\documentclass{article} 
\usepackage{iclr2026_conference}

\usepackage{times}
\usepackage{authblk}

\usepackage{amsmath,amsfonts,bm}

\def\eqref#1{equation~\ref{#1}}

\def\1{\bm{1}}

\DeclareMathAlphabet{\mathsfit}{\encodingdefault}{\sfdefault}{m}{sl}
\SetMathAlphabet{\mathsfit}{bold}{\encodingdefault}{\sfdefault}{bx}{n}

\usepackage{booktabs}
\usepackage{graphicx}
\usepackage{makecell}
\usepackage{hyperref}
\usepackage{url}
\usepackage{svg}
\usepackage{graphicx}
\usepackage{tabularx}
\usepackage{subcaption}
\usepackage{multirow}
\usepackage{siunitx}
\usepackage{amssymb}       
\usepackage{nicematrix}    
\usepackage[table]{xcolor}
\usepackage[inline]{enumitem}

\newcommand{\major}{$\blacksquare$}
\newcommand{\minor}{+}

\usepackage{tikz}
\usepackage{comicneue}
\usetikzlibrary{positioning, fit, shapes.geometric, arrows.meta, shadows.blur}

\usepackage{hyperref}
\usepackage{url}
\usepackage{longtable}
\usepackage{caption}

\newcommand{\MixtureVitae}{\textcolor{teal}{\textbf{MixtureVitae}}}
\newcommand{\DatasetSize}{422}

\usepackage{xcolor}
\usepackage{etoolbox} 
\newtoggle{comments}
\toggletrue{comments}     

\newcommand{\commentcolor}{red!65!black}
\newcommand{\DefaultInitials}{VM} 

\makeatletter
\newcommand{\cmt}[2][]{%
  \iftoggle{comments}{%
    \def\cmt@init{\if\relax\detokenize{#1}\relax \DefaultInitials \else #1\fi}%
    \ifmmode
      \text{\color{\commentcolor}\footnotesize\bfseries[\cmt@init:~#2]}%
    \else
      {\color{\commentcolor}\footnotesize\bfseries[\cmt@init:~#2]}%
    \fi
  }{}%
}
\makeatother

\newcommand{\mcmt}[2][]{%
  \iftoggle{comments}{%
    \marginpar{\raggedright\color{\commentcolor}\scriptsize
      \if\relax\detokenize{#1}\relax \DefaultInitials \else #1\fi:~#2}%
  }{}%
}

  {\iftoggle{comments}{%
     \par\begingroup\color{\commentcolor}\footnotesize\bfseries
     \noindent[\if\relax\detokenize{#1}\relax \DefaultInitials \else #1\fi]\quad
   }{\expandafter\cblock@swallow}}%
  {\iftoggle{comments}{\par\endgroup}{}}
\newenvironment{cblock@swallow}{}{}

\title{MixtureVitae: Open Web-Scale Pretraining Dataset With High Quality Instruction and Reasoning Data Built from Permissive-First Text Sources}

\author[1,3,4,*]{Huu Nguyen}
\author[1,*]{Victor May}
\author[1,2,*]{Harsh Raj}
\author[1,3,4,5]{Marianna Nezhurina}
\author[1,6]{Yishan Wang}
\author[7]{Yanqi Luo}
\author[8]{Minh Chien Vu}
\author[4,9]{Taishi Nakamura}
\author[4,15]{Ken Tsui}
\author[10]{Van Khue Nguyen}
\author[11,12]{David Salinas}
\author[13]{Aleksandra Krasnodębska}
\author[3]{Christoph Schuhmann}
\author[14]{Mats Leon Richter}
\author[16]{Xuan-Son Vu}
\author[1,3,4,5]{Jenia Jitsev}

\affil[*]{Equal contribution}
\affil[ ]{\textit{Correspondence to: \href{mailto:huu@ontocord.ai}{huu@ontocord.ai}}}
\affil[1]{Ontocord}
\affil[2]{Northeastern University}
\affil[3]{LAION}
\affil[4]{Open-$\Psi$ (Open-Sci) Collective}
\affil[5]{Juelich Supercomputing Center (JSC), Research Center Juelich (FZJ)}
\affil[6]{Carnegie Mellon University}
\affil[7]{Salesforce}
\affil[8]{Detomo Inc.}
\affil[9]{Institute of Science Tokyo}
\affil[10]{École Polytechnique, IP Paris}
\affil[11]{ELLIS Institute Tuebingen}
\affil[12]{University of Freiburg}
\affil[13]{NASK}
\affil[14]{Montreal Institute for Learning Algorithms, Université de Montréal}
\affil[15]{Independent Researcher}
\affil[16]{Lund University and DeepTensor AB}

\iclrfinalcopy 

\begin{document}

\maketitle

\begin{abstract}

We present \MixtureVitae{}, an open‑access\footnote{ \href{https://github.com/ontocord/mixturevitae}{https://github.com/ontocord/mixturevitae}} pretraining corpus built to minimize legal risk while providing strong downstream performance.  \MixtureVitae{} follows a permissive‑first, risk‑mitigated sourcing strategy that combines public‑domain and permissively licensed text (e.g., CC‑BY/Apache) with carefully justified low‑risk additions (e.g., government works and EU TDM‑eligible sources). \MixtureVitae{} adopts a simple, single-stage pretraining recipe that integrates a large proportion of permissive synthetic instruction and reasoning data—signals typically introduced during post-training and generally scarce in permissive web corpora. We categorize all sources into a three-tier scheme that reflects varying risk levels and provide shard-level provenance metadata to enable risk-aware usage. In controlled experiments using the open‑sci‑ref training protocol (fixed architectures and hyperparameters; 50B and 300B token budgets across 130M–1.7B parameters), models trained on \MixtureVitae{} consistently outperform other permissive datasets across a suite of standard benchmarks, and at the 1.7B-parameters/300B-tokens setting, they surpass FineWeb‑Edu and approach DCLM late in training. Performance is particularly strong on MMLU and on math and code benchmarks: a 1.7B model pretrained on 300B \MixtureVitae{} tokens matches or exceeds a strong 1.7B instruction‑tuned baseline on GSM8K, HumanEval, and MBPP, despite using over 36$\times$ fewer tokens (300B vs.\ $\approx$11T). Supported by a thorough decontamination analysis, these results show that permissive‑first data with high instruction and reasoning density, tiered by licensing and provenance-related risk, can provide a practical and risk-mitigated foundation for training capable LLMs, reducing reliance on broad web scrapes without sacrificing competitiveness.


\end{abstract}

\section{Introduction}

The proliferation of large language models (LLMs) has transformed the landscape of artificial intelligence, yet their development often relies on a legally and ethically precarious foundation. The vast majority of performant models are pretrained on massive web scrapes, indiscriminately mixing public-domain content with copyrighted materials such as books, news articles, and personal websites without explicit permission \citep{raffel2020exploring, gao2020pile}. This practice has led to a growing number of copyright infringement lawsuits,  creating significant legal uncertainty for both
academic researchers and commercial developers and threatening the future of the field. At the same
time, practitioners who wish to avoid this risk have few alternatives, as most high-performing
pretraining mixtures rely, at least in part, on opaque or non-permissive web scrapes.

Compounding this uncertainty is the prevailing assumption that state-of-the-art performance is inextricably linked to the sheer scale and diversity offered by these legally ambiguous web scrapes. The absence of a high-performance, large-scale pretraining dataset that actively mitigates these risks has forced a difficult choice between performance and compliance. In practice, the strongest open
baselines such as FineWeb-Edu~\citep{penedo2024} and DCLM~\citep{li2024datacomplm} still rely on mixed-license or unspecified web data,
whereas strictly permissive corpora tend to lag behind them on reasoning-heavy benchmarks.
This raises a critical question: Can
a powerful language model be trained on a dataset that provides a more legally robust foundation?

To this question, we answer "yes": We introduce \MixtureVitae{}, a \textbf{\DatasetSize{}}-billion-token, open-access pretraining dataset constructed to minimize copyright risk while explicitly demonstrating that a
reasoning- and instruction-dense, permissive-first mixture can substantially close the performance gap to leading
non-permissive corpora. The core of \MixtureVitae{}'s "permissive-first" data comprise (1) text with clear and permissive licenses (e.g., CC-BY-*, Apache 2.0), public-domain text, and copyright-exempt text such as US federal works (see Appendix~\ref{app:govt_works}) and (2) risk-mitigated text. Following Phi-4~\citep{abdin2024phi4technicalreport}, which shows that the addition of synthetic and web-rewrite data boosts performance, we address the scarcity of organic reasoning and conversational dialogue in strictly permissive sources significantly augmenting \MixtureVitae{} with targeted synthetic data, which is derived from permissive models and sources. We call this combination of expressly licensed and risk-mitigated methods the \textbf{"permissive-first"} approach.

To validate our approach, we train models with \textbf{130M, 400M, 1.3B, and 1.7B parameters} on \MixtureVitae{} and compare their performance against several prominent open datasets. The results first confirm that \MixtureVitae{} \textbf{significantly outperforms all other permissively licensed baselines}, with the performance gap widening as the model scale increases. The more critical test, however, is against popular non-permissive datasets containing higher proportions of copyrighted or ambiguously-licensed material. In this setting, our models achieve competitive performance, and on math and code benchmarks, our 1.7B base model matches or exceeds a strong 1.7B instruction-tuned baseline (SmolLM2) despite being trained on a dramatically smaller budget (over $36\times$ fewer tokens).

In summary, our contributions are threefold:
\begin{itemize}
\item
\textbf{Permissive-first, risk-mitigated, and performant recipe for pretraining corpora.} We present \MixtureVitae{}, the first highly-performant, permissive-first, and risk-mitigated pretraining corpus that deliberately front-loads high-quality reasoning and instruction data to drive capability gains in small models. It is organized into auditable provenance tiers and constructed via a positive-inclusion pipeline, avoiding the need for retroactive filtering.
\item
\textbf{We demonstrate that reliance on indiscriminately scraped, high-risk copyrighted data is not a prerequisite for training capable LLMs.}
Leveraging the \texttt{open-sci-ref}~\citep{nezhurina2025opensciref001openreproduciblereference} protocol to ensure rigorous comparison across 130M–1.7B parameter scales, we demonstrate the value of front-loading instruction and reasoning data into pre-training. Our \DatasetSize{}B-token, permissive-first mixture closes the gap to mixed-license baselines while providing an auditable legal provenance. Furthermore, we show that our 1.7B base model, despite a limited 300B token budget, is comparable across multiple reasoning benchmarks to a strong 1.7B instruction‑tuned baseline—trained on roughly $36\times$ more tokens ($\approx$11T).

\item
\textbf{Evaluation integrity and reusable artifacts}. We perform a large-scale 13-gram decontamination analysis across all benchmarks, showing that \MixtureVitae{}’s gains persist on decontaminated test sets and when removing shards responsible for most detected overlap, and we release the corpus, shard-level provenance metadata, and curation code to enable compliant, reproducible pretraining in future work.

\end{itemize}

\section{Dataset}

We adopt a permissive-first, risk-mitigated strategy, combining sources with clear permissive licenses (e.g. CC-BY, Apache, public domain) with narrowly justified inclusions (government works, EU TDM-eligible data) and targeted synthetic data. Within this framework, the \MixtureVitae{} dataset is constructed from three primary categories:  curated sources for domain-specific expertise, diverse web data for language and general knowledge and instruction-following and reasoning datasets to enhance reasoning and task-completion abilities. 

\begin{figure}[htbp]
    \centering

    \begin{subfigure}{0.48\textwidth}
        \centering
        \includegraphics[width=\linewidth]{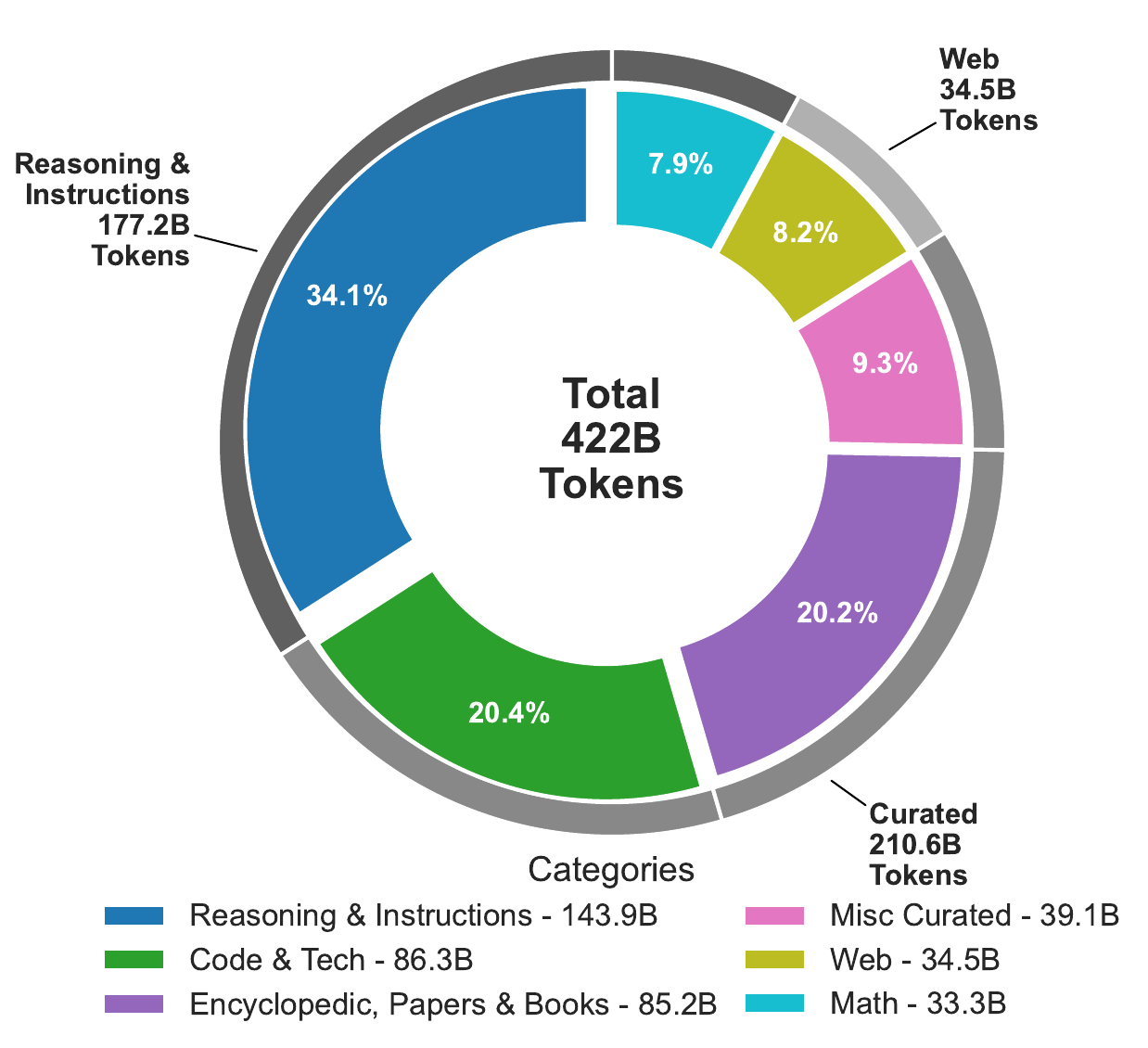}
        \caption{Dataset Composition by Top-Level Category and Content Domain}
        \label{fig:domain_comp}
    \end{subfigure}%
    \begin{subfigure}{0.48\textwidth}
        \centering
        \includegraphics[width=\linewidth]{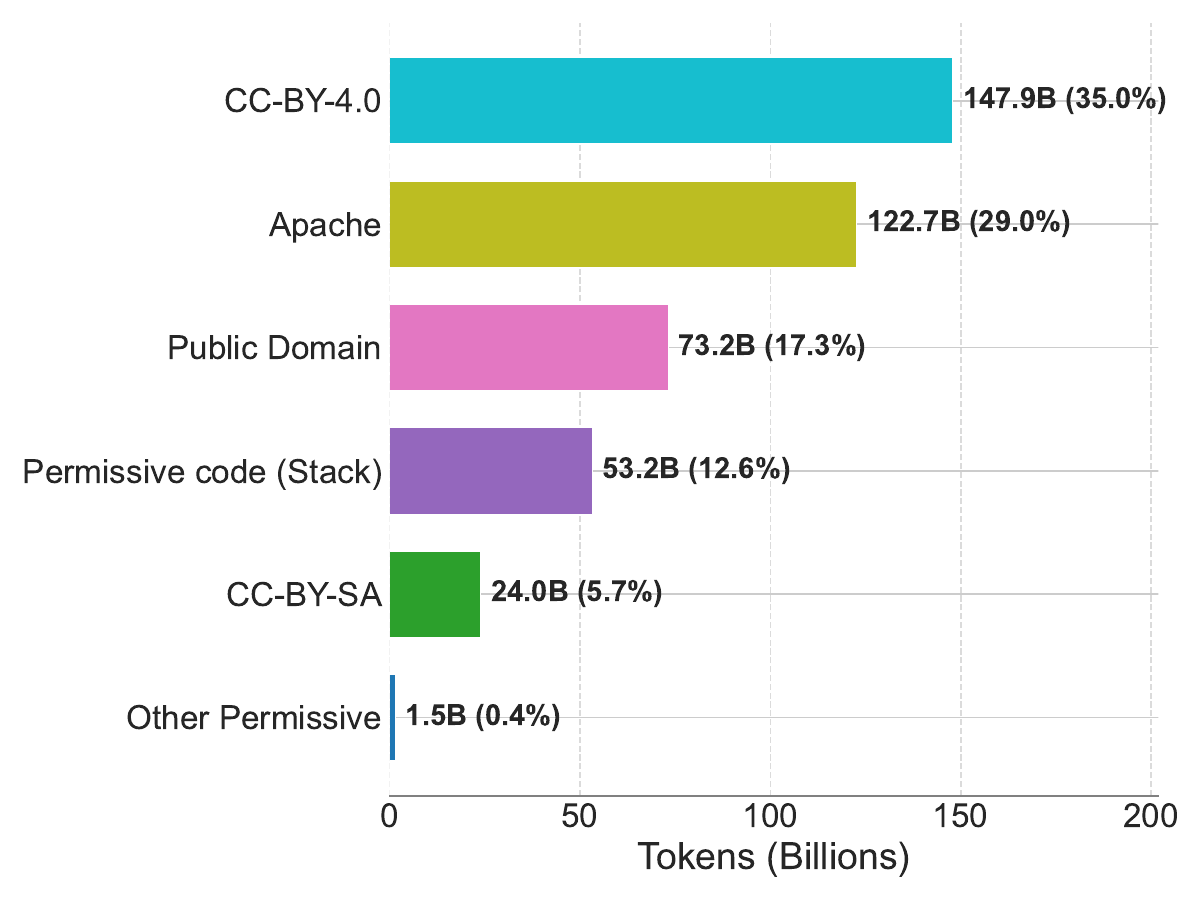}
        \caption{Token Distribution by Governing License}
        \label{fig:license_comp}
    \end{subfigure}

    \caption{Composition of the \MixtureVitae{} dataset (permissive-first, risk-mitigated composition).}
    \label{fig:composition_combined}
\end{figure}

The major categories of our corpus are visualized in Figure~\ref{fig:domain_comp}. We provide a granular breakdown showing the token count for each component (Figure~\ref{fig:data_comp_deets}), the license distribution (Figure~\ref{fig:license_comp}), and synthetic data usage (Figure~\ref{fig:synthetic_composition}). Specific data sources are detailed in the following subsections.

\subsection{Data Sources}
\label{sec:data_sources}
Our dataset selection process is governed by a two-layer  criteria, prioritizing risk mitigation followed by quality and capability objectives:

\begin{itemize}
    \item \textbf{Legal \& Licensing:} The primary filter is legal compliance. A dataset is considered only if it operates under a clear permissive license (e.g., CC-BY, Apache 2.0) or is in the public domain. For synthetic data, we further scrutinize the provenance of seed corpora and generator models (Appendix~\ref{app:synth_prov}). The majority of our synthetic sources satisfy full provenance transparency (classified as Tier 1), while a minority of community reasoning datasets with opaque provenance are categorized as Tier 2 to manage residual risk.
    \item \textbf{Quality \& Capability:} Among compliant sources, we prioritize datasets with prior evidence of high performance in community mixtures (e.g., \citealp{peS2o}). Furthermore, to address the reasoning deficits typical of strictly permissive web scrapes, we target high-density instruction and reasoning data, a choice driven by the need to boost performance on tasks such as GSM8K~\citep{cobbe2021training} and MMLU\citep{hendrycks2021mmlu}.
\end{itemize}

The following sections describe each of the three categories of data in \MixtureVitae{}: web, curated sources, and instruction and reasoning datasets.

\subsubsection{Web-Scale Corpora}
One subset of our pre-training data is derived from web-scale datasets including Nemotron-CC~\citep{su2025nemotroncc}, MGACorpus~\citep{hao2025reformulationpretrainingdataaugmentation}, and FineFineWeb~\citep{map2024finefineweb}. It also contains synthetic data generated by rephrasing web text from Nemotron-CC and MGACorpus.




\subsubsection{Curated Datasets}
To incorporate domain-specific knowledge and high-quality text, we curate diverse sources: public financial documents from SEC EDGAR~\citep{sec_edgar}, multilingual encyclopedic articles from MegaWika~\citep{barham2023megawikamillionsreportssources} and TxT360~\citep{txt360}, scientific papers from arXiv~\citep{clement2019use} and peS2o~\citep{peS2o}, medical data from Pubmed~\citep{PubMed}, code from The Stack v1~\citep{kocetkov2023the}, patents from the USPTO database~\citep{uspto} and EuroPat~\citep{heafield-etal-2022-europat}, mathematical problems from Deepmind Math~\citep{saxton2019analysingmathematicalreasoningabilities}, and video transcripts from both VALID~\citep{Huu2024VALID} and the YouTube Commons corpus~\citep{Langlais2024}, news and law data from the Open License Corpus~\citep{min2024silo}.

We source 12.6\% of our dataset from \textbf{The Stack v1}, a permissive-first, risk-mitigated code dataset governed by the OpenRAIL-M license. We discuss its permissiveness situation in Appendix~\ref{app:stack}.

\subsubsection{Instruction and Reasoning Datasets}
To enhance instruction-following and reasoning, we follow \cite{abdin2024phi4technicalreport} by including considerable synthetic and web-rewrite data. We extensively use fully and partially synthetic data --- all generated from permissive or public-domain seed data using models under permissive licenses.

\paragraph{General Instruction Following} We include a strong instruction-following baseline with the Magpie Collection~\citep{xu2024magpiealignmentdatasynthesis}, its derivatives (e.g., Magpie-Phi3-Pro). This is augmented with preference data from UltraFeedback~\citep{cui2024ultrafeedback} and NVIDIA's SFT data blend~\cite{nvidia_sft_datablend_v1}, which contains a curated mixture of permissively licensed subsets from public datasets, including OASST~\citep{köpf2023openassistant}, CodeContests~\citep{Li_2022}, FLAN~\citep{chung2022scalinginstructionfinetunedlanguagemodels}, OpenPlatypus~\citep{platypus2023}, and the training split of GSM8K~\citep{cobbe2021training}. Additionally, we augment the P3~\citep{sanh2021multitask} dataset with a few-shot and multiple-choice format.

\paragraph{Reasoning} To improve reasoning, we incorporate general corpora such as Glaive-AI Reasoning~\citep{glaive_reasoning_2023} and OpenThoughts~\citep{guha2025openthoughts} as well as domain-specific datasets: the legal dataset CaseHOLD~\citep{zheng2021does}, scientific Q\&A from the OpenScience collection~\citep{NVIDIAOpenScience2025}, and agent-focused instructions from OpenManus-RL~\citep{openmanus_rl}.

\paragraph{Mathematics and Coding} To strengthen quantitative reasoning, we combine our internally developed synthetic Math Word Problems dataset~(Appendix \ref{app:mwp}) with established datasets like MetaMathQA~\citep{yu2024metamathbootstrapmathematicalquestions} and DM-Math~\citep{saxton2019analysingmathematicalreasoningabilities}, further enriched with large-scale math instruction sets, including OpenMathInstruct-2~\citep{toshniwal2024openmathinstruct118millionmath}, DART-MATH~\citep{tong2024dartmath}, Nemo-Math~\citep{mahabadi2025nemotron}, and Prism-Math~\citep{nvidia2025prismmath}.
 For coding, we combine the Ling Coder collection~\cite{codefuse2025samplemattersleveragingmixtureofexperts} with executable instructions from the StarCoder dataset~\cite{kocetkov2023the} to target a wide range of software engineering tasks.
 

\begin{figure*}[htbp]
\centering
\begin{subfigure}[t]{0.49\textwidth}
    \centering
    \includegraphics[width=\textwidth]{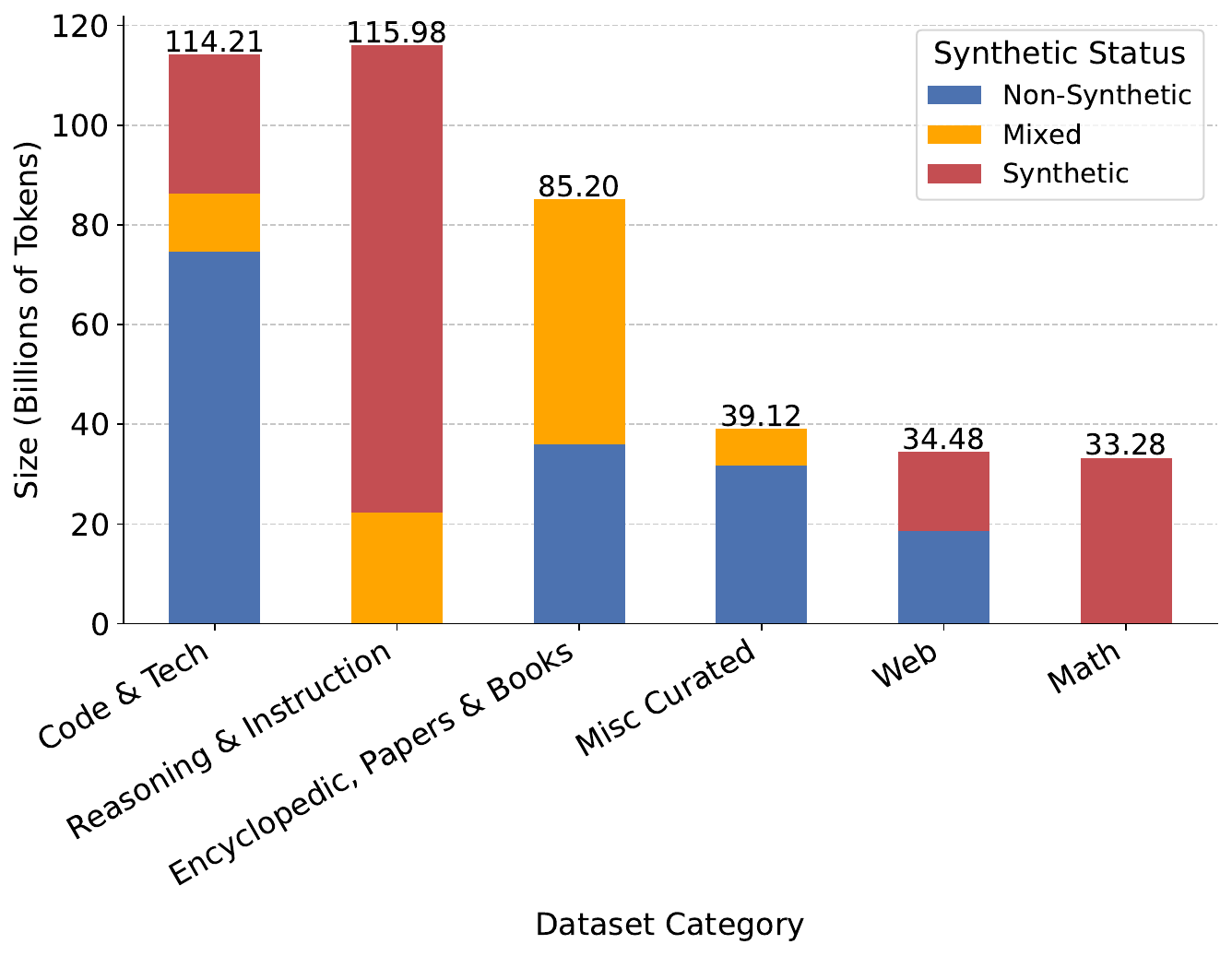}
    \caption{\MixtureVitae{} composition by origin (total token counts at the top in billions). Each bar represents one of the six primary content domains (as in Figure~\ref{fig:domain_comp}), segmented by source type: \textbf{Non-Synthetic} (real human-written text and code), \textbf{Mixed} (sources with partial synthetic data), and \textbf{Synthetic} (data generated by permissive models from permissive seeds).}
    \label{fig:synthetic_composition}
\end{subfigure}
\hfill 
\begin{subfigure}[t]{0.49\textwidth}
    \centering
    \includegraphics[width=\textwidth]{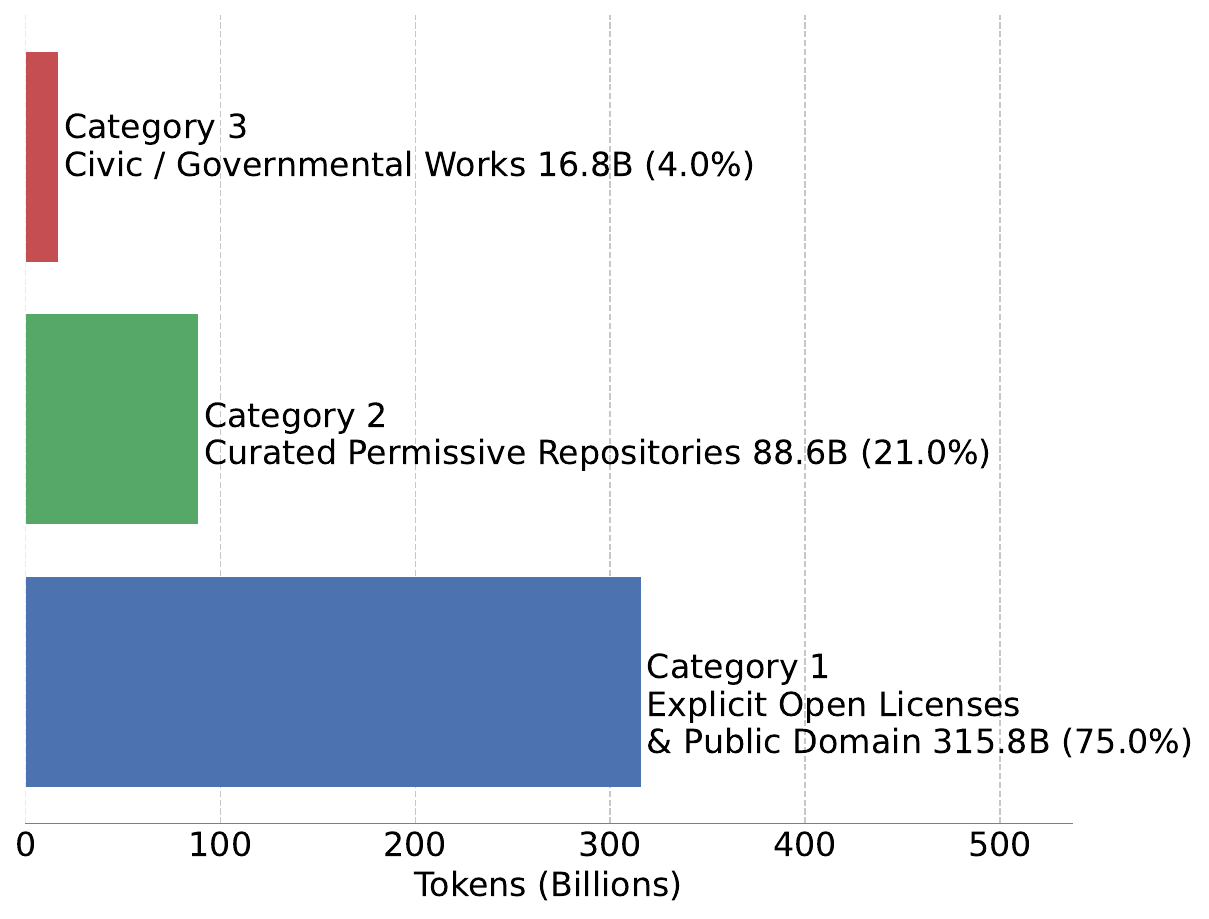}
    \caption{Legal provenance and risk-mitigation tiers of the \MixtureVitae{} corpus. The dataset is segmented into its three constituent legal categories, with all sources falling into a permissive-first or risk-mitigated tier. Token counts (billions) and total corpus percentages are shown for each category.}
    \label{fig:provenance_tiers}
\end{subfigure}
\caption{Composition and provenance of \MixtureVitae{}: \textbf{(a)} Synthetic-status distribution across the six content domains, \textbf{(b)} licensing tiers and risk posture for the corpus.}
\label{fig:mv-synth-legal}
\end{figure*}

\subsubsection{Licensing Tiers and Risk Profiles}
\label{sec:license_tiers}

To make the provenance and legal footing of \MixtureVitae{} transparent, we conceptualize all dataset components into \emph{tiers} based on license type and expected risk profile (see Figure~\ref{fig:provenance_tiers} and Table~\ref{tab:provenance}).\footnote{The high-level groupings presented in this section (e.g., ``Code \& Tech'', ``Reasoning'') and the shard breakdowns in the Appendices are primarily organizational abstractions for visualization and provenance tracking. In practice, the actual training data construction follows a granular \textit{domain-aware mixing strategy} (detailed in Section~\ref{sec:training_curation}), where documents are clustered by base URL or provenance to preserve domain coherence per sample, rather than strictly sampling from rigid high-level partitions.} 

\paragraph{Tier 1 --- Explicit Open Licenses \& Public Domain.}  
This tier encompasses text and code under clear permissive licenses (e.g., CC0, CC-BY, Apache 2.0, MIT, BSD, a permissive subset of P3) or in the public domain, such as encyclopedic resources, scientific papers, and portions of curated math corpora. Because licenses are explicit and permissive, the legal risk of reuse is minimal. This tier also includes synthetic data generated from permissively licensed models and seed data. 

\paragraph{Tier 2 --- Curated Permissive Corpora with Upstream Opacity.}

\begin{itemize}
    \item(a) \textbf{Permissive Corpora With Partial or Unverified Provenance.}
    This subset includes resources such as \textsc{The Stack V1} and Wikipedia-derived corpora. The released dataset all carries a permissive license, and curators apply filters (e.g., repository-level license heuristics). However, because provenance is only partially tracked at the file or example level, there remains some residual uncertainty about the licensing status of individual items, hence its separation from Tier~1. This Tier also includes datasets that have no license, but the underlying data is public domain or permissive and requiring the same license as the upstream data, or where the data is solely obtained synthetically from a model that is permissively licensed. 
    \item (b) \textbf{Synthetic Data with Non-Permissive or Unverifiable Generators or Seeds.} This tier contains datasets that are themselves permissively licensed (e.g., Apache/MIT/CC-BY), but where either (i) the generator model used to create the synthetic data operates under a more restrictive license (e.g., Llama-3 community license, OpenAI API terms), or (ii) the seed data contains slices whose provenance cannot be fully audited (e.g., partially opaque community mixtures). These datasets constitute only $\approx$4\% of \MixtureVitae{} and are isolated for transparency so that users who require a strictly permissive generator and seed provenance can exclude them (more detail in Table~\ref{tab:provenance}). 

\end{itemize}

\paragraph{Tier 3 --- Civic / Governmental Works.}  
This tier includes materials that are either statutory public domain (e.g., U.S. federal works) or under a strong public-purpose rationale for reuse (e.g., government websites, regulatory notices). While not always explicitly licensed, such work---typically created for dissemination---is widely recognized as low-risk for inclusion. Filtering with copyright keyword checks further reduces the possibility of inadvertently including restricted content.


\subsection{Data Processing Pipeline}
\label{sec:data_proc_pipe}
To transform the raw data sources into a high-quality and permissively licensed pretraining corpus, we develope a multistage data processing pipeline. Our curation pipeline includes the following stages: ensuring permissive licensing, filtering for CSAM and offensive language, improving overall content quality, and reducing data redundancy. The following sections detail each component.

\subsubsection{Permissiveness Filtering}
\label{sec:positive_inclusion}

In contrast to standard data pipelines that rely on the retroactive negative filtering of broad web scrapes (e.g., \citealp{fan2025can}), we employ a \textbf{positive inclusion} strategy for web data. Rather than ingesting broad web dumps and filtering post-hoc, we positively select sources based on auditable permissive status. Specifically, we (i) apply an explicit allowlist of governmental and international domains (Appendix~\ref{app:gov_domains}), (ii) curate a set of websites with known permissive licenses (Appendix~\ref{app:permissive_domains}), and (iii) expand this set with risk-mitigated documents by searching for permissive license keywords (e.g., “CC-BY-SA”), excluding documents with restrictive terms (e.g., “all rights reserved”). This upfront design minimizes the risk of including paywalled or opted-out content (e.g., commercial news). We justify the inclusion of governmental works under a strong fair-use rationale, considering their public purpose, content type, and minimal market impact (Appendix~\ref{app:govt_works}).


\subsubsection{Safety Filtering}
We remove obscene, adult and CSAM-related content with keyword-based blocklists adapted from prior work~\citep{NEURIPS2022_ce9e92e3, nakamura2024auroramopensourcecontinual}. For Wikipedia-based documents, we remove articles about films, sporting events, and biographies of living persons in English with applied targeted filtering, to minimize memorization of facts about people, in case of objection to incorrect facts about people being generated by models trained on \MixtureVitae{}.  Besides dataset-level filters, we also evaluate the final model's safety profile via standard red-teaming (Appendix~\ref{app:red_teaming}).

\subsubsection{Quality Filtering}
Per standard practices~\citep{raffel2020exploring}, we remove documents with base64-encoded text (which can disrupt training) and duplicative headers and footers (e.g., "Home | Search") from FineFineWeb. 

\subsubsection{Deduplication}

Informed by recent findings in large-scale data curation, our deduplication strategy prioritizes diversity over purity. While removing exact repetitions mitigates harmful memorization \citep{lee2022deduplicating}, prior research finds that aggressive, global near-duplicate removal can be detrimental. For example, the creators of the \textbf{FineWeb-Edu} dataset \citep{penedo2024} reported \textit{worsened} model performance by global fuzzy deduplication, postulating that it removed ``too much quality data.''

Therefore, we adopt a local-only approach. We first apply \textbf{intra-dataset deduplication} using prefix-based exact matching to remove verbatim boilerplate text ~\citep{lee2022deduplicating}. We \textbf{intentionally avoid full, cross-dataset fuzzy deduplication} to preserve near-duplicates (e.g., Wikipedia articles with different formatting across sources). We posit that doing so retains \textbf{``stylistic and domain diversity,''} a factor shown to be helpful for model generalization \citep{chen2024diversity}.



\subsubsection{Training Example Curation}
\label{sec:training_curation}
Our process for creating training examples involves several stages:
\begin{enumerate}
    \item \textbf{Heuristic Cleaning:} We remove boilerplate content by eliminating repetitive n-gram prefixes and suffixes, following standard web data cleaning pipelines~\citep{raffel2020exploring}.
    \item \textbf{Fine-grained Deduplication:} To enhance data quality, we segment documents into sentences and remove duplicate sentences within each document. Documents with high internal repetition (sentence duplication rate > 75\%) are discarded entirely, as this has been shown to improve model performance~\citep{lee2022deduplicating}.
    \item \textbf{Domain-Aware Mixing:} To construct the final training examples, we employ a domain-aware data mixing strategy~\citep{xie2023doremioptimizingdatamixtures}. Documents are clustered by their base URL (a proxy for domain), and sentences are concatenated first within their original document, then packed with other documents from the same cluster. 
\end{enumerate}

\subsubsection{Additional Filtering for Synthetic Datasets}
To ensure that the synthetic subsets of \MixtureVitae{} adhere to our permissive-first, risk-mitigated approach, we prioritize data originating from seeds that are sourced from permissive sources and generated with models that are themselves permissively licensed. A small portion ($\approx$4\%) of \MixtureVitae{} originates from sources with restricted, mixed, or opaque provenance and is isolated into Tier~2(b), as detailed in Appendix~\ref{app:synth_prov} and Table~\ref{tab:provenance}.

\section{Experiments}
\label{sec:experiment_results}
\subsection{Experimental Setup}
To empirically validate the quality of the \MixtureVitae{} pretraining dataset, we conduct a large-scale comparative study against a selection of prominent open pretraining datasets. We isolate the impact of the dataset on downstream performance using the \textbf{open-sci-ref} training procedure~\citep{nezhurina2025opensciref001openreproduciblereference}, which enables systematic control of factors affecting benchmark scores. As in \texttt{open-sci-ref}, we fix the model architecture (Table ~\ref{tab:opensciref_scales}, sizes: 0.13B, 0.4B, 1.3B, 1.7B) and training hyperparameters (Table~\ref{tab:training_schedules}), varying only the dataset. This design ensures that any performance difference can be attributed solely to the dataset. 

Also, following the numbers given in \texttt{open-sci-ref}, we train each model on two token budgets: 50B and 300B, to analyze scaling effects. Conducting separate training runs on each budget, rather than using intermediate checkpoints, thus ensuring a consistent data distribution and allowing for proper optimization of learning rate schedules for each specific token budget \citep{hoffmann2022}. This follows standard practice: Data mixtures effective at small token budgets may not generalize to larger ones \citep{albalak2023efficientonlinedatamixing}.

To guard against test‑set leakage, we also perform a large‑scale 13‑gram decontamination analysis and re‑evaluation; Section~\ref{sec:exp_contam} and Appendix~\ref{app:contamination} detail this procedure.



Within this controlled evaluation framework, we compare \MixtureVitae{} with the set of public baselines evaluated in \texttt{open-sci-ref}, with the addition of a representative selection of permissively licensed datasets. As detailed in Table~\ref{tab:dataset_comparison}, the comparison set includes two groups:
\begin{itemize}
    \item \textbf{Non-Permissive/Mixed-License Baselines.} C4~\citep{raffel2020exploring}, The Pile~\citep{gao2020pile}, SlimPajama~\citep{shen2024slimpajamadcunderstandingdatacombinations}, FineWeb-Edu~\citep{penedo2024}, Nemotron-CC-HQ~\citep{su2025nemotroncc}, DCLM-baseline~\citep{li2024datacomplm}, HPLT Monolingual Datasets v2.0~\citep{burchell2025};
    \item \textbf{Permissive Baselines.} CommonCorpus and its English subset~\citep{langlais2025commoncorpuslargestcollection}, as well as Comma-0.1~\citep{kandpal2025commonpilev018tb}. 
\end{itemize}



All datasets are tokenized using the GPT-NeoX-20B tokenizer~\citep{black2022gptneox}, resulting in a vocabulary size of 50,304. The models are trained using Megatron-LM~\citep{shoeybi2020megatronlmtrainingmultibillionparameter}, and the evaluations are performed using LM Evaluation Harness~\citep{eval-harness}. 

Model performance is evaluated on recognized downstream task benchmarks: MMLU~\citep{hendrycks2021mmlu}, COPA~\citep{copa}, LAMBADA~\citep{paperno2016lambada}, OpenBookQA~\citep{mihaylov2018openbookqa}, Winogrande~\citep{sakaguchi2020winogrande}, ARC (Challenge and Easy)~\citep{clark2018arc}, BoolQ~\citep{clark2019boolq}, HellaSwag~\citep{zellers2019hellaswag}, Commonsense-QA~\citep{talmor2019commonsenseqa} and PIQA~\citep{bisk2019piqareasoningphysicalcommonsense}. 

To ensure evaluation integrity, we perform a comprehensive decontamination analysis against all benchmark test sets, with full details and case studies provided in Appendix~\ref{app:contamination}.



\subsection{Experiment Results}

\begin{figure}[htb!]
    \centering

    \begin{subfigure}[t]{0.48\textwidth}
        \centering
        \includegraphics[width=\linewidth]{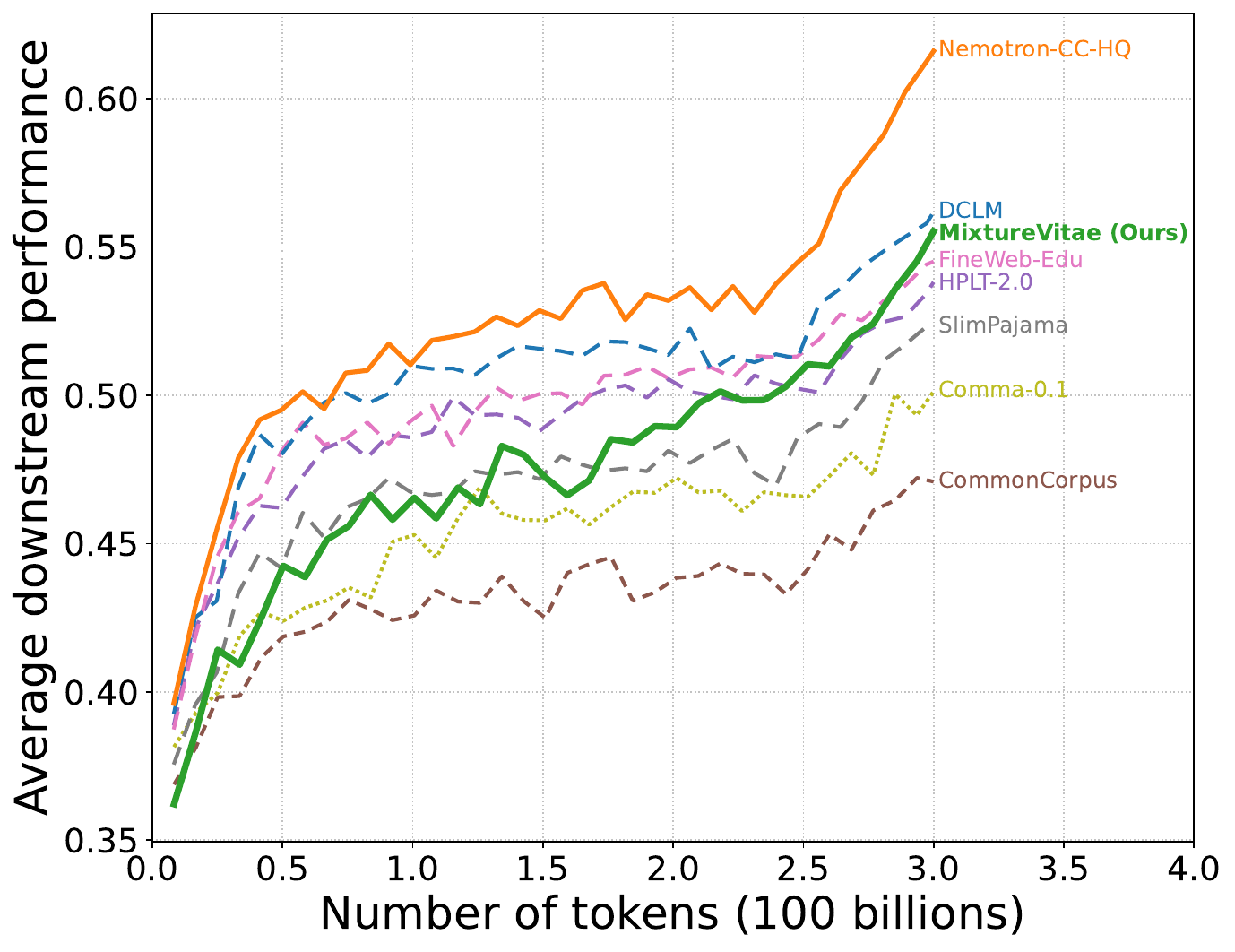}
          \vspace*{-0.5cm}
          \caption{Average performance on all 11 evaluated tasks.}
        \label{fig:average_all_performance}
    \end{subfigure}%
    \hfill 
    \begin{subfigure}[t]{0.48\textwidth}
        \centering
        \includegraphics[width=\linewidth]{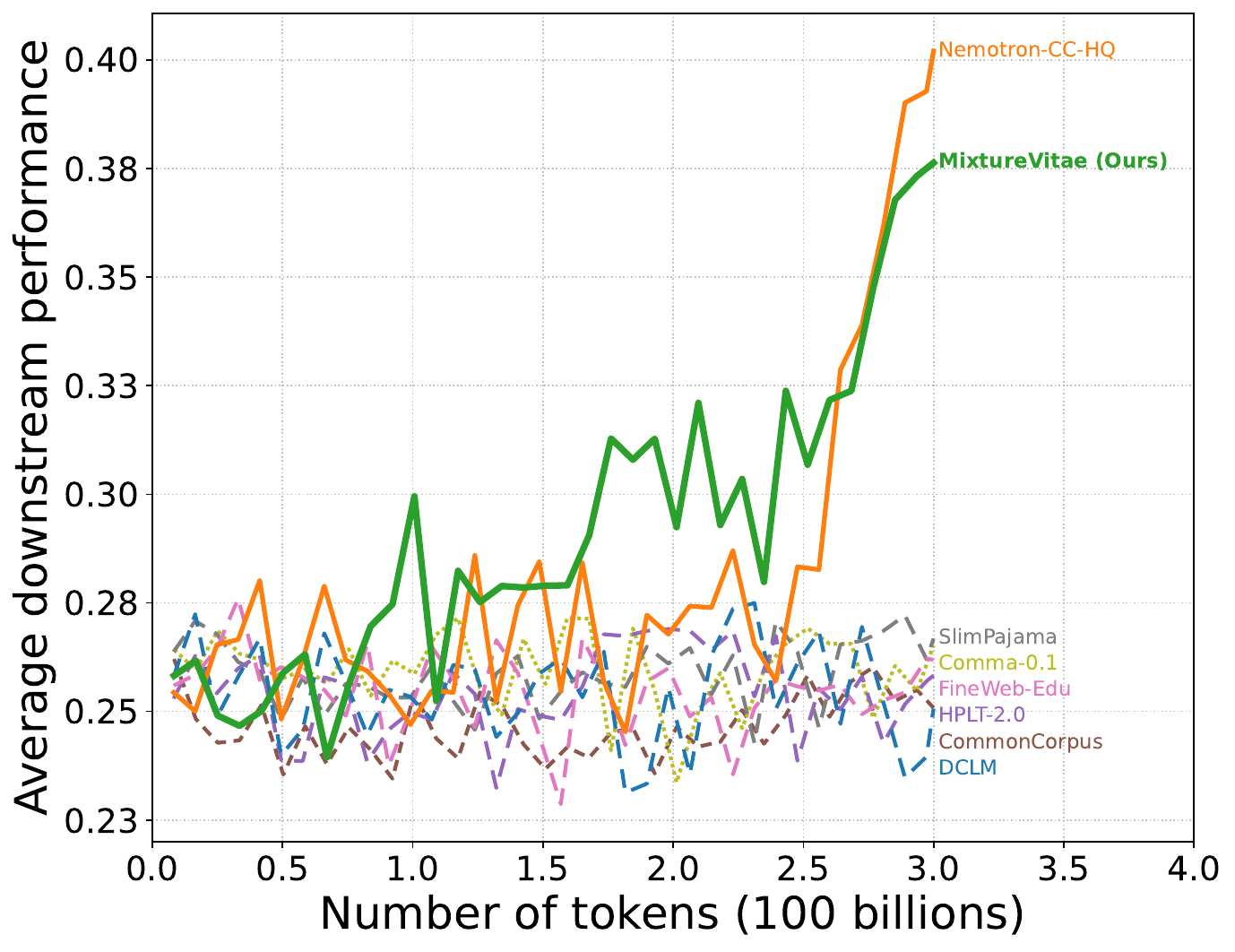}
          \vspace*{-0.5cm}
          \caption{Performance on MMLU.}
        \label{fig:average_mmlu_performance}
    \end{subfigure}%

    \caption{Performance comparison of pretraining datasets for a 1.7B-parameter model trained up to a 300B token budget, showing downstream accuracy as a function of the number of training tokens.
    }

\label{fig:300b_performance}
\end{figure}

\begin{table}[!htb]
    \centering
    \footnotesize
\caption{Performance comparison of 1.7B-parameter models trained on different pretraining datasets with a 300B token budget.
\textit{Italic} denotes the best result among permissive-only datasets, while \textbf{bold} indicates the best result overall, including mixed-license datasets.
\MixtureVitae{} outperforms other permissive datasets across most benchmarks. On reasoning related MMLU, BoolQ, and CommonSense-QA, it also outperforms strong non-permissive baselines.}
\label{tab:lang_results}
    \footnotesize
\begin{tabular}{lrrrrr}
\toprule
Benchmark & \makecell{\MixtureVitae{}\\\scriptsize (permissive)} &
\makecell{Comma-0.1\\\scriptsize (permissive)} &
\makecell{CommonCorpus\\\scriptsize (permissive)} &
\makecell{FineWeb-Edu\\\scriptsize (mixed-license)} &
\makecell{DCLM\\\scriptsize (mixed-license)}\\
\midrule
COPA  & \textit{0.73} & 0.71 & 0.71 & 0.76 & \textbf{0.81} \\
Lambada  & 0.48 & \textit{0.54} & 0.49 & 0.52 & \textbf{0.65} \\
OpenBookQA  & \textit{0.35} & 0.33 & 0.31 & \textbf{0.42} & 0.39 \\
Winogrande & 0.58 & \textit{0.60} & 0.56 & 0.61 & \textbf{0.62} \\
MMLU & \textbf{\textit{0.38}} & 0.27 & 0.25 & 0.26 & 0.25 \\
ARC-Challenge  & \textit{0.40} & 0.36 & 0.32 & \textbf{0.44} & 0.40 \\
ARC-Easy & \textit{0.71} & 0.63 & 0.61 & \textbf{0.75} & 0.73 \\
BoolQ  & \textbf{\textit{0.75}} & 0.62 & 0.62 & 0.67 & 0.69 \\
CommonSense-QA  & \textbf{\textit{0.49}} & 0.21 & 0.19 & 0.19 & 0.20 \\
HellaSwag  & \textit{0.54} & 0.53 & 0.45 & 0.63 & \textbf{0.67} \\
PIQA  & 0.70 & \textit{0.71} & 0.66 & \textbf{0.76} & \textbf{0.76} \\
\midrule
Average & \textbf{\textit{0.56}} & 0.50 & 0.47 & \textbf{0.55} & \textbf{0.56} \\
\bottomrule
\end{tabular}
\end{table}

\textbf{Overall average performance.} At a 300B-token budget, \MixtureVitae{} shows strong performance when compared to the reference permissive datasets and is almost comparable to the non-permissive datasets (Figure~\ref{fig:300b_performance}, Tab.~\ref{tab:lang_results}). \MixtureVitae{} outperforms all permissive dataset baselines by a significant margin, with gaps widening considerably for larger model sizes, in terms of average performance across all 10 tasks (see Figure~\ref{fig:average_all_performance}, Tab.~\ref{tab:lang_results}). Non-permissive datasets, particularly Nemotron-CC-HQ and DCLM, still achieve the highest overall performance. Approaching the 300B token budget, \MixtureVitae{} catches up to FineWeb-Edu and DCLM. More importantly, while the top-performing models are still trained on non-permissive datasets like Nemotron-CC-HQ and DCLM, our results demonstrate that this performance gap is no longer an inevitability. \MixtureVitae{} proves that a dataset built on a fully permissive, risk-mitigated foundation can achieve highly competitive results—significantly outperforming all other permissive baselines and landing within a small, practical margin of top-tier, legally-ambiguous corpora. This finding directly challenges the prevailing assumption that reliance on high-risk, indiscriminately scraped copyrighted data is a prerequisite for training capable LLMs. \MixtureVitae{} performs particularly well relative to others on reasoning related tasks like \textbf{MMLU} (Figure~\ref{fig:average_mmlu_performance}, Tab.~\ref{tab:lang_results}), where most baselines are near random chance. Among all the baselines, only Nemotron-CC-HQ catches up to \MixtureVitae{} at around 260B and overtakes it past that point. Our findings also hold at the 50B token budget scale (App. Sec.~\ref{app:50b_exp}).

\textbf{Performance on single tasks.} We show performance on each single task in Tab.~\ref{tab:lang_results} and in the App. Sec.~\ref{app:additional_exp_tasks_300B} (App. Fig.~\ref{fig:per_dataset_performance}). \MixtureVitae{} outperforms other permissive datasets on MMLU, Arc Challenge, Arc Easy and BoolQ, while closely matching DCLM and FineWeb-Edu. On PIQA, HellaSwag, Winogrande, OpenBookQA, \MixtureVitae{} is on par with Comma-0.1, while both are behind non-permissive datasets. Lambada is the only task where \MixtureVitae{} falls behind Comma-0.1. We thus observe \MixtureVitae{} to be particularly strong on reasoning-related tasks.

\subsection{Results on Problem Solving and Instruction-Based Downstream Tasks}
\label{sec:instruct_results}

To further demonstrate the performance of the \MixtureVitae{} dataset, we evaluate the model on a set of math, code, and instruction benchmarks: GSM8k~\citep{cobbe2021training}, MBPP~\citep{austin2021programsynthesislargelanguagen}, IF-Eval~\citep{zhou2023instruction}.  Our evaluation uses the final 1.7B model checkpoints after training for 300B tokens using the \texttt{open-sci-ref} protocol (exact evaluation setup in Table~\ref{tab:reasoning-eval-settings}).

Unlike traditional web-only baselines (e.g., C4, FineWeb, DCLM), \MixtureVitae{} utilizes a \textit{reasoning and instruction-heavy} pretraining mixture. Compared against base models with same architecture and matched training compute, this front-loading strategy shows capabilities typically associated with post-training. This pretraining composition leads to a more token-efficient and simple path to reasoning competence already after single base model pre-training stage, matching or outperforming conventional multi-stage extensive pre- and post-training procedures.




\begin{table}[!htb]
    \centering
\caption{\textbf{Performance on math, code, and instruction-following tasks for 1.7B models.}
We compare \MixtureVitae{}—trained on a reasoning- and instruction-heavy, permissive-first
mixture—against standard \texttt{open-sci-ref} baselines trained on predominantly web-based
corpora. \MixtureVitae{} shows a substantial lead in math and code tasks. Notably, the
1.7B \MixtureVitae{} base model exceeds \textbf{SmolLM2-1.7B-Instruct} on GSM8K,
HumanEval, and MBPP despite training on 300B rather than $\approx$11T tokens.}

\label{tab:instruct_results}
    \footnotesize
\begin{tabular}{llrrrrr}
\toprule
Training Dataset & Tokens & IF-Eval & GSM8K & HumanEval & MBPP & Average \\
\midrule
\multicolumn{7}{@{}l}{\textit{Models Trained with \texttt{open-sci-ref}} for 300B Tokens} \\
\MixtureVitae{} & 300B & 0.19 & \textbf{0.53} & \textbf{0.32} & \textbf{0.38} & \textbf{0.36} \\
Comma-0.1 & 300B & 0.19 & 0.06 & 0.13 & 0.22 & 0.15 \\
CommonCorpus & 300B & 0.13 & 0.02 & 0.05 & 0.05 & 0.06 \\
C4 & 300B & 0.20 & 0.02 & 0.00 & 0.00 & 0.06 \\
SlimPajama & 300B & 0.14 & 0.02 & 0.05 & 0.00 & 0.05 \\
HPLT-2.0 & 300B & 0.17 & 0.02 & 0.00 & 0.00 & 0.05 \\
DCLM & 300B & 0.13 & 0.02 & 0.01 & 0.01 & 0.04 \\
Nemotron-CC-HQ & 300B & 0.09 & 0.03 & 0.02 & 0.00 & 0.03 \\
\multicolumn{7}{@{}l}{\textit{Models Trained with \texttt{open-sci-ref}} for 1T Tokens} \\
\midrule
FineWeb-Edu & 1T & 0.20 & 0.03 & 0.00 & 0.00 & 0.06 \\
Nemotron-CC-HQ & 1T & 0.13 & 0.03 & 0.01 & 0.04 & 0.05 \\
DCLM & 1T & 0.15 & 0.03 & 0.00 & 0.01 & 0.05 \\
\multicolumn{7}{@{}l}{\textit{Other Models}} \\
\midrule
SmolLM2-1.7B & 11T & 0.18 & 0.31 & 0.01 & 0.35 & 0.21 \\
SmolLM2-1.7B-Instruct & 11T & \textbf{0.28} & 0.37 & 0.28 & 0.37 & 0.33 \\
\bottomrule
\end{tabular}
\end{table}


The results (Table~\ref{tab:instruct_results}) show a dramatic difference on math (GSM8K) and coding (HumanEval, MBPP). \MixtureVitae{} achieves scores of \textbf{0.53}, \textbf{0.32}, and \textbf{0.38}, respectively. This performance is considerably stronger than any other dataset, all of which remain near random performance on GSM8K (0.02-0.06) and cap at 0.13 on HumanEval and 0.22 on MBPP. Most notably, our base model outperforms the post-trained SmolLM2-1.7B-Instruct \citep{lozhkovsmollm2} model on GSM8K, HumanEval, and MBPP --- despite the latter being trained on $\approx$11T tokens (over $36\times$ our budget).

\subsection{Test leakage and decontamination}
\label{sec:exp_contam}
To rule out test-set leakage as an alternative explanation for these gains, we perform a 13-gram exact-match decontamination sweep between \MixtureVitae{} and all benchmarks
(Appendix~\ref{app:contamination}). Document-level overlap is negligible for most tasks (e.g., at or below
$0.0003\%$ for ARC, HellaSwag, LAMBADA, OpenBookQA, and PIQA; see Table~\ref{tab:global_contam_summary}); contamination rates are modest for MMLU and BoolQ; for code benchmarks such as HumanEval and MBPP, contamination rates are higher but still small.

\textbf{Decontaminated Test Set Performance.} We re-evaluate all models on decontaminated test sets with all overlapping items removed. As shown in Table~\ref{tab:decontam-results-validation}, the performance of \MixtureVitae{} is consistent between the original and decontaminated versions. Crucially, the scores on GSM8K (0.54 decontaminated vs. 0.53 original) and MBPP (0.38 for both) remain stable, ruling out the possibility that our strong performance on math and coding is due to memorization of test items.

\begin{figure}[htbp]
    \centering
    \begin{subfigure}[t]{0.48\textwidth}
\centering        \includegraphics[width=0.95\linewidth]{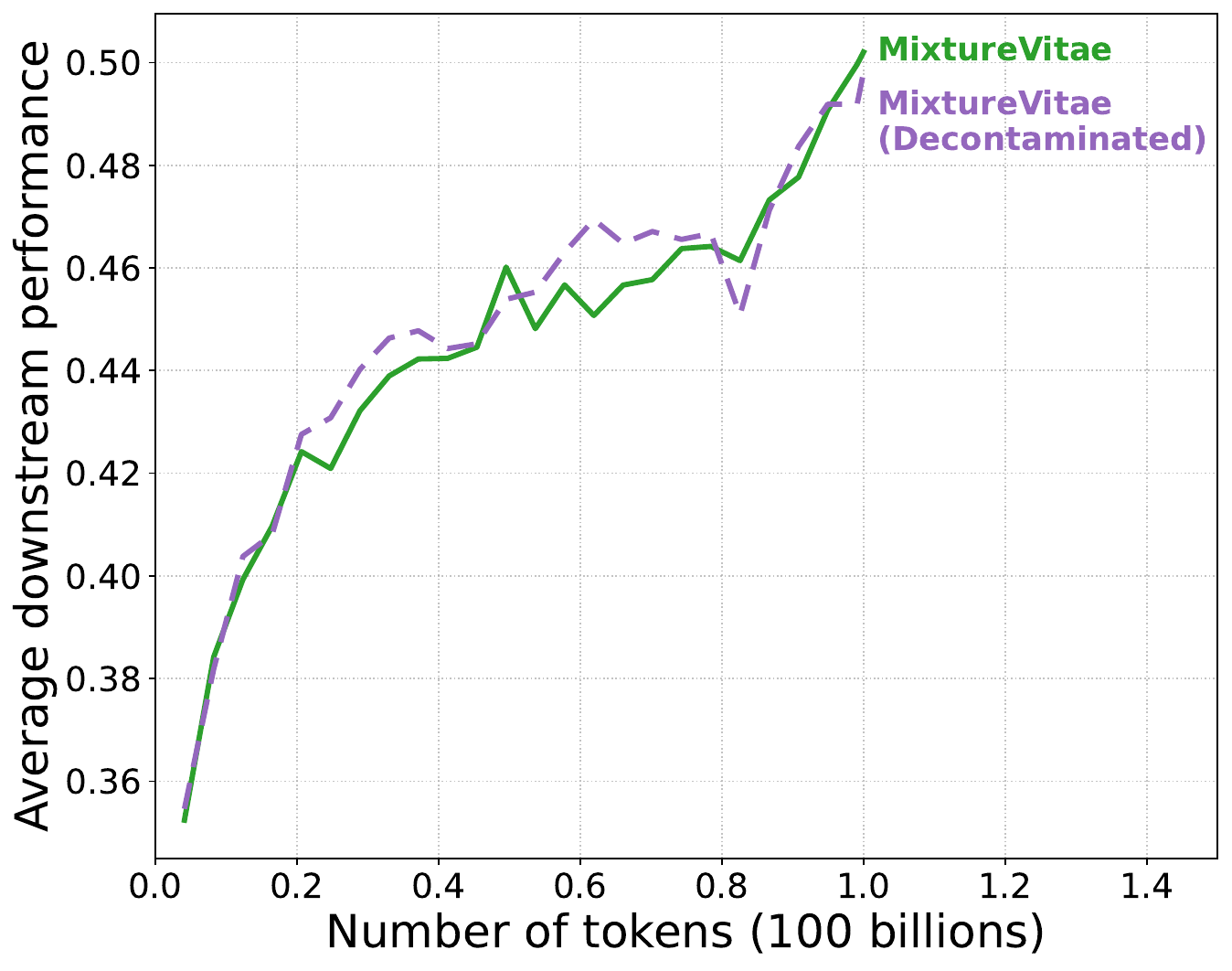}
\caption{Average accuracy across all tasks (as listed in Table~\ref{tab:eval_settings_general}) as a function of number of training steps.}
\end{subfigure}%
\hfill
     \begin{subfigure}[t]{0.48\textwidth}
        \centering        \includegraphics[width=0.95\linewidth]{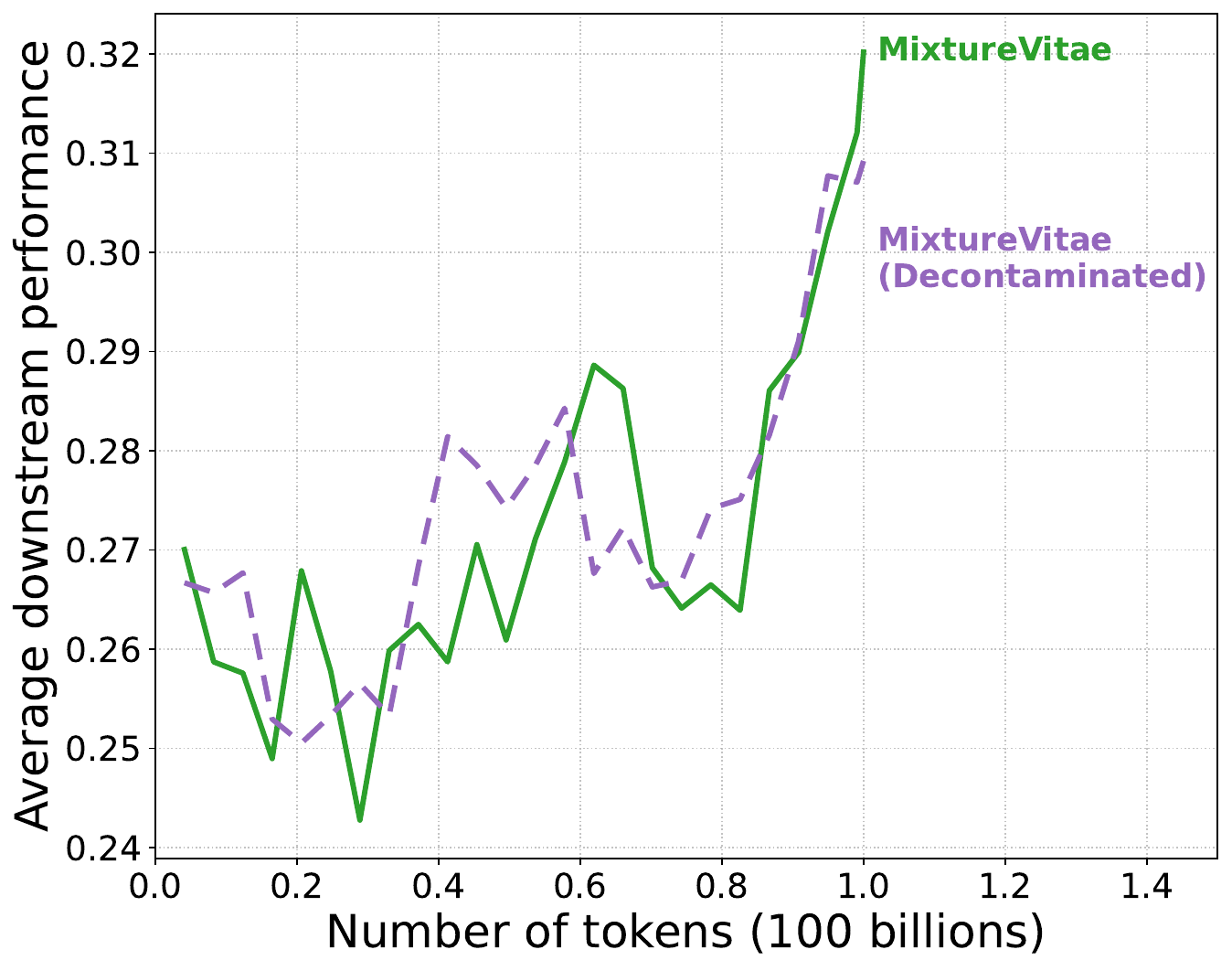}
        \caption{Accuracy on MMLU as a function of number of training steps.}
        \label{fig:mmlu_contam}
    \end{subfigure}
  \caption{\textbf{Validation of 1.7B model performance}. The \textbf{\MixtureVitae{} (Decontaminated)} model (purple, dashed), trained with dataset shards responsible for benchmark leakage removed, performs closely to the full \MixtureVitae{} (green, solid) model. This confirms our results are not an artifact of test set leakage.}
  \label{fig:ablate_app}
\end{figure}

\textbf{Retraining on Decontaminated Shards}. To further alleviate concerns, we train a 1.7B model, removing the shards responsible for the majority of the contamination signal. As illustrated in Figure~\ref{fig:ablate_app}, removing these shards had no negative effect on downstream performance. The training trajectory of the decontaminated model tracks closely with the full \MixtureVitae{} model, confirming that our results are not an artifact of dataset contamination.

\begin{table*}[htb]
\caption{Validating math, code, and instruction performance by comparing original (Orig) vs.\ decontaminated (Decont) test sets for 1.7B models trained for 300B tokens. \MixtureVitae{}'s high scores are shown to be genuine, as performance is maintained after removing all overlapping test items. This confirms the model's capabilities are not an artifact of test set leakage.}
\centering
\footnotesize
\begin{tabular}{l
                *{2}{c}
                *{2}{c}
                *{2}{c}
                *{2}{c}
                *{2}{c}}
\toprule
\multirow{2}{*}{\textbf{Training Dataset}} 
& \multicolumn{2}{c}{\textbf{GSM8K}} 
& \multicolumn{2}{c}{\textbf{GSM8K-CoT}} 
& \multicolumn{2}{c}{\textbf{MBPP}} 
& \multicolumn{2}{c}{\textbf{MBPP+}} 
& \multicolumn{2}{c}{\textbf{IFEval}} \\
\cmidrule(lr){2-3}\cmidrule(lr){4-5}\cmidrule(lr){6-7}\cmidrule(lr){8-9}\cmidrule(lr){10-11}
& \textbf{Orig} & \textbf{Decont} 
& \textbf{Orig} & \textbf{Decont} 
& \textbf{Orig} & \textbf{Decont} 
& \textbf{Orig} & \textbf{Decont} 
& \textbf{Orig} & \textbf{Decont} \\
\midrule
\textbf{\MixtureVitae{}}     & 0.53 & 0.54 & 0.50 & 0.50 & 0.38 & 0.38 & 0.55 & 0.59 & 0.19 & 0.23 \\
SmolLM2          & 0.30 & 0.30 & 0.28 & 0.29 & 0.35 & 0.35 & 0.48 & 0.48 & 0.17 & 0.20 \\
Comma-0.1        & 0.06 & 0.06 & 0.09 & 0.09 & 0.21 & 0.23 & 0.28 & 0.28 & 0.18 & 0.20 \\
CommonCorpus & 0.02 & 0.01 & 0.01 & 0.01 & 0.02 & 0.02 & 0.04 & 0.05 & 0.12 & 0.16 \\
C4               & 0.01 & 0.01 & 0.01 & 0.02 & 0.00 & 0.00 & 0.00 & 0.00 & 0.20 & 0.21 \\
DCLM             & 0.01 & 0.02 & 0.02 & 0.02 & 0.01 & 0.00 & 0.02 & 0.02 & 0.12 & 0.13 \\
FineWeb          & 0.02 & 0.01 & 0.03 & 0.03 & 0.00 & 0.00 & 0.00 & 0.00 & 0.18 & 0.20 \\
HPLT             & 0.02 & 0.02 & 0.02 & 0.02 & 0.00 & 0.00 & 0.00 & 0.00 & 0.17 & 0.21 \\
Nemotron-CC-HQ   & 0.03 & 0.02 & 0.03 & 0.03 & 0.00 & 0.00 & 0.00 & 0.00 & 0.09 & 0.10 \\
SlimPajama       & 0.02 & 0.02 & 0.02 & 0.02 & 0.00 & 0.00 & 0.00 & 0.00 & 0.14 & 0.15 \\
\bottomrule
\end{tabular}

\label{tab:decontam-results-validation}

\end{table*}

\subsection{Ablation Studies}
\label{app:ablations}

To isolate the impact of primary data components in \MixtureVitae{}, we define \textbf{Web} and \textbf{Instructions} subsets (see Figure~\ref{fig:composition_combined}; \textbf{Instructions} encompasses Reasoning \& Instruction and Math parts of the full mixture) and conduct an ablation study on a 100B-token scale. We train three separate models: (1) \textbf{\MixtureVitae{} (full)}, the complete dataset; (2) \textbf{\MixtureVitae{} (w/o Web)}, removing the \textbf{Web} component; (3) \textbf{\MixtureVitae{} (w/o Instructions)}, removing the \textbf{Instructions} component.

\begin{figure}[htbp]
    \centering
     \begin{subfigure}[c]{0.48\textwidth}
        \centering        \includegraphics[width=\linewidth]{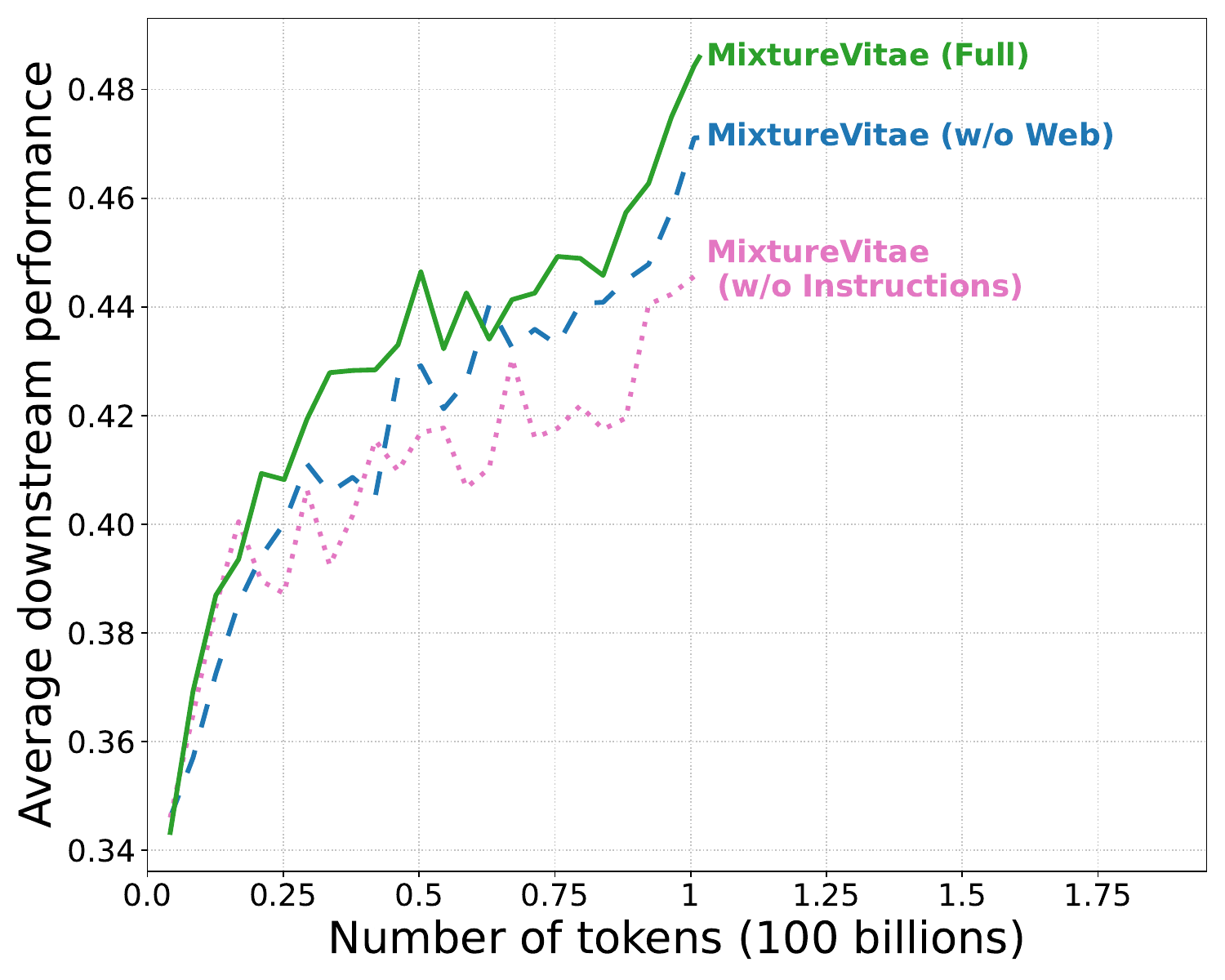}
          \caption{Ablation on full \MixtureVitae{} against two versions, each excluding a data subset as indicated by \texttt{w/o}. Average performance on 10 downstream tasks.}
        \label{fig:ablate}
    \end{subfigure}%
\begin{subfigure}[c]{0.48\textwidth}
\centering

\caption{A performance breakdown on math, coding and instruction following tasks for the ablated dataset variants. Best results are in \textbf{bold}. Numbers in \textcolor{red}{red} indicate strong performance drop.}

\resizebox{1\linewidth}{!}{
\begin{tabular}{lrrrr}
\toprule
Training Dataset  & IF-Eval & GSM8K & MBPP & Average \\
\midrule
\MixtureVitae{} & 0.14 & \textbf{0.47} & \textbf{0.34} & \textbf{0.25} \\
\shortstack{\MixtureVitae{}\\(w/o Web)} & 0.18 & 0.41 & 0.33 & \textbf{0.25} \\
\shortstack{\MixtureVitae{} \\ (w/o Instructions)} & \textbf{0.19} & \textcolor{red}{0.03} & \textcolor{red}{0.14} & 0.14 \\
\bottomrule
\end{tabular}}
\label{fig:ablate_table}
\end{subfigure} 
 \caption{An ablation study on components of the \MixtureVitae{} dataset. Fig. \ref{fig:ablate} shows performance average on 10 downstream evals during training, while Fig. \ref{fig:ablate_table} shows scores on further separate math, code and instruction benchmarks which are not part of the average in (a). The evaluation setup is given in Table~\ref{tab:reasoning-eval-settings}.}
\label{fig:ablate_all}
\end{figure}

The average downstream performance of these models (Figure~\ref{fig:ablate}) shows varying contributions by each component: The \textbf{Instructions} data is the most critical driver of performance, as its removal results in the largest, consistent drop of average performance compared to other configurations. Removing \textbf{Instructions} particularly leads to severe drop on GSM8k (from $0.47$ to $0.03$) and MBPP, as shown in Figure~\ref{fig:ablate_table}.  Absent the \textbf{Instructions} data, the model fails to match the gains of the full mix, underscoring the essential role of instruction-following data in generalization.

Removing the \textbf{Web} component (\textbf{w/o Web}, blue dashed line) also results in a performance drop below the full dataset, albeit less dramatically. Figure~\ref{fig:ablate_table} shows a drop from $0.47$ to $0.41$ on GSM8k, far less severe than the drop close to $0$ for \textbf{w/o Instructions} and only slight changes on code evals. The comparison of ablation effects again highlights the \textbf{crucial role of instruction and reasoning data in achieving high performance}.





\section{Related Work}
\label{sec:related_work}

LLM development is intrinsically linked to the scale and quality of pretraining datasets, which have become larger, more diverse, with a growing emphasis on provenance and licensing recently.

\subsection{Pioneering Large-Scale Datasets}
Early large-scale text corpora for language modeling often rely on web-crawled data for scale. C4~\citep{raffel2020exploring}, derived from Common Crawl, is instrumental in training the T5 model, setting standards for large-scale data cleaning and deduplication. \citet{gao2020pile} then introduce The Pile, demonstrating the benefit of a more varied data mixture on model generalization and downstream performance. Similarly, ROOTS~\citep{NEURIPS2022_ce9e92e3} supports the training of the BLOOM model with its 498 Common Crawl multilingual scrapes. While foundational, these datasets often have complex or unspecified licenses, mixing permissive data with content of unknown or non-commercial licensing, creating potential legal risks for commercial applications.

\subsection{Open and Reproducible Datasets}
Amidst many proprietary "black box" datasets, the community has pushed for more openness and reproducibility, moving toward permissive datasets that are also performant, e.g., RedPajama-1T~\citep{weber2024redpajama} and its processing recipes \citep{touvron2023llamaopenefficientfoundation}, Dolma~\citep{soldaini2024dolma} and its open-source curation toolkits, SILO~\citep{min2024silo}. Our work joins this effort, contributing a new risk-mitigated dataset featuring explicit consideration for the underlying copyright.

\subsection{Permissively Licensed and Synthetic Data}
Growing awareness of copyright and data ownership has spurred interest in datasets built solely from permissively licensed materials. The Stack \citep{kocetkov2023the} curates such data for code-generation models, but creating a large, diverse, and high-quality corpus for natural language from exclusively permissive sources remains a challenge. Recent efforts like Common Corpus~\citep{langlais2025commoncorpuslargestcollection} and The Common Pile~\citep{kandpal2025commonpilev018tb} advance the creation of large-scale corpora of permissively licensed and public-domain text. While foundational, our experiments (Section~\ref{sec:experiment_results}) show that models trained on them can lag in complex reasoning, math, and instruction following, suggesting that strictly permissive human text alone is insufficient to instill these advanced skills.

With this scarcity of high-quality reasoning and instruction data, researchers have turned to synthetic data. Alpaca~\citep{taori2023alpaca} and OpenMathInstruct-1 \citep{toshniwal2024openmathinstruct} use instructional data for fine-tuning. Phi4 proposes using synthetic data for reasoning tasks \citep{abdin2024phi4technicalreport}. Our work, \MixtureVitae{}, extends these trends with a meticulously curated, permissive-first, risk-mitigated dataset augmented with targeted synthetic data, providing a strong, legally considered foundation for LLM training to mitigate copyright risks in many existing corpora.

While both our work and the concurrent Apertus project~\citep{apertus2025apertusdemocratizingopencompliant} value openness and legal safety, they represent distinct, complementary design philosophies. 
First, regarding scale versus efficiency, Apertus optimizes for breadth, processing 15T tokens across 1800+ languages using retroactive filtering (e.g., \texttt{robots.txt}) on large web-scale datasets. 
In contrast, \MixtureVitae{} focuses on data efficiency through a \textit{positive inclusion strategy}, curating sources known to be permissive (e.g., government works, The Stack) and prioritizing English-centric reasoning density. 
Our results demonstrate that a reasoning-heavy mixture can achieve strong performance on MMLU, GSM8K, and MBPP with roughly 2\% of the pretraining token budget of a dataset in the size range of Apertus. 
Finally, whereas Apertus primarily releases recipes and reconstruction scripts, \MixtureVitae{} provides a single, ready-to-use pretraining dataset, which strongly simplifies reproducibility, validation and further experimentation by broad research community. 



\subsection{Mixing Reasoning Data into Pre-Training}

Concurrent with our work, \citet{akter2025front} systematically investigate the ``front-loading'' of reasoning data, finding that injecting reasoning data into the pretraining phase establishes foundational capabilities that cannot be replicated by scaling supervised fine-tuning (SFT) alone. They observe an asymmetric principle where pretraining benefits most from the scale and diversity of reasoning patterns, while SFT relies more heavily on data quality. Similarly, \cite{wang2025thinking} augment pre-training text data with synthetically generated thinking trajectories. They observe that pre-training augmented with thinking traces strongly outperforms vanilla pretraining using matched compute and token budget (8B model, 100BT) on reasoning/math/language understanding evals. Our findings with \MixtureVitae{} align with and extend this observation to the permissive dataset landscape: we show that by front-loading a diverse, risk-mitigated mixture of reasoning and instruction data, we can achieve competitive performance against non-permissive baselines even with a constrained token budget. For a dataset composition comparison of \MixtureVitae{} to other permissive and non-permissive baselines, see Tab.~\ref{tab:dataset_comparison}.




\section{Discussion \& Conclusion}

We have introduced \MixtureVitae{}, a pretraining corpus serving as a proof-of-concept: \textbf{Permissively licensed and permissively-sourced} real and synthetic data can achieve high performance. Our results suggest a shift in the \textbf{compliance--performance frontier}. \MixtureVitae{} demonstrates that  capabilities previously associated with mixed-license corpora are reachable with a permissive first, risk-mitigated approach. In our controlled 300B-token experiments, not only does \MixtureVitae{} catch up to leading non-permissive baselines like DCLM and FineWeb-Edu, but our 1.7B base model also outperforms the \textit{post-trained} SmolLM2-1.7B-Instruct---a model trained on $\approx$11T tokens---on GSM8K, HumanEval and MBPP. 

\textbf{Mixing dominant fraction of reasoning \& instruction data into pre-training.} \MixtureVitae{}’s performance is enhanced by the large proportion of reasoning and instruction data, as demonstrated in the ablation study in Section~\ref{app:ablations}. Removing this subset (“w/o Instructions” in Fig.~\ref{fig:ablate_all}) causes a substantial degradation across tasks—far larger than the impact of removing the web component. This observation validates and extends the findings of Phi-4~\citep{abdin2024phi4technicalreport}, showing that a permissive-first, risk-mitigated, and reasoning-heavy mixture can substitute vast quantities of generic web text, particularly under constrained token budgets. Importantly, while strongly boosting the performance on math/code tasks (Tab. \ref{tab:instruct_results}), language understanding evals also stays strong, matching non-permissive baselines and outperforming other permissive datasets (Fig.~\ref{fig:300b_performance}, Tab. \ref{tab:lang_results}). We thus provide evidence that heavily increasing reasoning and instruction data fraction on expense of generic web text creates overall boost in performance \textit{without} hurting core language understanding capabilities.

Beyond this specific corpus, the three‑tier licensing scheme and its shard‑level annotations
provide a \textbf{concrete template for structuring risk‑mitigated mixtures in future work}, and
\MixtureVitae{} as a whole serves as a reusable blueprint for compliant pretraining.
We demonstrate a fully open, reproducible pipeline built on positive‑inclusion “pseudo‑crawling,” tiered provenance
tracking, targeted synthetic generation with audited seeds and decontamination controlling for test set leakage. As detailed in our scaling outlook (Appendix~\ref{app:scaling_outlook}), this recipe provides a path to extend compliant pretraining to the multi-trillion-token regime---via subset upsampling, multilingual expansion, and synthetic growth---providing the community with a sustainable alternative to the legal uncertainty of broad web scrapes.
\section{Reproducibility statement}

We release our code at \url{https://github.com/ontocord/mixturevitae}.

\subsection{Dataset and Curation Recipes}

\begin{itemize}
    \item \textbf{Public Release:} The full \DatasetSize{}\textbf{B} token dataset, along with the 100B and 50B subsets used for scaling ablations experiments, will be made publicly available upon acceptance of this paper.
    \item \textbf{Curation Methodology:}
        \begin{itemize}
            \item \textbf{Dataset Composition} The detailed list of sources and their composition are shown in Figure~\ref{fig:data_comp_deets}.
            \item \textbf{Code}: We are including our  data curation and math word problem generation scripts with the submission.
        \end{itemize}
\end{itemize}

\subsection{Training Procedure}

To ensure our experiments are directly comparable and reproducible, we adhered to a controlled, public framework.
\begin{itemize}
    \item \textbf{Framework:} All experiments were conducted using the \textbf{open-sci-ref} training procedure~\citep{nezhurina2025opensciref001openreproduciblereference}, which standardizes key factors affecting performance.
    \item \textbf{Architectures:} The exact model architectures for all four scales (0.13B, 0.4B, 1.3B, 1.7B) are detailed in Table~\ref{tab:opensciref_scales}.
    \item \textbf{Hyperparameters:} The complete training schedules and hyperparameters (learning rate, batch size, warmup, etc.) for both the 50B and 300B token budgets are specified in Table~\ref{app:hyperparams}.
    \item \textbf{Software:} Models were trained using Megatron-LM~\citep{shoeybi2020megatronlmtrainingmultibillionparameter} with the GPT-NeoX-20B tokenizer\citep{black2022gptneox}.
    \item \textbf{Code}: We are including our training script with the submission.
\end{itemize}

\subsection{Evaluation and Analysis}

Our evaluation protocol is fully specified to allow for independent verification of our results.
\begin{itemize}
    \item \textbf{Framework:} All general and reasoning task evaluations were performed using the public LM Evaluation Harness~\citep{eval-harness}.
    \item \textbf{Settings:} The exact settings for each benchmark, including the number of few-shot examples, are provided in Table~\ref{tab:eval_settings_general} and Table~\ref{tab:reasoning-eval-settings}.
    \item \textbf{Decontamination:} Our 13-gram decontamination protocol is detailed in Appendix~\ref{app:contamination}.
    \item \textbf{Code}: We are including our evaluation and decontamination scripts with the submission.
\end{itemize}

While model checkpoints and training logs are not included in the initial submission due to size and anonymity constraints, we plan to release these upon publication to facilitate future research.

\section*{Acknowledgements}
\label{appendix:acknowledgements}

Huu Nguyen is thankful for and acknowledges discussions with Robert Kaczmarczyk on the ethical implications of training data; Colin Raffel on the inclusion of large scale instruction data in pretraining; Stella Biderman on various best practices for permissive datasets; Wojciech Kusa and the members of  NASK – National Research Institute on data safety; Veronika Laippala and Sampo Pyysalo of the University of Turku for their advice, and he espeically thanks his wife Thao Tran for her enduring support. 

Marianna Nezhurina, David Salinas and Jenia Jitsev acknowledge co-funding by EU from Digital Europe Programme under grant no. 101195233 (openEuroLLM). Marianna Nezhurina and Jenia Jitsev acknowledge co-funding from EuroHPC Joint Undertaking programm under grant no. 101182737 (MINERVA), funding by the Federal Ministry of Education and Research of Germany (BMBF) under grant no. 01IS24085C (OPENHAFM), under the grant 01IS22094B (WestAI - AI Service Center West), and under the grant 16HPC117K (MINERVA). 

We gratefully acknowledge the Gauss Centre for Supercomputing e.V. for funding this work by providing computing time through the John von Neumann Institute for Computing (NIC) on the supercomputer JUWELS Booster at Jülich Supercomputing Centre (JSC), EuroHPC Joint Undertaking for computing time and storage on the EuroHPC supercomputer LEONARDO, hosted by CINECA (Italy) and the LEONARDO consortium through an EuroHPC Extreme Access grant EHPC-EXT-2023E02-068 and through EuroHPC AI Factory Large Scale Access grant EHPC-AIF-2025LS01-028, storage resources on JUST granted and operated by JSC and supported by Helmholtz Data Federation (HDF), computing time granted by the JARA and JSC on the supercomputer JURECA at JSC.

We thank Robert Kaczmarczyk for coordination and support of EuroHPC Extreme Scale grant applications. Further thanks go for support provided by supercomputing facilities and their teams, especially to Damian Alvarez and Mathis Bode from Juelich Supercomputer Center (JSC, Germany) and to Laura Morselli from CINECA (Italy).

We also would like to thank all the members of the Ontocord\footnote{\url{https://huggingface.co/datasets/ontocord}}, LAION \footnote{\url{https://discord.gg/BZqhreFazY}}, and Open-$\Psi$ (open-sci) communities\footnote{\url{https://discord.gg/GsKh4mBVcv}} for providing a fertile ground for scientific exchange and open-source development.

\bibliography{main_bibliography, added}

@misc{akter2025front,
      title={Front-Loading Reasoning: The Synergy between Pretraining and Post-Training Data}, 
      author={Syeda Nahida Akter and Shrimai Prabhumoye and Eric Nyberg and Mostofa Patwary and Mohammad Shoeybi and Yejin Choi and Bryan Catanzaro},
      year={2025},
      eprint={2510.03264},
      archivePrefix={arXiv},
      primaryClass={cs.LG},
      url={https://arxiv.org/abs/2510.03264}, 
}

@article{wang2025thinking,
  title={Thinking Augmented Pre-training},
  author={Wang, Liang and Yang, Nan and Huang, Shaohan and Dong, Li and Wei, Furu},
  journal={arXiv preprint arXiv:2509.20186},
  year={2025}
}

@misc{eu_directive_tdm_2019,
  title = {{Directive (EU) 2019/790 of the European Parliament and of the Council of 17 April 2019 on copyright and related rights in the Digital Single Market}},
  author = {{European Union}},
  journal = {Official Journal of the European Union},
  year = {2019},
  note = {L 130/92}
}

@misc{PubMed,
  author       = {{National Library of Medicine (U.S.)}},
  title        = {PubMed},
  howpublished = {\url{https://pubmed.ncbi.nlm.nih.gov/}},
  year         = {1996}
}

@misc{uspto,
      title={{USPTO Patent Public Data Sets}},
      author={{United States Patent and Trademark Office}},
      year={2024},
      howpublished={\url{https://developer.uspto.gov/product/patent-public-data-sets}}
}

@misc{sec_edgar,
      title={{EDGAR: Electronic Data Gathering, Analysis, and Retrieval System}},
      author={{U.S. Securities and Exchange Commission}},
      year={2024},
      howpublished={\url{https://www.sec.gov/edgar}}
}

@article{lemley2017fair,
  title={Fair Learning},
  author={Lemley, Mark A and Casey, Bryan},
  journal={Texas Law Review},
  volume={95},
  pages={1},
  year={2017}
}

@misc{NVIDIAOpenScience2025,
  author = {{NVIDIA Corporation}},
  title = {{OpenScience Dataset}},
  year = {2025},
  publisher = {{Hugging Face}},
  url = {https://huggingface.co/datasets/nvidia/OpenScience}
}

@misc{openmanus_rl,
  author       = {{Ulab-UIUC and MetaGPT}},
  title        = {{OpenManus-RL Dataset}},
  year         = {2024},
  publisher    = {Hugging Face},
  journal      = {Hugging Face repository},
  howpublished = {\url{https://huggingface.co/datasets/CharlieDreemur/OpenManus-RL}}
}

@misc{nvidia2025prismmath,
  author       = {{NVIDIA}},
  title        = {{Nemotron-PrismMath Dataset}},
  year         = {2025},
  publisher    = {Hugging Face},
  journal      = {Hugging Face repository},
  howpublished = {\url{https://huggingface.co/datasets/nvidia/Nemotron-PrismMath}}
}

@techreport{peS2o,
    author = {Luca Soldaini and Kyle Lo},
    year = 2023,
    title = {{peS2o (Pretraining Efficiently on S2ORC) Dataset}},
    institution = {{Allen Institute for AI}},
    note = {ODC-By, \url{https://github.com/allenai/pes2o}}
}

@misc{glaive_reasoning_2023,
  author = {{Glaive AI}},
  title = {Glaive-AI Reasoning Dataset},
  year = {2023},
  publisher = {Hugging Face},
  journal = {Hugging Face repository},
  howpublished = {\url{https://huggingface.co/datasets/glaiveai/reasoning-v1-20m}}
}

@misc{nvidia_sft_datablend_v1,
  author       = {{NVIDIA}},
  title        = {{SFT DataBlend v1}},
  year         = {2024},
  publisher    = {Hugging Face},
  journal      = {Hugging Face repository},
  howpublished = {\url{https://huggingface.co/datasets/nvidia/sft_datablend_v1}}
}

@article{Langlais2024,
  author = {Langlais, Pierre-Carl},
  title = {Releasing Youtube-Commons: a massive open corpus for conversational and multimodal data},
  journal = {Hugging Face blog},
  year = {2024},
  month = {April},
  url = {https://huggingface.co/blog/Pclanglais/youtube-commons}
}

@misc{commoncrawl,
  author = {{Common Crawl Foundation}},
  title = {{Common Crawl}},
  year = {2025},
  howpublished = {\url{https://commoncrawl.org/}},
  note = {Accessed: 2025-08-25}
}

@misc{uscopyright,
  title        = {Copyright Act of 1976},
  year         = {1976},
  url          = {https://www.copyright.gov/title17/},
author = {United States Congress},
note = {Public Law 94-553, Enacted October 19, 1976}
}

@misc{Meta2024_Llama3,
   author = {Meta},
   journal = {Meta},
   title = {Introducing Meta Llama 3: The most capable openly available LLM to date},
   url = {https://ai.meta.com/blog/meta-llama-3/},
   lastchecked =  {22.05.2024},
   originalyear = {18.04.2024},
   year = {2024},
}

@inproceedings{hoffmann2022,
title={An empirical analysis of compute-optimal large language model training},
author={Jordan Hoffmann and Sebastian Borgeaud and Arthur Mensch and Elena Buchatskaya and Trevor Cai and Eliza Rutherford and Diego de las Casas and Lisa Anne Hendricks and Johannes Welbl and Aidan Clark and Tom Hennigan and Eric Noland and Katherine Millican and George van den Driessche and Bogdan Damoc and Aurelia Guy and Simon Osindero and Karen Simonyan and Erich Elsen and Oriol Vinyals and Jack William Rae and Laurent Sifre},
booktitle={Advances in Neural Information Processing Systems},
editor={Alice H. Oh and Alekh Agarwal and Danielle Belgrave and Kyunghyun Cho},
year={2022},
url={https://openreview.net/forum?id=iBBcRUlOAPR}
}

@article{brown2020language,
  title={Language models are few-shot learners},
  author={Brown, Tom and Mann, Benjamin and Ryder, Nick and Subbiah, Melanie and Kaplan, Jared D and Dhariwal, Prafulla and Neelakantan, Arvind and Shyam, Pranav and Sastry, Girish and Askell, Amanda and others},
  journal={Advances in neural information processing systems},
  volume={33},
  pages={1877--1901},
  year={2020}
}

@article{clark2018arc,
  title={Think you have solved question answering? try arc, the ai2 reasoning challenge},
  author={Clark, Peter and Cowhey, Isaac and Etzioni, Oren and Khot, Tushar and Sabharwal, Ashish and Schoenick, Carissa and Tafjord, Oyvind},
  journal={arXiv preprint arXiv:1803.05457},
  year={2018}
}

@inproceedings{bisk2019piqareasoningphysicalcommonsense,
  title={Piqa: Reasoning about physical commonsense in natural language},
  author={Bisk, Yonatan and Zellers, Rowan and Gao, Jianfeng and Choi, Yejin and others},
  booktitle={Proceedings of the AAAI conference on artificial intelligence},
  volume={34},
  pages={7432--7439},
  year={2020}
}

@article{cobbe2021training,
  title={Training verifiers to solve math word problems},
  author={Cobbe, Karl and Kosaraju, Vineet and Bavarian, Mohammad and Chen, Mark and Jun, Heewoo and Kaiser, Lukasz and Plappert, Matthias and Tworek, Jerry and Hilton, Jacob and Nakano, Reiichiro and others},
  journal={arXiv preprint arXiv:2110.14168},
  year={2021}
}

@inproceedings{zellers2019hellaswag,
   author = {Rowan Zellers and Ari Holtzman and Yonatan Bisk and Ali Farhadi and Yejin Choi},
   booktitle = {Proceedings of the 57th Annual Meeting of the Association for Computational Linguistics},
   title = {HellaSwag: Can a Machine Really Finish Your Sentence?},
   url = {https://arxiv.org/abs/1905.07830},
   year = {2019},
}

@inproceedings{
hendrycks2021mmlu,
title={Measuring Massive Multitask Language Understanding},
author={Dan Hendrycks and Collin Burns and Steven Basart and Andy Zou and Mantas Mazeika and Dawn Song and Jacob Steinhardt},
booktitle={International Conference on Learning Representations},
year={2021},
url={https://openreview.net/forum?id=d7KBjmI3GmQ}
}

@article{sakaguchi2020winogrande,
  title={Winogrande: An adversarial winograd schema challenge at scale},
  author={Sakaguchi, Keisuke and Bras, Ronan Le and Bhagavatula, Chandra and Choi, Yejin},
  journal={Communications of the ACM},
  volume={64},
  number={9},
  pages={99--106},
  year={2021},
  publisher={ACM New York, NY, USA}
}

@misc{eval-harness,
  author       = {Gao, Leo and
                  Tow, Jonathan and
                  Biderman, Stella and
                  Black, Sid and
                  DiPofi, Anthony and
                  Foster, Charles and
                  Golding, Laurence and
                  Hsu, Jeffrey and
                  McDonell, Kyle and
                  Muennighoff, Niklas and
                  Phang, Jason and
                  Reynolds, Laria and
                  Tang, Eric and
                  Thite, Anish and
                  Wang, Ben and
                  Wang, Kevin and
                  Zou, Andy},
  title        = {A framework for few-shot language model evaluation},
  month        = sep,
  year         = 2021,
  publisher    = {Zenodo},
  version      = {v0.0.1},
  doi          = {10.5281/zenodo.5371628},
  url          = {https://doi.org/10.5281/zenodo.5371628}
}

@misc{touvron2023llamaopenefficientfoundation,
      title={LLaMA: Open and Efficient Foundation Language Models}, 
      author={Hugo Touvron and Thibaut Lavril and Gautier Izacard and Xavier Martinet and Marie-Anne Lachaux and Timothée Lacroix and Baptiste Rozière and Naman Goyal and Eric Hambro and Faisal Azhar and Aurelien Rodriguez and Armand Joulin and Edouard Grave and Guillaume Lample},
      year={2023},
      eprint={2302.13971},
      archivePrefix={arXiv},
      primaryClass={cs.CL},
      url={https://arxiv.org/abs/2302.13971}, 
}

@misc{black2022gptneox,
      title={{GPT-NeoX-20B}: An Open-Source Autoregressive Language Model}, 
      author={Sid Black and Stella Biderman and Eric Hallahan and Quentin Anthony and Leo Gao and Laurence Golding and Horace He and Connor Leahy and Kyle McDonell and Jason Phang and Michael Pieler and USVSN Sai Prashanth and Shivanshu Purohit and Laria Reynolds and Jonathan Tow and Ben Wang and Samuel Weinbach},
      year={2022},
      eprint={2204.06745},
      archivePrefix={arXiv},
      primaryClass={cs.CL},
      url={https://arxiv.org/abs/2204.06745}, 
}

@misc{gao2020pile,
      title={The Pile: An 800GB Dataset of Diverse Text for Language Modeling}, 
      author={Leo Gao and Stella Biderman and Sid Black and Laurence Golding and Travis Hoppe and Charles Foster and Jason Phang and Horace He and Anish Thite and Noa Nabeshima and Shawn Presser and Connor Leahy},
      year={2020},
      eprint={2101.00027},
      archivePrefix={arXiv},
      primaryClass={cs.CL},
      url={https://arxiv.org/abs/2101.00027}, 
}

@misc{penedo2024,
      title={The FineWeb Datasets: Decanting the Web for the Finest Text Data at Scale}, 
      author={Guilherme Penedo and Hynek Kydlíček and Loubna Ben allal and Anton Lozhkov and Margaret Mitchell and Colin Raffel and Leandro Von Werra and Thomas Wolf},
      year={2024},
      eprint={2406.17557},
      archivePrefix={arXiv},
      primaryClass={cs.CL},
      url={https://arxiv.org/abs/2406.17557}, 
}

@misc{li2024datacomplm,
      title={DataComp-LM: In search of the next generation of training sets for language models}, 
      author={Jeffrey Li and Alex Fang and Georgios Smyrnis and Maor Ivgi and Matt Jordan and Samir Gadre and Hritik Bansal and Etash Guha and Sedrick Keh and Kushal Arora and Saurabh Garg and Rui Xin and Niklas Muennighoff and Reinhard Heckel and Jean Mercat and Mayee Chen and Suchin Gururangan and Mitchell Wortsman and Alon Albalak and Yonatan Bitton and Marianna Nezhurina and Amro Abbas and Cheng-Yu Hsieh and Dhruba Ghosh and Josh Gardner and Maciej Kilian and Hanlin Zhang and Rulin Shao and Sarah Pratt and Sunny Sanyal and Gabriel Ilharco and Giannis Daras and Kalyani Marathe and Aaron Gokaslan and Jieyu Zhang and Khyathi Chandu and Thao Nguyen and Igor Vasiljevic and Sham Kakade and Shuran Song and Sujay Sanghavi and Fartash Faghri and Sewoong Oh and Luke Zettlemoyer and Kyle Lo and Alaaeldin El-Nouby and Hadi Pouransari and Alexander Toshev and Stephanie Wang and Dirk Groeneveld and Luca Soldaini and Pang Wei Koh and Jenia Jitsev and Thomas Kollar and Alexandros G. Dimakis and Yair Carmon and Achal Dave and Ludwig Schmidt and Vaishaal Shankar},
      year={2025},
      eprint={2406.11794},
      archivePrefix={arXiv},
      primaryClass={cs.LG},
      url={https://arxiv.org/abs/2406.11794}, 
}

@inproceedings{su2025nemotroncc,
  title={Nemotron-cc: Transforming common crawl into a refined long-horizon pretraining dataset},
  author={Su, Dan and Kong, Kezhi and Lin, Ying and Jennings, Joseph and Norick, Brandon and Kliegl, Markus and Patwary, Mostofa and Shoeybi, Mohammad and Catanzaro, Bryan},
  booktitle={Proceedings of the 63rd Annual Meeting of the Association for Computational Linguistics (Volume 1: Long Papers)},
  pages={2459--2475},
  year={2025}
}

@inproceedings{burchell2025,
  title={An expanded massive multilingual dataset for high-performance language technologies (HPLT)},
  author={Burchell, Laurie and Bonet, Ona De Gibert and Arefyev, Nikolay and Aulamo, Mikko and Ba{\~n}{\'o}n, Marta and Chen, Pinzhen and Fedorova, Mariia and Guillou, Liane and Haddow, Barry and Hajic, Jan and others},
  booktitle={Proceedings of the 63rd Annual Meeting of the Association for Computational Linguistics (Volume 1: Long Papers)},
  pages={17452--17485},
  year={2025}
}

@misc{langlais2025commoncorpuslargestcollection,
      title={Common Corpus: The Largest Collection of Ethical Data for LLM Pre-Training}, 
      author={Pierre-Carl Langlais and Carlos Rosas Hinostroza and Mattia Nee and Catherine Arnett and Pavel Chizhov and Eliot Krzystof Jones and Irène Girard and David Mach and Anastasia Stasenko and Ivan P. Yamshchikov},
      year={2025},
      eprint={2506.01732},
      archivePrefix={arXiv},
      primaryClass={cs.CL},
      url={https://arxiv.org/abs/2506.01732}, 
}

@inproceedings{talmor2019commonsenseqa,
    title = "{C}ommonsense{QA}: A Question Answering Challenge Targeting Commonsense Knowledge",
    author = "Talmor, Alon  and
      Herzig, Jonathan  and
      Lourie, Nicholas  and
      Berant, Jonathan",
    editor = "Burstein, Jill  and
      Doran, Christy  and
      Solorio, Thamar",
    booktitle = "Proceedings of the 2019 Conference of the North {A}merican Chapter of the Association for Computational Linguistics: Human Language Technologies, Volume 1 (Long and Short Papers)",
    month = jun,
    year = "2019",
    address = "Minneapolis, Minnesota",
    publisher = "Association for Computational Linguistics",
    url = "https://aclanthology.org/N19-1421/",
    doi = "10.18653/v1/N19-1421",
    pages = "4149--4158",
}

@inproceedings{paperno2016lambada,
    title = "The {LAMBADA} dataset: Word prediction requiring a broad discourse context",
    author = "Paperno, Denis  and
      Kruszewski, Germ{\'a}n  and
      Lazaridou, Angeliki  and
      Pham, Ngoc Quan  and
      Bernardi, Raffaella  and
      Pezzelle, Sandro  and
      Baroni, Marco  and
      Boleda, Gemma  and
      Fern{\'a}ndez, Raquel",
    editor = "Erk, Katrin  and
      Smith, Noah A.",
    booktitle = "Proceedings of the 54th Annual Meeting of the Association for Computational Linguistics (Volume 1: Long Papers)",
    month = aug,
    year = "2016",
    address = "Berlin, Germany",
    publisher = "Association for Computational Linguistics",
    url = "https://aclanthology.org/P16-1144/",
    doi = "10.18653/v1/P16-1144",
    pages = "1525--1534"
}

@inproceedings{copa,
author = {Roemmele, Melissa and Bejan, Cosmin and Gordon, Andrew},
year = {2011},
title = {Choice of Plausible Alternatives: An Evaluation of Commonsense Causal Reasoning.},
booktitle = {AAAI spring symposium: logical formalizations of commonsense reasoning},
pages = "90--95"
}

@inproceedings{clark2019boolq,
    title = "{B}ool{Q}: Exploring the Surprising Difficulty of Natural Yes/No Questions",
    author = "Clark, Christopher  and
      Lee, Kenton  and
      Chang, Ming-Wei  and
      Kwiatkowski, Tom  and
      Collins, Michael  and
      Toutanova, Kristina",
    editor = "Burstein, Jill  and
      Doran, Christy  and
      Solorio, Thamar",
    booktitle = "Proceedings of the 2019 Conference of the North {A}merican Chapter of the Association for Computational Linguistics: Human Language Technologies, Volume 1 (Long and Short Papers)",
    month = jun,
    year = "2019",
    address = "Minneapolis, Minnesota",
    publisher = "Association for Computational Linguistics",
    url = "https://aclanthology.org/N19-1300/",
    doi = "10.18653/v1/N19-1300",
    pages = "2924--2936",
}

@inproceedings{mihaylov2018openbookqa,
  title={Can a Suit of Armor Conduct Electricity? A New Dataset for Open Book Question Answering},
  author={Mihaylov, Todor and Clark, Peter and Khot, Tushar and Sabharwal, Ashish},
  booktitle={Proceedings of the 2018 Conference on Empirical Methods in Natural Language Processing},
  pages={2381--2391},
  year={2018}
}

@misc{kandpal2025commonpilev018tb,
      title={The Common Pile v0.1: An 8TB Dataset of Public Domain and Openly Licensed Text}, 
      author={Nikhil Kandpal and Brian Lester and Colin Raffel and Sebastian Majstorovic and Stella Biderman and Baber Abbasi and Luca Soldaini and Enrico Shippole and A. Feder Cooper and Aviya Skowron and John Kirchenbauer and Shayne Longpre and Lintang Sutawika and Alon Albalak and Zhenlin Xu and Guilherme Penedo and Loubna Ben Allal and Elie Bakouch and John David Pressman and Honglu Fan and Dashiell Stander and Guangyu Song and Aaron Gokaslan and Tom Goldstein and Brian R. Bartoldson and Bhavya Kailkhura and Tyler Murray},
      year={2025},
      eprint={2506.05209},
      archivePrefix={arXiv},
      primaryClass={cs.CL},
      url={https://arxiv.org/abs/2506.05209}, 
}

@misc{abdin2024phi4technicalreport,
      title={Phi-4 Technical Report}, 
      author={Marah Abdin and Jyoti Aneja and Harkirat Behl and Sébastien Bubeck and Ronen Eldan and Suriya Gunasekar and Michael Harrison and Russell J. Hewett and Mojan Javaheripi and Piero Kauffmann and James R. Lee and Yin Tat Lee and Yuanzhi Li and Weishung Liu and Caio C. T. Mendes and Anh Nguyen and Eric Price and Gustavo de Rosa and Olli Saarikivi and Adil Salim and Shital Shah and Xin Wang and Rachel Ward and Yue Wu and Dingli Yu and Cyril Zhang and Yi Zhang},
      year={2024},
      eprint={2412.08905},
      archivePrefix={arXiv},
      primaryClass={cs.CL},
      url={https://arxiv.org/abs/2412.08905}, 
}

@misc{
cui2024ultrafeedback,
title={UltraFeedback: Boosting Language Models with High-quality Feedback},
author={Ganqu Cui and Lifan Yuan and Ning Ding and Guanming Yao and Wei Zhu and Yuan Ni and Guotong Xie and Zhiyuan Liu and Maosong Sun},
year={2024},
url={https://openreview.net/forum?id=pNkOx3IVWI}
}

@misc{xu2024magpiealignmentdatasynthesis,
      title={Magpie: Alignment Data Synthesis from Scratch by Prompting Aligned LLMs with Nothing}, 
      author={Zhangchen Xu and Fengqing Jiang and Luyao Niu and Yuntian Deng and Radha Poovendran and Yejin Choi and Bill Yuchen Lin},
      year={2024},
      eprint={2406.08464},
      archivePrefix={arXiv},
      primaryClass={cs.CL},
      url={https://arxiv.org/abs/2406.08464}, 
}

@article{köpf2023openassistant,
  title={Openassistant conversations-democratizing large language model alignment},
  author={K{\"o}pf, Andreas and Kilcher, Yannic and Von R{\"u}tte, Dimitri and Anagnostidis, Sotiris and Tam, Zhi Rui and Stevens, Keith and Barhoum, Abdullah and Nguyen, Duc and Stanley, Oliver and Nagyfi, Rich{\'a}rd and others},
  journal={Advances in neural information processing systems},
  volume={36},
  pages={47669--47681},
  year={2023}
}

@misc{chung2022scalinginstructionfinetunedlanguagemodels,
      title={Scaling Instruction-Finetuned Language Models}, 
      author={Hyung Won Chung and Le Hou and Shayne Longpre and Barret Zoph and Yi Tay and William Fedus and Yunxuan Li and Xuezhi Wang and Mostafa Dehghani and Siddhartha Brahma and Albert Webson and Shixiang Shane Gu and Zhuyun Dai and Mirac Suzgun and Xinyun Chen and Aakanksha Chowdhery and Alex Castro-Ros and Marie Pellat and Kevin Robinson and Dasha Valter and Sharan Narang and Gaurav Mishra and Adams Yu and Vincent Zhao and Yanping Huang and Andrew Dai and Hongkun Yu and Slav Petrov and Ed H. Chi and Jeff Dean and Jacob Devlin and Adam Roberts and Denny Zhou and Quoc V. Le and Jason Wei},
      year={2022},
      eprint={2210.11416},
      archivePrefix={arXiv},
      primaryClass={cs.LG},
      url={https://arxiv.org/abs/2210.11416}, 
}

@inproceedings{sanh2021multitask,
      title={Multitask Prompted Training Enables Zero-Shot Task Generalization},
      author={Victor Sanh and Albert Webson and Colin Raffel and Stephen H. Bach and Lintang Sutawika and Zaid Alyafeai and Antoine Chaffin and Arnaud Stiegler and Teven Le Scao and Arun Raja and Manan Dey and M Saiful Bari and Canwen Xu and Urmish Thakker and Shanya Sharma Sharma and Eliza Szczechla and Taewoon Kim and Gunjan Chhablani and Nihal Nayak and Debajyoti Datta and Jonathan Chang and Mike Tian-Jian Jiang and Han Wang and Matteo Manica and Sheng Shen and Zheng Xin Yong and Harshit Pandey and Rachel Bawden and Thomas Wang and Trishala Neeraj and Jos Rozen and Abheesht Sharma and Andrea Santilli and Thibault Fevry and Jason Alan Fries and Ryan Teehan and Stella Biderman and Leo Gao and Tali Bers and Thomas Wolf and Alexander M. Rush},
      year={2022},
      booktitle={International Conference on Learning Representations},
url={https://openreview.net/forum?id=9Vrb9D0WI4}
}

@inproceedings{yu2024metamathbootstrapmathematicalquestions,
  title={MetaMath: Bootstrap Your Own Mathematical Questions for Large Language Models},
  author={Yu, Longhui and Jiang, Weisen and Shi, Han and YU, Jincheng and Liu, Zhengying and Zhang, Yu and Kwok, James and Li, Zhenguo and Weller, Adrian and Liu, Weiyang},
  booktitle={The Twelfth International Conference on Learning Representations},
year={2024}
}

@misc{toshniwal2024openmathinstruct118millionmath,
      title={OpenMathInstruct-1: A 1.8 Million Math Instruction Tuning Dataset}, 
      author={Shubham Toshniwal and Ivan Moshkov and Sean Narenthiran and Daria Gitman and Fei Jia and Igor Gitman},
      year={2024},
      eprint={2402.10176},
      archivePrefix={arXiv},
      primaryClass={cs.CL},
      url={https://arxiv.org/abs/2402.10176}, 
}

@article{guha2025openthoughts,
  title={OpenThoughts: Data Recipes for Reasoning Models},
  author={Guha, Etash and Marten, Ryan and Keh, Sedrick and Raoof, Negin and Smyrnis, Georgios and Bansal, Hritik and Nezhurina, Marianna and Mercat, Jean and Vu, Trung and Sprague, Zayne and others},
  journal={arXiv preprint arXiv:2506.04178},
  year={2025}
}

@inproceedings{NEURIPS2022_ce9e92e3,
 author = {Lauren\c{c}on, Hugo and Saulnier, Lucile and Wang, Thomas and Akiki, Christopher and Villanova del Moral, Albert and Le Scao, Teven and Von Werra, Leandro and Mou, Chenghao and Gonz\'{a}lez Ponferrada, Eduardo and Nguyen, Huu and Frohberg, J\"{o}rg and \v{S}a\v{s}ko, Mario and Lhoest, Quentin and McMillan-Major, Angelina and Dupont, Gerard and Biderman, Stella and Rogers, Anna and Ben allal, Loubna and De Toni, Francesco and Pistilli, Giada and Nguyen, Olivier and Nikpoor, Somaieh and Masoud, Maraim and Colombo, Pierre and de la Rosa, Javier and Villegas, Paulo and Thrush, Tristan and Longpre, Shayne and Nagel, Sebastian and Weber, Leon and Mu\~{n}oz, Manuel and Zhu, Jian and Van Strien, Daniel and Alyafeai, Zaid and Almubarak, Khalid and Vu, Minh Chien and Gonzalez-Dios, Itziar and Soroa, Aitor and Lo, Kyle and Dey, Manan and Ortiz Suarez, Pedro and Gokaslan, Aaron and Bose, Shamik and Adelani, David and Phan, Long and Tran, Hieu and Yu, Ian and Pai, Suhas and Chim, Jenny and Lepercq, Violette and Ilic, Suzana and Mitchell, Margaret and Luccioni, Sasha Alexandra and Jernite, Yacine},
 booktitle = {Advances in Neural Information Processing Systems},
 editor = {S. Koyejo and S. Mohamed and A. Agarwal and D. Belgrave and K. Cho and A. Oh},
 pages = {31809--31826},
 publisher = {Curran Associates, Inc.},
 title = {The BigScience ROOTS Corpus: A 1.6TB Composite Multilingual Dataset},
 url = {https://proceedings.neurips.cc/paper_files/paper/2022/file/ce9e92e3de2372a4b93353eb7f3dc0bd-Paper-Datasets_and_Benchmarks.pdf},
 volume = {35},
 year = {2022}
}

@inproceedings{nakamura2024auroramopensourcecontinual,
  title={Aurora-m: Open source continual pre-training for multilingual language and code},
  author={Nakamura, Taishi and Mishra, Mayank and Tedeschi, Simone and Chai, Yekun and Stillerman, Jason T and Friedrich, Felix and Yadav, Prateek and Laud, Tanmay and Chien, Vu Minh and Zhuo, Terry Yue and others},
  booktitle={Proceedings of the 31st International Conference on Computational Linguistics: Industry Track},
  pages={656--678},
  year={2025}
}

@misc{shen2024slimpajamadcunderstandingdatacombinations,
      title={SlimPajama-DC: Understanding Data Combinations for LLM Training}, 
      author={Zhiqiang Shen and Tianhua Tao and Liqun Ma and Willie Neiswanger and Zhengzhong Liu and Hongyi Wang and Bowen Tan and Joel Hestness and Natalia Vassilieva and Daria Soboleva and Eric Xing},
      year={2024},
      eprint={2309.10818},
      archivePrefix={arXiv},
      primaryClass={cs.CL},
      url={https://arxiv.org/abs/2309.10818}, 
}

@inproceedings{
albalak2023efficientonlinedatamixing,
title={Efficient Online Data Mixing For Language Model Pre-Training},
author={Alon Albalak and Liangming Pan and Colin Raffel and William Yang Wang},
booktitle={R0-FoMo:Robustness of Few-shot and Zero-shot Learning in Large Foundation Models},
year={2023},
url={https://openreview.net/forum?id=9Tze4oy4lw}
}

@inproceedings{xie2023doremioptimizingdatamixtures,
  title={Doremi: Optimizing data mixtures speeds up language model pretraining},
  author={Xie, Sang Michael and Pham, Hieu and Dong, Xuanyi and Du, Nan and Liu, Hanxiao and Lu, Yifeng and Liang, Percy S and Le, Quoc V and Ma, Tengyu and Yu, Adams Wei},
  booktitle={Advances in Neural Information Processing Systems},
  volume={36},
  pages={69798--69818},
  year={2023}
}

@misc{shoeybi2020megatronlmtrainingmultibillionparameter,
      title={Megatron-LM: Training Multi-Billion Parameter Language Models Using Model Parallelism}, 
      author={Mohammad Shoeybi and Mostofa Patwary and Raul Puri and Patrick LeGresley and Jared Casper and Bryan Catanzaro},
      year={2020},
      eprint={1909.08053},
      archivePrefix={arXiv},
      primaryClass={cs.CL},
      url={https://arxiv.org/abs/1909.08053}, 
}

@misc{nezhurina2025opensciref001openreproduciblereference,
      title={Open-sci-ref-0.01: open and reproducible reference baselines for language model and dataset comparison}, 
      author={Marianna Nezhurina and Jörg Franke and Taishi Nakamura and Timur Carstensen and Niccolò Ajroldi and Ville Komulainen and David Salinas and Jenia Jitsev},
      year={2025},
      eprint={2509.09009},
      archivePrefix={arXiv},
      primaryClass={cs.LG},
      url={https://arxiv.org/abs/2509.09009}, 
}

@misc{bercovich2025llamanemotronefficientreasoningmodels,
      title={Llama-Nemotron: Efficient Reasoning Models}, 
      author={Akhiad Bercovich and Itay Levy and Izik Golan and Mohammad Dabbah and Ran El-Yaniv and Omri Puny and Ido Galil and Zach Moshe and Tomer Ronen and Najeeb Nabwani and Ido Shahaf and Oren Tropp and Ehud Karpas and Ran Zilberstein and Jiaqi Zeng and Soumye Singhal and Alexander Bukharin and Yian Zhang and Tugrul Konuk and Gerald Shen and Ameya Sunil Mahabaleshwarkar and Bilal Kartal and Yoshi Suhara and Olivier Delalleau and Zijia Chen and Zhilin Wang and David Mosallanezhad and Adi Renduchintala and Haifeng Qian and Dima Rekesh and Fei Jia and Somshubra Majumdar and Vahid Noroozi and Wasi Uddin Ahmad and Sean Narenthiran and Aleksander Ficek and Mehrzad Samadi and Jocelyn Huang and Siddhartha Jain and Igor Gitman and Ivan Moshkov and Wei Du and Shubham Toshniwal and George Armstrong and Branislav Kisacanin and Matvei Novikov and Daria Gitman and Evelina Bakhturina and Prasoon Varshney and Makesh Narsimhan and Jane Polak Scowcroft and John Kamalu and Dan Su and Kezhi Kong and Markus Kliegl and Rabeeh Karimi Mahabadi and Ying Lin and Sanjeev Satheesh and Jupinder Parmar and Pritam Gundecha and Brandon Norick and Joseph Jennings and Shrimai Prabhumoye and Syeda Nahida Akter and Mostofa Patwary and Abhinav Khattar and Deepak Narayanan and Roger Waleffe and Jimmy Zhang and Bor-Yiing Su and Guyue Huang and Terry Kong and Parth Chadha and Sahil Jain and Christine Harvey and Elad Segal and Jining Huang and Sergey Kashirsky and Robert McQueen and Izzy Putterman and George Lam and Arun Venkatesan and Sherry Wu and Vinh Nguyen and Manoj Kilaru and Andrew Wang and Anna Warno and Abhilash Somasamudramath and Sandip Bhaskar and Maka Dong and Nave Assaf and Shahar Mor and Omer Ullman Argov and Scot Junkin and Oleksandr Romanenko and Pedro Larroy and Monika Katariya and Marco Rovinelli and Viji Balas and Nicholas Edelman and Anahita Bhiwandiwalla and Muthu Subramaniam and Smita Ithape and Karthik Ramamoorthy and Yuting Wu and Suguna Varshini Velury and Omri Almog and Joyjit Daw and Denys Fridman and Erick Galinkin and Michael Evans and Shaona Ghosh and Katherine Luna and Leon Derczynski and Nikki Pope and Eileen Long and Seth Schneider and Guillermo Siman and Tomasz Grzegorzek and Pablo Ribalta and Monika Katariya and Chris Alexiuk and Joey Conway and Trisha Saar and Ann Guan and Krzysztof Pawelec and Shyamala Prayaga and Oleksii Kuchaiev and Boris Ginsburg and Oluwatobi Olabiyi and Kari Briski and Jonathan Cohen and Bryan Catanzaro and Jonah Alben and Yonatan Geifman and Eric Chung},
      year={2025},
      eprint={2505.00949},
      archivePrefix={arXiv},
      primaryClass={cs.CL},
      url={https://arxiv.org/abs/2505.00949}, 
}

@article{kocetkov2023the,
  title={The Stack: 3 TB of permissively licensed source code},
  author={Kocetkov, Denis and Li, Raymond and Mou, Chenghao and Jernite, Yacine and Mitchell, Margaret and Ferrandis, Carlos Mu{\~n}oz and Hughes, Sean and Wolf, Thomas and Bahdanau, Dzmitry and Von Werra, Leandro and others},
  journal={Transactions on Machine Learning Research},
year={2023}
}

@article{tong2024dartmath,
  title={Dart-math: Difficulty-aware rejection tuning for mathematical problem-solving},
  author={Tong, Yuxuan and Zhang, Xiwen and Wang, Rui and Wu, Ruidong and He, Junxian},
  journal={Advances in Neural Information Processing Systems},
  volume={37},
  pages={7821--7846},
  year={2024}
}

@misc{bommarito2025kl3mdataprojectcopyrightclean,
      title={The KL3M Data Project: Copyright-Clean Training Resources for Large Language Models}, 
      author={Bommarito, II, Michael J. and Bommarito, Jillian and Katz, Daniel Martin},
      year={2025},
      eprint={2504.07854},
      archivePrefix={arXiv},
      primaryClass={cs.CL},
      url={https://arxiv.org/abs/2504.07854}, 
}

@inproceedings{
min2024silo,
title={{SILO} Language Models: Isolating Legal Risk In a Nonparametric Datastore},
author={Sewon Min and Suchin Gururangan and Eric Wallace and Weijia Shi and Hannaneh Hajishirzi and Noah A. Smith and Luke Zettlemoyer},
booktitle={The Twelfth International Conference on Learning Representations},
year={2024},
url={https://openreview.net/forum?id=ruk0nyQPec}
}

@article{zhou2023instruction,
  title={Instruction-following evaluation for large language models},
  author={Zhou, Jeffrey and Lu, Tianjian and Mishra, Swaroop and Brahma, Siddhartha and Basu, Sujoy and Luan, Yi and Zhou, Denny and Hou, Le},
  journal={arXiv preprint arXiv:2311.07911},
  year={2023}
}

@inproceedings{wang2024not,
  title={Do-not-answer: Evaluating safeguards in LLMs},
  author={Wang, Yuxia and Li, Haonan and Han, Xudong and Nakov, Preslav and Baldwin, Timothy},
  booktitle={Findings of the Association for Computational Linguistics: EACL 2024},
  pages={896--911},
  year={2024}
}

@misc{inan2023llamaguardllmbasedinputoutput,
      title={Llama Guard: LLM-based Input-Output Safeguard for Human-AI Conversations}, 
      author={Hakan Inan and Kartikeya Upasani and Jianfeng Chi and Rashi Rungta and Krithika Iyer and Yuning Mao and Michael Tontchev and Qing Hu and Brian Fuller and Davide Testuggine and Madian Khabsa},
      year={2023},
      eprint={2312.06674},
      archivePrefix={arXiv},
      primaryClass={cs.CL},
      url={https://arxiv.org/abs/2312.06674}, 
}

@misc{rae2022scalinglanguagemodelsmethods,
      title={Scaling Language Models: Methods, Analysis \& Insights from Training Gopher}, 
      author={Jack W. Rae and Sebastian Borgeaud and Trevor Cai and Katie Millican and Jordan Hoffmann and Francis Song and John Aslanides and Sarah Henderson and Roman Ring and Susannah Young and Eliza Rutherford and Tom Hennigan and Jacob Menick and Albin Cassirer and Richard Powell and George van den Driessche and Lisa Anne Hendricks and Maribeth Rauh and Po-Sen Huang and Amelia Glaese and Johannes Welbl and Sumanth Dathathri and Saffron Huang and Jonathan Uesato and John Mellor and Irina Higgins and Antonia Creswell and Nat McAleese and Amy Wu and Erich Elsen and Siddhant Jayakumar and Elena Buchatskaya and David Budden and Esme Sutherland and Karen Simonyan and Michela Paganini and Laurent Sifre and Lena Martens and Xiang Lorraine Li and Adhiguna Kuncoro and Aida Nematzadeh and Elena Gribovskaya and Domenic Donato and Angeliki Lazaridou and Arthur Mensch and Jean-Baptiste Lespiau and Maria Tsimpoukelli and Nikolai Grigorev and Doug Fritz and Thibault Sottiaux and Mantas Pajarskas and Toby Pohlen and Zhitao Gong and Daniel Toyama and Cyprien de Masson d'Autume and Yujia Li and Tayfun Terzi and Vladimir Mikulik and Igor Babuschkin and Aidan Clark and Diego de Las Casas and Aurelia Guy and Chris Jones and James Bradbury and Matthew Johnson and Blake Hechtman and Laura Weidinger and Iason Gabriel and William Isaac and Ed Lockhart and Simon Osindero and Laura Rimell and Chris Dyer and Oriol Vinyals and Kareem Ayoub and Jeff Stanway and Lorrayne Bennett and Demis Hassabis and Koray Kavukcuoglu and Geoffrey Irving},
      year={2022},
      eprint={2112.11446},
      archivePrefix={arXiv},
      primaryClass={cs.CL},
      url={https://arxiv.org/abs/2112.11446}, 
}

@inproceedings{hartvigsen2022toxigen,
    title = "{T}oxi{G}en: A Large-Scale Machine-Generated Dataset for Adversarial and Implicit Hate Speech Detection",
    author = "Hartvigsen, Thomas  and
      Gabriel, Saadia  and
      Palangi, Hamid  and
      Sap, Maarten  and
      Ray, Dipankar  and
      Kamar, Ece",
    editor = "Muresan, Smaranda  and
      Nakov, Preslav  and
      Villavicencio, Aline",
    booktitle = "Proceedings of the 60th Annual Meeting of the Association for Computational Linguistics (Volume 1: Long Papers)",
    month = may,
    year = "2022",
    address = "Dublin, Ireland",
    publisher = "Association for Computational Linguistics",
    url = "https://aclanthology.org/2022.acl-long.234/",
    doi = "10.18653/v1/2022.acl-long.234",
    pages = "3309--3326"
}

@article{weber2024redpajama,
  title={Redpajama: an open dataset for training large language models},
  author={Weber, Maurice and Fu, Dan and Anthony, Quentin and Oren, Yonatan and Adams, Shane and Alexandrov, Anton and Lyu, Xiaozhong and Nguyen, Huu and Yao, Xiaozhe and Adams, Virginia and others},
  journal={Advances in neural information processing systems},
  volume={37},
  pages={116462--116492},
  year={2024}
}

@inproceedings{soldaini2024dolma,
  title={Dolma: An open corpus of three trillion tokens for language model pretraining research},
  author={Soldaini, Luca and Kinney, Rodney and Bhagia, Akshita and Schwenk, Dustin and Atkinson, David and Authur, Russell and Bogin, Ben and Chandu, Khyathi and Dumas, Jennifer and Elazar, Yanai and others},
  booktitle={Proceedings of the 62nd annual meeting of the association for computational linguistics (volume 1: long papers)},
  pages={15725--15788},
  year={2024}
}

@misc{barham2023megawikamillionsreportssources,
      title={MegaWika: Millions of reports and their sources across 50 diverse languages}, 
      author={Samuel Barham and Orion Weller and Michelle Yuan and Kenton Murray and Mahsa Yarmohammadi and Zhengping Jiang and Siddharth Vashishtha and Alexander Martin and Anqi Liu and Aaron Steven White and Jordan Boyd-Graber and Benjamin Van Durme},
      year={2023},
      eprint={2307.07049},
      archivePrefix={arXiv},
      primaryClass={cs.CL},
      url={https://arxiv.org/abs/2307.07049}, 
}

@misc{Huu2024VALID,
title = {VALID (Video-Audio Large Interleaved Dataset)},
author = {Huu Nguyen and Ken Tsui and Andrej Radonjic and Christoph Schuhmann},
year = {2024},
url = {https://huggingface.co/datasets/ontocord/VALID},
}

@inproceedings{heafield-etal-2022-europat,
    title = "The {E}uro{P}at Corpus: A Parallel Corpus of {E}uropean Patent Data",
    author = "Heafield, Kenneth  and
      Farrow, Elaine  and
      van der Linde, Jelmer  and
      Ram{\'i}rez-S{\'a}nchez, Gema  and
      Wiggins, Dion",
    editor = "Calzolari, Nicoletta  and
      B{\'e}chet, Fr{\'e}d{\'e}ric  and
      Blache, Philippe  and
      Choukri, Khalid  and
      Cieri, Christopher  and
      Declerck, Thierry  and
      Goggi, Sara  and
      Isahara, Hitoshi  and
      Maegaard, Bente  and
      Mariani, Joseph  and
      Mazo, H{\'e}l{\`e}ne  and
      Odijk, Jan  and
      Piperidis, Stelios",
    booktitle = "Proceedings of the Thirteenth Language Resources and Evaluation Conference",
    month = jun,
    year = "2022",
    address = "Marseille, France",
    publisher = "European Language Resources Association",
    url = "https://aclanthology.org/2022.lrec-1.78/",
    pages = "732--740"
}

@inproceedings{saxton2019analysingmathematicalreasoningabilities,
  title={Analysing Mathematical Reasoning Abilities of Neural Models},
  author={Saxton, David and Grefenstette, Edward and Hill, Felix and Kohli, Pushmeet},
  booktitle={International Conference on Learning Representations},
  year={2019},
}

@article{platypus2023,
  title={Platypus: Quick, Cheap, and Powerful Refinement of LLMs},
  author={Ariel N. Lee and Cole J. Hunter and Nataniel Ruiz},
  journal={NeurIPS 2023 Workshop on Instruction Tuning and Instruction Following},
  year={2023}
}

@article{mahabadi2025nemotron,
  title={Nemotron-cc-math: A 133 billion-token-scale high quality math pretraining dataset},
  author={Mahabadi, Rabeeh Karimi and Satheesh, Sanjeev and Prabhumoye, Shrimai and Patwary, Mostofa and Shoeybi, Mohammad and Catanzaro, Bryan},
  journal={arXiv preprint arXiv:2508.15096},
  year={2025}
}

@misc{hao2025reformulationpretrainingdataaugmentation,
      title={Reformulation for Pretraining Data Augmentation}, 
      author={Xintong Hao and Ruijie Zhu and Ge Zhang and Ke Shen and Chenggang Li},
      year={2025},
      eprint={2502.04235},
      archivePrefix={arXiv},
      primaryClass={cs.CL},
      url={https://arxiv.org/abs/2502.04235}, 
}

@inproceedings{zheng2021does,
  title={When does pretraining help? assessing self-supervised learning for law and the casehold dataset of 53,000+ legal holdings},
  author={Zheng, Lucia and Guha, Neel and Anderson, Brandon R and Henderson, Peter and Ho, Daniel E},
  booktitle={Proceedings of the eighteenth international conference on artificial intelligence and law},
  pages={159--168},
  year={2021}
}

@inproceedings{logacheva-etal-2022-paradetox,
    title = "{P}ara{D}etox: Detoxification with Parallel Data",
    author = "Logacheva, Varvara  and
      Dementieva, Daryna  and
      Ustyantsev, Sergey  and
      Moskovskiy, Daniil  and
      Dale, David  and
      Krotova, Irina  and
      Semenov, Nikita  and
      Panchenko, Alexander",
    editor = "Muresan, Smaranda  and
      Nakov, Preslav  and
      Villavicencio, Aline",
    booktitle = "Proceedings of the 60th Annual Meeting of the Association for Computational Linguistics (Volume 1: Long Papers)",
    month = may,
    year = "2022",
    address = "Dublin, Ireland",
    publisher = "Association for Computational Linguistics",
    url = "https://aclanthology.org/2022.acl-long.469/",
    doi = "10.18653/v1/2022.acl-long.469",
    pages = "6804--6818"
}

@inproceedings{lee2022deduplicating,
  title={Deduplicating training data makes language models better},
  author={Lee, Katherine and Ippolito, Daphne and Nystrom, Andrew and Zhang, Chiyuan and Eck, Douglas and Callison-Burch, Chris and Carlini, Nicholas},
  booktitle={Proceedings of the 60th Annual Meeting of the Association for Computational Linguistics (Volume 1: Long Papers)},
  pages={8424--8445},
  year={2022}
}

@article{raffel2020exploring,
  title={Exploring the limits of transfer learning with a unified text-to-text transformer},
  author={Raffel, Colin and Shazeer, Noam and Roberts, Adam and Lee, Katherine and Narang, Sharan and Matena, Michael and Zhou, Yanqi and Li, Wei and Liu, Peter J},
  journal={Journal of machine learning research},
  volume={21},
  number={140},
  pages={1--67},
  year={2020}
}

@misc{taori2023alpaca,
  title={Stanford alpaca: An instruction-following llama model},
  author={Taori, Rohan and Gulrajani, Ishaan and Zhang, Tianyi and Dubois, Yann and Li, Xuechen and Guestrin, Carlos and Liang, Percy and Hashimoto, Tatsunori B},
  year={2023},
  publisher={Stanford, CA, USA}
}

@article{chen2024diversity,
  title={On the diversity of synthetic data and its impact on training large language models},
  author={Chen, Hao and Waheed, Abdul and Li, Xiang and Wang, Yidong and Wang, Jindong and Raj, Bhiksha and Abdin, Marah I},
  journal={arXiv preprint arXiv:2410.15226},
  year={2024}
}

@misc{austin2021programsynthesislargelanguagen,
      title={Program Synthesis with Large Language Models}, 
      author={Jacob Austin and Augustus Odena and Maxwell Nye and Maarten Bosma and Henryk Michalewski and David Dohan and Ellen Jiang and Carrie Cai and Michael Terry and Quoc Le and Charles Sutton},
      year={2021},
      eprint={2108.07732},
      archivePrefix={arXiv},
      primaryClass={cs.PL},
      url={https://arxiv.org/abs/2108.07732}, 
}

@article{uddin2025advbench,
  title={AdvBench: A Comprehensive Benchmark of Adversarial Attacks on Deepfake Detectors in Real-World Consumer Applications},
  author={Uddin, Kutub and Farooq, Muhammad Umar and Khan, Awais and Saeed, Muhammad Saad and Haq, Ijaz Ul and Tasnim, Nusrat and Malik, Khalid Mahmood},
  journal={Authorea Preprints},
  year={2025},
  publisher={Authorea}
}

@article{clement2019use,
  title={On the use of arxiv as a dataset},
  author={Clement, Colin B and Bierbaum, Matthew and O'Keeffe, Kevin P and Alemi, Alexander A},
  journal={arXiv preprint arXiv:1905.00075},
  year={2019}
}

@article{apertus2025apertusdemocratizingopencompliant,
  title={Apertus: Democratizing open and compliant llms for global language environments},
  author={Hern{\'a}ndez-Cano, Alejandro and H{\"a}gele, Alexander and Huang, Allen Hao and Romanou, Angelika and Solergibert, Antoni-Joan and Pasztor, Barna and Messmer, Bettina and Garbaya, Dhia and {\v{D}}urech, Eduard Frank and Hakimi, Ido and others},
  journal={arXiv preprint arXiv:2509.14233},
  year={2025}
}

@article{Li_2022,
   title={Competition-level code generation with AlphaCode},
   volume={378},
   ISSN={1095-9203},
   url={http://dx.doi.org/10.1126/science.abq1158},
   DOI={10.1126/science.abq1158},
   number={6624},
   journal={Science},
   publisher={American Association for the Advancement of Science (AAAS)},
   author={Li, Yujia and Choi, David and Chung, Junyoung and Kushman, Nate and Schrittwieser, Julian and Leblond, Rémi and Eccles, Tom and Keeling, James and Gimeno, Felix and Dal Lago, Agustin and Hubert, Thomas and Choy, Peter and de Masson d’Autume, Cyprien and Babuschkin, Igor and Chen, Xinyun and Huang, Po-Sen and Welbl, Johannes and Gowal, Sven and Cherepanov, Alexey and Molloy, James and Mankowitz, Daniel J. and Sutherland Robson, Esme and Kohli, Pushmeet and de Freitas, Nando and Kavukcuoglu, Koray and Vinyals, Oriol},
   year={2022},
   month=dec, pages={1092–1097} }

@misc{kaplan2020scalinglawsneurallanguage,
      title={Scaling Laws for Neural Language Models}, 
      author={Jared Kaplan and Sam McCandlish and Tom Henighan and Tom B. Brown and Benjamin Chess and Rewon Child and Scott Gray and Alec Radford and Jeffrey Wu and Dario Amodei},
      year={2020},
      eprint={2001.08361},
      archivePrefix={arXiv},
      primaryClass={cs.LG},
      url={https://arxiv.org/abs/2001.08361}, 
}

@misc{codefuse2025samplemattersleveragingmixtureofexperts,
      title={Every Sample Matters: Leveraging Mixture-of-Experts and High-Quality Data for Efficient and Accurate Code LLM}, 
      author={{Codefuse Team} and {Ling Team} and Wenting Cai and Yuchen Cao and Chaoyu Chen and Chen Chen and Siba Chen and Qing Cui and Peng Di and Junpeng Fang and Zi Gong and Ting Guo and Zhengyu He and Yang Huang and Cong Li and Jianguo Li and Zheng Li and Shijie Lian and BingChang Liu and Songshan Luo and Shuo Mao and Min Shen and Jian Wu and Jiaolong Yang and Wenjie Yang and Tong Ye and Hang Yu and Wei Zhang and Zhenduo Zhang and Hailin Zhao and Xunjin Zheng and Jun Zhou},
      year={2025},
      eprint={2503.17793},
      archivePrefix={arXiv},
      primaryClass={cs.LG},
      url={https://arxiv.org/abs/2503.17793}, 
}

@misc{map2024finefineweb,
title={FineFineWeb: A Comprehensive Study on Fine-grained Domain Web Corpus},
url={[https://huggingface.co/datasets/m-a-p/FineFineWeb](https://huggingface.co/datasets/m-a-p/FineFineWeb)},
author = {{M-A-P} and Ge Zhang and Xinrun Du and Zhimiao Yu and Zili Wang and Zekun Wang and Shuyue Guo and Tianyu Zheng and Kang Zhu and Jerry Liu and Shawn Yue and Binbin Liu and Zhongyuan Peng and Yifan Yao and Jack Yang and Ziming Li and Bingni Zhang and Minghao Liu and Tianyu Liu and Yang Gao and Wenhu Chen and Xiaohuan Zhou and Qian Liu and Taifeng Wang and Wenhao Huang},
publisher={huggingface},
verision={v0.1.0},
month={December},
year={2024}
}

@article{10.1093/grurint/ikac054,
    author = {Margoni, Thomas and Kretschmer, Martin},
    title = {A Deeper Look into the EU Text and Data Mining Exceptions: Harmonisation, Data Ownership, and the Future of Technology},
    journal = {GRUR International},
    volume = {71},
    number = {8},
    pages = {685-701},
    year = {2022},
    month = {07},
    issn = {2632-8623},
    doi = {10.1093/grurint/ikac054},
    url = {https://doi.org/10.1093/grurint/ikac054},
    eprint = {https://academic.oup.com/grurint/article-pdf/71/8/685/45289348/ikac054.pdf},
}

@inproceedings{
fan2025can,
title={Can Performant {LLM}s Be Ethical? Quantifying the Impact of Web Crawling Opt-Outs},
author={Dongyang Fan and Vinko Sabol{\v{c}}ec and Matin Ansaripour and Ayush Kumar Tarun and Martin Jaggi and Antoine Bosselut and Imanol Schlag},
booktitle={Second Conference on Language Modeling},
year={2025},
url={https://openreview.net/forum?id=a6QsOjr3wo}
}

@article{txt360,
  title = {{TxT360: A Top-Quality LLM Pre-training Dataset Requires the Perfect Blend}},
  author={Liping Tang and Nikhil Ranjan and Omkar Pangarkar and Xuezhi Liang and Zhen Wang and Li An and Bhaskar Rao and Linghao Jin and Huijuan Wang and Zhoujun Cheng and Suqi Sun and Cun Mu and Victor Miller and Xuezhe Ma and Yue Peng and Zhengzhong Liu and Eric P Xing},
  year={2024},
  url={https://huggingface.co/spaces/LLM360/TxT360}
}

@inproceedings{maini2024rephrasing,
  title={Rephrasing the web: A recipe for compute and data-efficient language modeling},
  author={Maini, Pratyush and Seto, Skyler and Bai, Richard and Grangier, David and Zhang, Yizhe and Jaitly, Navdeep},
  booktitle={Proceedings of the 62nd Annual Meeting of the Association for Computational Linguistics (Volume 1: Long Papers)},
  pages={14044--14072},
  year={2024}
}

@article{toshniwal2024openmathinstruct,
  title={Openmathinstruct-1: A 1.8 million math instruction tuning dataset},
  author={Toshniwal, Shubham and Moshkov, Ivan and Narenthiran, Sean and Gitman, Daria and Jia, Fei and Gitman, Igor},
  journal={Advances in Neural Information Processing Systems},
  volume={37},
  pages={34737--34774},
  year={2024}
}

@inproceedings{lozhkovsmollm2,
  title={Smollm2: When smol goes big—data-centric training of a fully open small language model},
  author={Ben allal, Loubna and Lozhkov, Anton and Bakouch, Elie and Blazquez, Gabriel Martin and Penedo, Guilherme and Tunstall, Lewis and Marafioti, Andr{\'e}s and Lajar{\'\i}n, Agust{\'\i}n Piqueres and Kydl{\'\i}{\v{c}}ek, Hynek and Srivastav, Vaibhav and Lochner, Joshua and others},
  booktitle={Second Conference on Language Modeling},
year={2025}
}
\bibliographystyle{iclr2026_conference}

\clearpage

\begin{center}
{\Large\bf Appendix: MixtureVitae -- Open Web-Scale Pretraining Dataset With High Quality Instruction and Reasoning Data Built from Permissive Text Sources}
\end{center}

\appendix
\section{Dataset Composition and Comparison}
This appendix provides a detailed view of the \MixtureVitae{} corpus, both in relation to other datasets and in its internal construction.

\begin{table*}[htbp]
\center
\caption{Comparison of large-scale pretraining datasets, grouped by their licensing philosophy to provide context for our performance results. \MixtureVitae{} is unique in its combination of a risk-mitigated licensing approach and the inclusion of a large subset of reasoning, coding and instruction synthetic data.} 
\label{tab:dataset_comparison}
\footnotesize
\renewcommand{\arraystretch}{1.5} 
\scalebox{0.88}{
\begin{tabular}{@{}l l l l@{}}
\toprule
\textbf{Dataset} & \textbf{\makecell[l]{Size \\ (Tokens)}} & \textbf{\makecell[l]{Primary \\ Data Types}} & \textbf{\makecell[l]{Licensing \\ Philosophy}} \\
\midrule
\multicolumn{4}{@{}l}{\textit{Non-Permissive / Mixed-License Baselines}} \\
\quad Nemotron-CC-HQ \citep{su2025nemotroncc} & $\approx$ 1.1T & Web, Synthetic & Unspecified \\
\quad DCLM-baseline \citep{li2024datacomplm} & $\approx$ 3.8T & Web, Code, Academic & Mixed / Unspecified \\
\quad FineWeb-Edu \citep{penedo2024} & $\approx$ 1.3T & Web (Educational) & Unspecified \\
\quad The Pile \citep{gao2020pile} & $\approx$ 183.28B & Web, Books, Code & Mixed / Unspecified \\
\quad SlimPajama \citep{shen2024slimpajamadcunderstandingdatacombinations} & $\approx$ 627B & Web, Books, Code & Mixed / Unspecified \\
\quad C4 \citep{raffel2020exploring} & $\approx$ 156B & Web & ODC-BY \\
\quad HPLT-2.0 (eng.) \citep{burchell2025} & $\approx$ 2.86T & Web, Books, News & Mixed / Unspecified \\
\midrule
\multicolumn{4}{@{}l}{\textit{Permissive Baselines}} \\
\quad CommonCorpus \citep{langlais2025commoncorpuslargestcollection} & $\approx$ 2T & Web, Curated & Strictly Permissive \\
\quad Comma-0.1 \citep{kandpal2025commonpilev018tb} & $\approx$ 1T & Web, Curated & Strictly Permissive \\
\quad KL3M \citep{bommarito2025kl3mdataprojectcopyrightclean} & $\approx$ 580B & Web, Curated & Strictly Permissive \\
\quad OLC \citep{min2024silo} & $\approx$ 228B & Web, Curated & Strictly Permissive \\
\midrule
\multicolumn{4}{@{}l}{\textit{Our Contribution}} \\
\quad \MixtureVitae{} & \textbf{$\approx$ \DatasetSize{}B} & \textbf{\makecell[l]{Web, Curated, \\  Synthetic}} & \textbf{\makecell[l]{Permissive-First, \\ Risk-Mitigated}} \\
\bottomrule
\end{tabular}
}

\end{table*}

\paragraph{Shard Definitions and Mixing.}
It is important to note that the dataset shards and categories listed in this appendix serve as logical groupings for transparency, licensing audits, and ablation analysis. They do not dictate a rigid sequential training order. As noted in the main text, the physical construction of training batches utilizes domain-aware packing to maximize local coherence, prioritizing the density of reasoning and factual tokens over these high-level taxonomic boundaries.

Table~\ref{tab:dataset_comparison} presents a high-level comparison of \MixtureVitae{} against the other prominent pretraining datasets evaluated in our experiments, detailing their respective sizes, primary data types, and licensing philosophies.

\begin{figure}[htbp] 
    \centering    
    \includegraphics[width=1\linewidth]{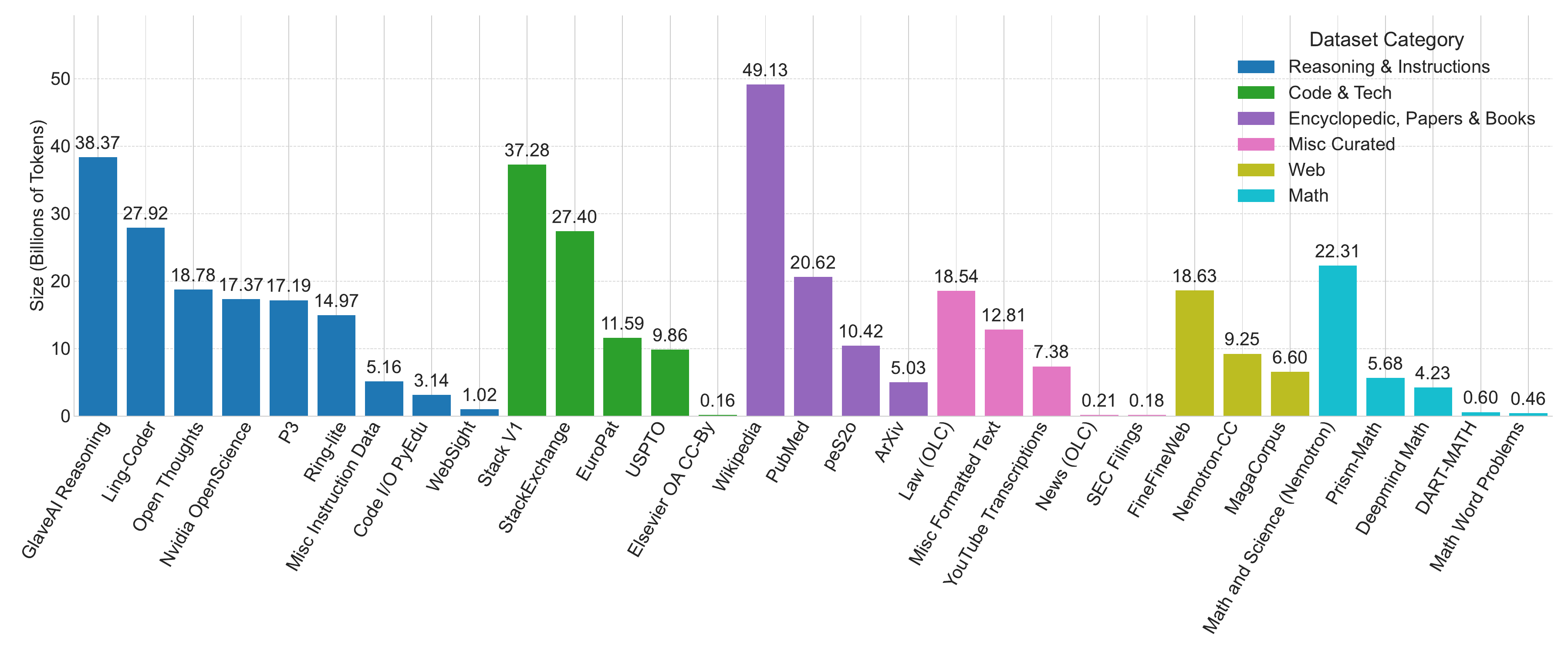}
      \caption[Detailed Composition of the MixtureVitae Dataset]{Detailed composition of the \MixtureVitae{} dataset.} 
    \label{fig:data_comp_deets}
\end{figure}

Figure~\ref{fig:data_comp_deets} presents the detailed composition of the \MixtureVitae{} dataset. The individual components are color-coded by their primary dataset category, as presented in the main text.

\begin{itemize}
    \item \textbf{Code \& Tech (Blue):} This domain is anchored by our largest code sources, Stack V1 and Ling-Coder, and supplemented by StackExchange.
    
    \item \textbf{Reasoning \& Instruction (Green):} The largest contributor to this category is Open Thoughts , followed by P3 and NVIDIA OpenScience.
    
    \item \textbf{Encyclopedic, Papers \& Books (Purple):} This category is dominated by Wikipedia, the single largest component in the dataset. It is complemented by large-scale text from PubMed  and arXiv.
    
    \item \textbf{Math (Cyan):} The math component is a diverse mixture of sources, led by the Math and Science (Nemotron) corpus and Prism-Math.
    
    \item \textbf{Web (Yellow):} Our web data is primarily sourced from corpora such as SEC Filings, MegaCorpus, and FineFineWeb.
    
    \item \textbf{Misc Curated (Pink):} This category includes a variety of high-quality curated sources, notably Law (Open License Corpus) and YouTube Transcriptions.
\end{itemize}

\section{Experiment Setup Details}
\label{app:exp_setup}
To ensure full reproducibility, this appendix details the complete experimental setup. This includes the model architectures for all scales, the training hyperparameters for both 50B and 300B token budgets, and the specific settings used for all general evaluation benchmarks.



\begin{table*}[htbp]
\centering
\caption{\textbf{open-sci-ref}~\citep{nezhurina2025opensciref001openreproduciblereference} model architecture and scales. We used tied embedding weights in all experiments.}
\label{tab:opensciref_scales}

\renewcommand{\arraystretch}{1.5}
\begin{tabular}{@{}l c c c c c c@{}}
\toprule
\textbf{\makecell[l]{Parameters (B) \\ (Non-Emb + Emb)}} &
\textbf{Layers} &
\textbf{Hidden} &
\textbf{Heads} &
\textbf{\makecell[l]{FFN \\ Hidden}} &
\textbf{Memory} &
\textbf{\makecell[l]{FLOPs}} \\
\midrule
$0.1 + 0.03 = 0.13$ & 22 & 512  & 8  & 2256 & $0.89\,\mathrm{GB}$  & $7.8\times10^{8}$ \\
$0.35 + 0.05 = 0.40$ & 22 & 1024 & 16 & 3840 & $2.88\,\mathrm{GB}$ & $2.4\times10^{9}$ \\
$1.21 + 0.10 = 1.31$ & 24 & 2048 & 32 & 5440 & $7.544\,\mathrm{GB}$ & $7.9\times10^{9}$ \\
$1.61 + 0.10 = 1.71$ & 24 & 2048 & 32 & 8192 & $9.884\,\mathrm{GB}$ & $1.0\times10^{10}$ \\
\bottomrule
\end{tabular}
\end{table*}

\begin{table*}[htbp]
\centering
\caption{The training schedules used in our experiments.}
\label{tab:training_schedules}
\renewcommand{\arraystretch}{1.5}
\begin{tabular}{@{}l c c c c c@{}}
\toprule
\textbf{Tokens} &
\textbf{\makecell[c]{Global Batch Size \\ (tokens)}} &
\textbf{Iterations} &
\textbf{\makecell[c]{Learning \\ Rate}} &
\textbf{Warmup} &
\textbf{\makecell[c]{Cooldown \\ (20\%)}} \\
\midrule
50B  & 4.12M & 11,921  & $4\times10^{-3}$ & 1,000  & 2,384  \\
300B & 4.12M & 72,661  & $4\times10^{-3}$ & 25,000 & 14,532 \\
\bottomrule
\end{tabular}

\end{table*}

\subsection{Training Setup Parameters}
\label{app:hyperparams}

This appendix details the exact model architectures and training hyperparameters used for all experiments, ensuring full reproducibility. 

We adopt the standard architectures and scales from the \textbf{open-sci-ref} framework to allow for a fair and direct comparison against other published baselines. All models were trained with tied embedding weights. 

\paragraph{Model Architecture}
Table~\ref{tab:opensciref_scales} defines the four model scales used in our study. The columns are defined as follows:

\begin{description}
    \item[Parameters (B) (Non-Emb + Emb)] The total model parameters in billions, separated into \textbf{Non-Embedding} (Non-Emb) parameters (the core transformer blocks) and \textbf{Embedding} (Emb) parameters (the token lookup tables). As noted in the caption, we used tied embedding weights.
    \item[Layers] The total number of transformer blocks stacked in the model.
    \item[Hidden] The hidden size (or embedding dimension, $d_{\text{model}}$) of the model.
    \item[Heads] The number of attention heads in the multi-head attention mechanism.
    \item[FFN Hidden] The inner dimension of the Feed-Forward Network (FFN) layer within each transformer block.
    \item[Memory] The approximate VRAM required to store the model weights, in bfloat16.
    \item[FLOPs] An approximation of the training compute cost using the  \textbf{6N} rule: a standard estimate for a transformer's forward-and-backward pass, where \textbf{N} is the number of \textit{non-embedding} parameters~\citep{kaplan2020scalinglawsneurallanguage}.
\end{description}

\paragraph{Training Schedules}
Table~\ref{tab:training_schedules} defines the training hyperparameters for our two main experimental runs (50B and 300B tokens). We use a single stage training with no post-training.

\begin{description}
    \item[Tokens] The total number of tokens in the training run.
    \item[Global Batch Size (tokens)] The total number of tokens processed in a single training step (i.e., one gradient update) across all GPUs.
    \item[Iterations] The total number of training steps.
    \item[Learning Rate] The peak learning rate used during training.
    \item[Warmup] The number of initial \textit{iterations} (steps) over which the learning rate linearly increases from 0 to its peak value.
    \item[Cooldown (20\%)] The number of final \textit{iterations} (the last 20\% of training) over which the learning rate decays to zero.
\end{description}


\subsection{Evaluation Settings}
\label{sec:appendix_eval_settings}

We used the \texttt{lm-evaluation-harness} \citep{eval-harness} for all general evaluations. The specific tasks and few-shot counts are detailed in Table~\ref{tab:eval_settings_general}. The settings for the reasoning tasks (e.g., GSM8K, IFEval) are listed separately in Table\ref{tab:reasoning-eval-settings}.

\begin{table}[ht]
\centering
\caption{General evaluation benchmark settings. All tasks use Accuracy as the primary metric.}
\label{tab:eval_settings_general}
\begin{tabular}{@{}llc@{}}
\toprule
Task & Citation & \# of Shots \\
\midrule
MMLU & \cite{hendrycks2021mmlu} & 5 \\
HellaSwag & \cite{zellers2019hellaswag} & 10 \\
CommonSenseQA & \cite{talmor2019commonsenseqa} & 10 \\
ARC-Challenge & \cite{clark2018arc} & 10 \\
ARC-Easy & \cite{clark2018arc} & 10 \\
PIQA & \cite{bisk2019piqareasoningphysicalcommonsense} & 10 \\
BoolQ & \cite{clark2019boolq} & 10 \\
Winogrande & \cite{sakaguchi2020winogrande} & 0 \\
OpenBookQA & \cite{mihaylov2018openbookqa} & 0 \\
COPA & \cite{copa} & 0 \\
LAMBADA & \cite{paperno2016lambada} & 0 \\
\bottomrule
\end{tabular}
\end{table}


\begin{table}[ht]
\centering
\caption{Evaluation settings for reasoning tasks. All tasks use Accuracy as the primary metric. To execute the evaluation, we used LM Evaluation Harness \cite{eval-harness}.}
\label{tab:reasoning-eval-settings}
\begin{tabular}{llccl}
\toprule
Task & Citation & \# of Shots \\
\midrule
GSM8k & ~\cite{cobbe2021training} &  4 \\
IFEval & ~\cite{zhou2023instruction} &  0 \\
MBPP & ~\cite{austin2021programsynthesislargelanguagen} & 4 \\
\bottomrule
\end{tabular}
\end{table}
\section{Additional Experiments}
\label{app:additional_exp}

This appendix provides additional experimental results to supplement the findings presented in the main paper. We offer a more granular breakdown of the 300B token experiment, analyze performance at a smaller 50B token scale to assess the generalization of our results, and report the results of a model red-teaming analysis to evaluate the model's safety profile.

\subsection{300B Experiment - Detailed Results}
\label{app:additional_exp_tasks_300B}

\begin{figure}[ht] 
    \centering
    \includegraphics[width=1\linewidth]{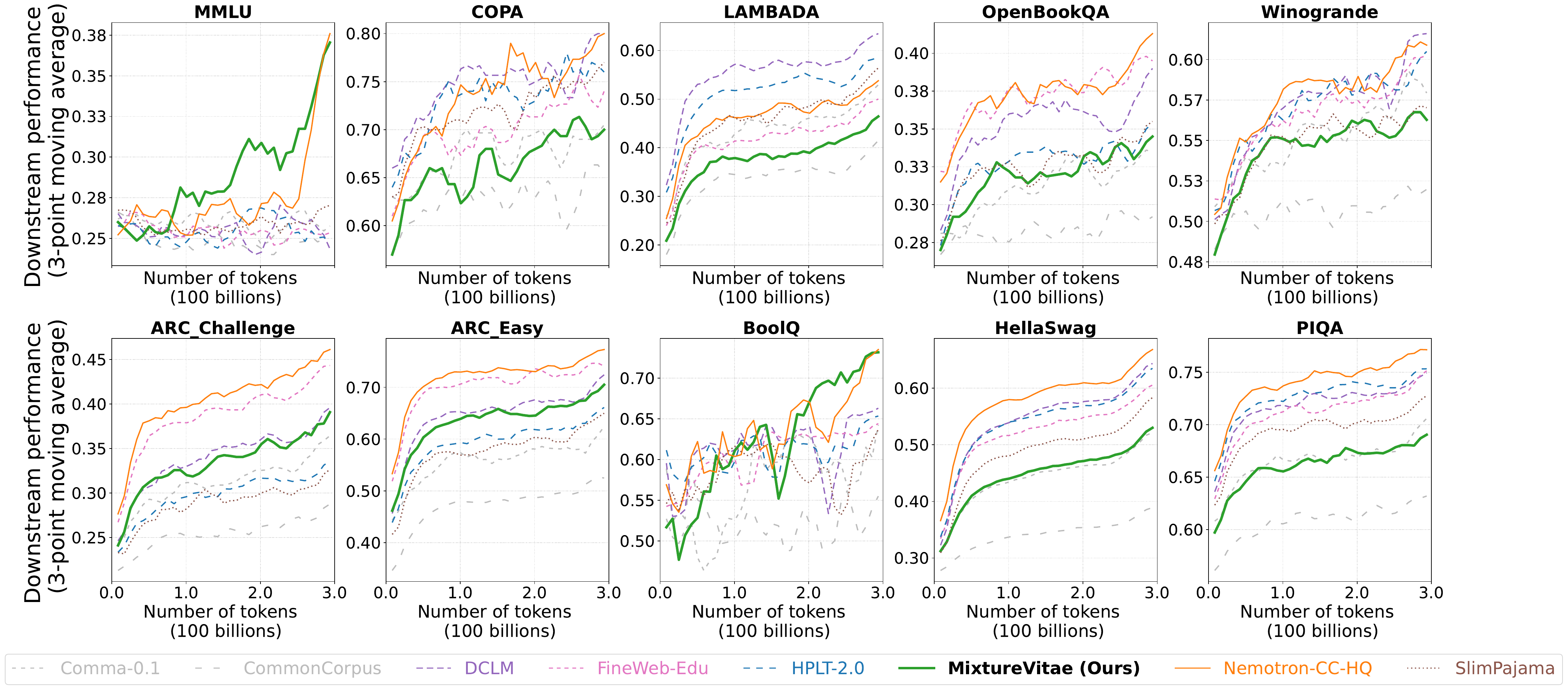}
    \caption{Comparing performance of 1.7B model trained on \MixtureVitae{} and baseline datasets for a 300B token budget. While some evaluations provide clear dataset rankings (e.g. ARC, Hellaswag, Lambada), others do not provide a good signal for dataset comparison, on an individual basis.}
    \label{fig:per_dataset_performance}
\end{figure}

The detailed results for each evaluated task (contributing to the average over 10 tasks as shown in Figure \ref{fig:300b_performance}) are given in Figure~\ref{fig:per_dataset_performance}. Despite its substantial proportion of instruction and reasoning data which gives \MixtureVitae{} exceptional performance for base model of ts scale on reasoning related tasks, \MixtureVitae{} demonstrates also strong performance on language tasks that are typically associated with pretraining on broad web scrapes (see also Table~\ref{tab:lang_results} in main results Sec. \ref{sec:experiment_results}).

\subsection{Performance at 50B Tokens Scale.} 

\begin{figure}[htbp]
    \centering
    \includegraphics[width=\linewidth]{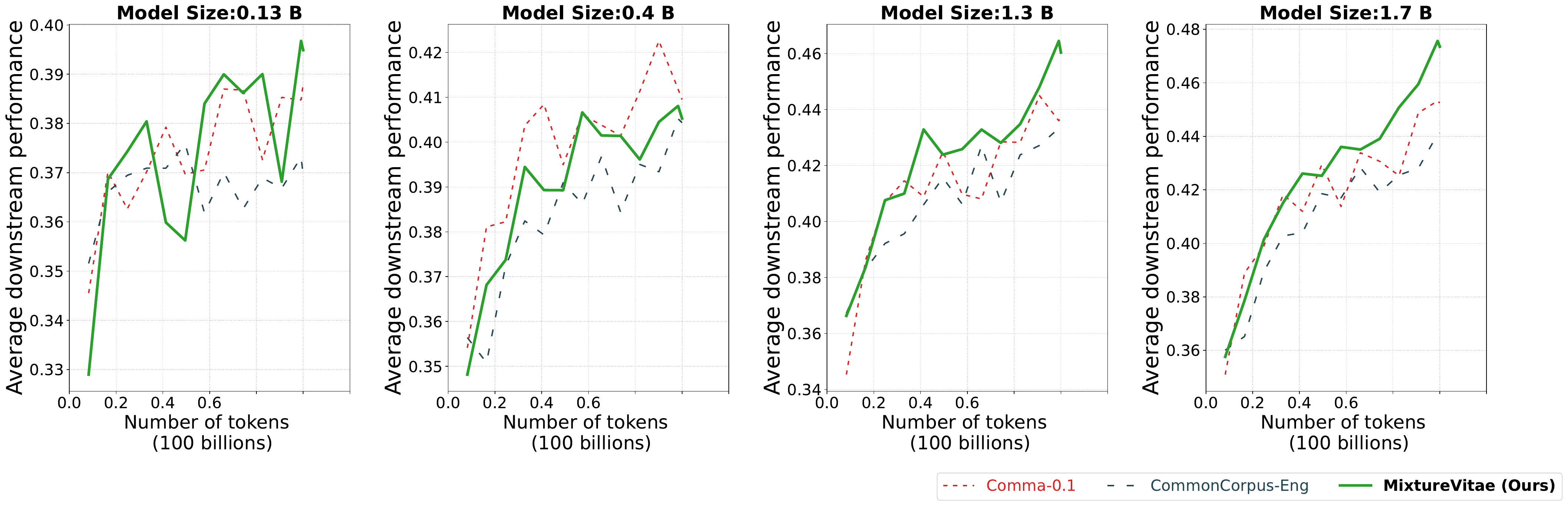}
    \caption{Average performance of permissive datasets after 50B training tokens. \MixtureVitae{} shows an early and consistent lead at larger model scales.}
    \label{fig:50b_avg_perf}
\end{figure}

\begin{figure}[htbp]
    \centering
    \includegraphics[width=\linewidth]{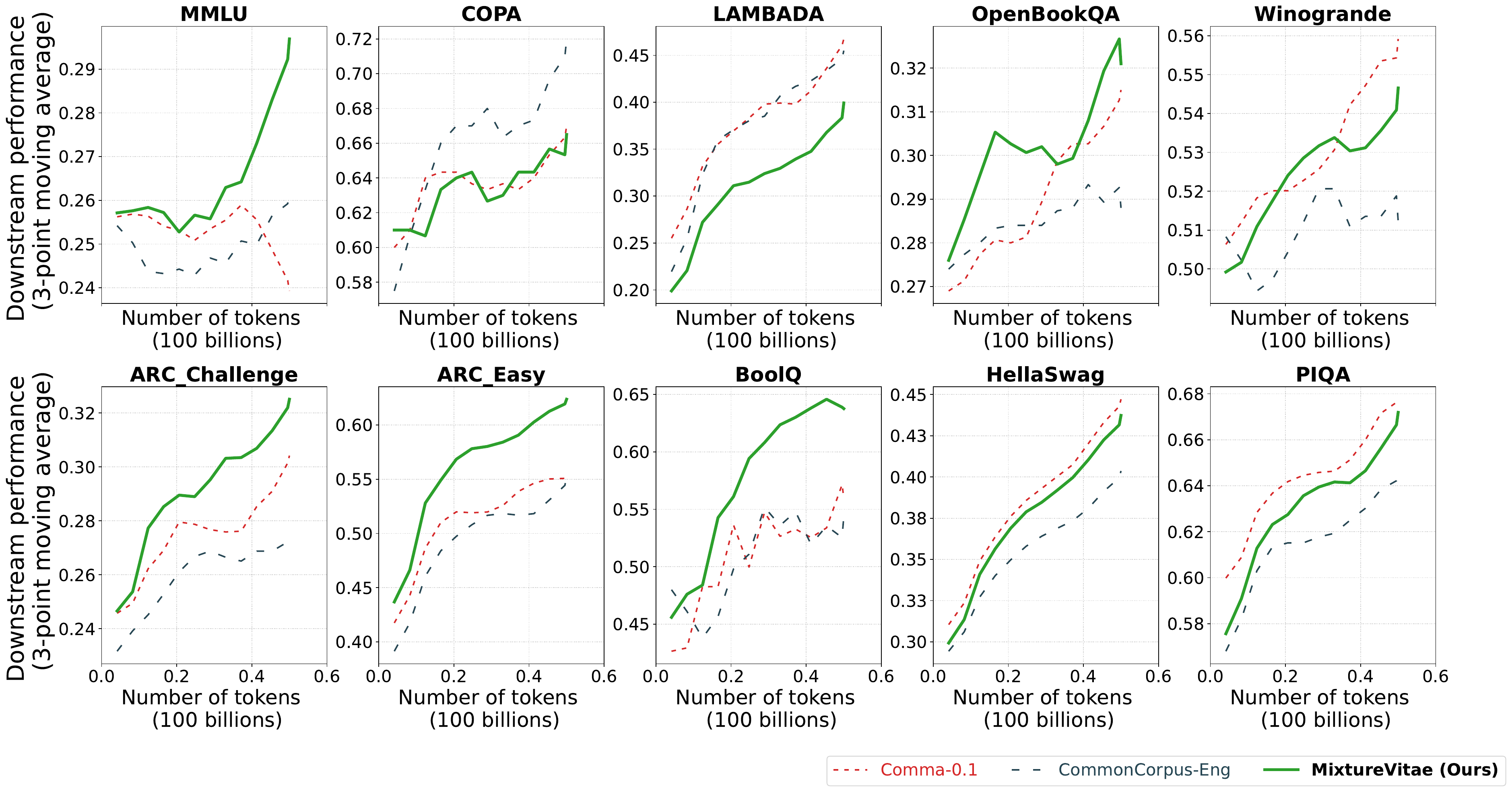}
    \caption{Per-benchmark performance of permissive datasets after 50B training tokens. \MixtureVitae{}'s advantage on MMLU is apparent even at this early stage.}
    \label{fig:50b_per_benchmark_perf}
\end{figure}

\label{app:50b_exp} To assess performance on a smaller reference tokens scale, we also evaluated models trained on a 50B token subset of each  dataset. The results, shown in Figure~\ref{fig:50b_avg_perf} and Figure~\ref{fig:50b_per_benchmark_perf}, indicate that the advantages of \MixtureVitae{} manifest already at the smaller token scales. Figure~\ref{fig:50b_avg_perf} shows that \MixtureVitae{} establishes a consistent performance lead over other permissive datasets within the first 50B tokens, especially at the 1.3B and 1.7B model scales. The per-benchmark analysis further reinforces this finding (see Figure~\ref{fig:50b_per_benchmark_perf}). On MMLU, \MixtureVitae{} is the only permissive dataset to show a significant learning signal early in training, demonstrating that its composition provides immediate benefits, which might be both due to knowledge rich and instruction like content. Arguably, this suggests that the reasoning capability shown by \MixtureVitae{} is not a late-stage phenomenon but rather an indication of efficient instillation from the early stages of training. This strong initial performance underscores the learning efficiency of \MixtureVitae{}, making it a compelling choice for achieving high performance with less computational cost.

\subsection{Model Red Teaming}
\label{app:red_teaming}
To evaluate the safety of the model trained on \MixtureVitae{} for 300B tokens, we performed a red-teaming analysis to measure the Attack Success Rate (ASR) against three standard benchmarks: \textbf{ToxiGen} \citep{hartvigsen2022toxigen}, \textbf{Do-Not-Answer} \citep{wang2024not}, and \textbf{AdvBench} \citep{uddin2025advbench}. The results (Table~\ref{tab:model_performance}) shows that our model is competitive with the baselines.

The model responses were evaluated using two safety classifiers: (i) \textbf{Llama Guard-8B} \citep{inan2023llamaguardllmbasedinputoutput}, used to evaluate the \textbf{Do-Not-Answer} and \textbf{AdvBench} datasets, while (ii) the \textbf{toxigen\_roberta} classifier \citep{logacheva-etal-2022-paradetox} was used for the \textbf{ToxiGen} benchmark.

\begin{table}[ht]
\centering
\caption{Attack Success Rate in \%, lower is better. All models are trained with the same \textbf{open-sci-ref} procedure (300B-token budget) while varying only the pretraining dataset.}
\label{tab:model_performance}
\begin{tabular}{lcccc}
\toprule
\textbf{Benchmark} & \textbf{\MixtureVitae} & \textbf{Comma} & \textbf{CommonCorpus-Eng} & \textbf{Nemotron-HQ-CC} \\
\midrule
ToxiGen & 8.07 & 9.04 & 12.77 & 10.21 \\
Do-Not-Answer & 28.22 & 24.71 & 21.62 & 20.98 \\
AdvBench & 86.92 & 92.12 & 70.58 & 85.77 \\
\bottomrule
\end{tabular}
\end{table}

\section{Contamination Analysis}
\label{app:contamination}
\subsection{Contamination Detection Protocol}

To ensure the integrity of our evaluation, we implemented a comprehensive decontamination protocol to measure the overlap between our training dataset and all evaluation benchmarks we report results on. This protocol consists of three main stages: Index Construction, Dataset Scanning, and Leakage Reporting.

\subsubsection{Index Construction}
The first stage creates a compact, indexed set of unique n-grams from all benchmark evaluation data.

\begin{enumerate}
    \item \textbf{Text Normalization:} All text from the benchmarks is processed through a  normalization pipeline, similar to ~\cite{NEURIPS2022_ce9e92e3}: (1) Unicode normalization (NFKC), (2) conversion to lowercase, (3) tokenization, and (4) removal of a predefined list of common English stop words. This procedure focuses the resulting n-grams on substantive content.
    
    \item \textbf{N-gramming and Filtering:} We generate 13-grams, a common n-gram size for this task~\cite{brown2020language, gao2020pile} from the normalized token lists. As in~\cite{NEURIPS2022_ce9e92e3}, a set of regular expressions is used to filter out common boilerplate, exam instructions, and formatting artifacts.

     \item \textbf{Train/Test De-duplication:} as in~\cite{gao2020pile}, we compute the set of all 13-gram hashes from the \texttt{train} split and subtract this set from the 13-gram hashes generated from the \texttt{test} split. This ensures our index only contains n-grams that are unique to the evaluation set.
\end{enumerate}
\subsubsection{Dataset Scanning}
The second stage analyzes the target training dataset against the generated index.

\begin{enumerate}
    \item \textbf{Document Processing:} Each document in the training dataset is processed using the \textit{exact same} normalization, 13-gramming, and hashing pipeline used for index construction.
    
    \item \textbf{Contamination Criteria:} A document is flagged as "contaminated" if it meets two criteria, based on the set intersection of its n-gram hashes with the benchmark index:
    \begin{itemize}
        \item \textbf{Minimum Hits:} The number of distinct matching n-grams is $\geq 3$.
        \item \textbf{Minimum Coverage:} As proposed in~\cite{rae2022scalinglanguagemodelsmethods}, the coverage of matching n-grams is $\geq 0.1\%$. Coverage is defined as:
        $$
        \text{Coverage} = \frac{\text{distinct\_hits}}{\text{total\_unique\_13grams\_in\_doc}}
        $$
    \end{itemize}
\end{enumerate}

\subsubsection{Leakage Reporting}
The final stage aggregates the scan results into a summary report.

\begin{enumerate}
    \item \textbf{Numerator (Leaked N-grams):} The procedure aggregates the reports from all scanned partitions. It performs a global \textit{set union} to find all unique n-gram hashes that were found \textit{at least once} in the target dataset, aggregated by benchmark source. This provides the $\text{unique\_ngrams\_leaked}$ count for each benchmark.
    
    \item \textbf{Denominator (Total N-grams):} The procedure retrieves the pre-computed metadata to obtain the total unique n-gram count for each benchmark.
    
    \item \textbf{Final Metric:} As proposed in~\cite{touvron2023llamaopenefficientfoundation}, the \textbf{Leak Percentage} for each benchmark is then calculated as:
    $$
    \text{Leak Percentage} = \frac{\text{unique\_ngrams\_leaked}_{\text{benchmark}}}{\text{total\_unique\_ngrams\_in\_index}_{\text{benchmark}}} \times 100
    $$
\end{enumerate}

\subsection{Contamination Report}
\label{app:contam_report}

We executed our 13-gram contamination scan across the entire \num{345697271} documents of the \MixtureVitae{} dataset. The global summary of contaminated documents per benchmark is presented in Table~\ref{tab:global_contam_summary}.

The results confirm that for the vast majority of benchmarks—including ARC, HellaSwag, LAMBADA, OpenBookQA, and PIQA—the document-level contamination rate is negligible (at or below 0.0003\%), strongly validating the integrity of our evaluation on these tasks.

The scan did, however, flag a minor overlap for MMLU (0.0098\%) and BoolQ (0.0087\%), and a more significant overlap for our key code benchmarks: HumanEval (0.0988\%) and MBPP (0.0878\%). This overlap in code benchmarks is a known challenge when including large-scale permissive code corpora like The Stack, which may naturally contain snippets of common coding problems (a "source overlap" rather than a direct "test-set leak").

To ensure this overlap did not artificially inflate our model's strong performance on these key tasks, we conducted case studies for the benchmarks with the highest overlap. This analysis is detailed in the following section (Appendix~\ref{app:decontam_gsm8k}).

\begin{table}[htb]
\caption{Global contamination summary by document count, based on a 13-gram overlap scan.
This table shows the total number of documents in \MixtureVitae{} that contained
at least one overlapping n-gram from each benchmark's test set.
The total documents in \MixtureVitae{} is \num{345697271} and the overall contamination rate is 0.1420\%.}
\centering
\begin{tabular}{lrr}
\toprule
Benchmark & {Contaminated Docs} & {Contamination Rate (\%)} \\
\midrule
ALERT         & \num{12}      & 0.0000\% \\
ARC           & \num{17}      & 0.0000\% \\
BoolQ         & \num{30144}   & 0.0087\% \\
CommonSenseQA          & \num{0}    & 0.0000\% \\
GPQA          & \num{1077}    & 0.0003\% \\
GSM8K         & \num{230}     & 0.0001\% \\
HellaSwag     & \num{186}     & 0.0001\% \\
HumanEval     & \num{341554}  & 0.0988\% \\
IfEval        & \num{756}     & 0.0002\% \\
LAMBADA       & \num{23}      & 0.0000\% \\
MBPP          & \num{303558}  & 0.0878\% \\
MMLU          & \num{33922}   & 0.0098\% \\
OpenBookQA    & \num{60}      & 0.0000\% \\
PIQA          & \num{5}       & 0.0000\% \\
SimpleQA      & \num{98}      & 0.0000\% \\
\bottomrule
\end{tabular}

\label{tab:global_contam_summary}
\end{table}

\subsection{Decontaminated Test Set Performance}
\label{app:decontam_gsm8k}
To understand how test data leakage affects final performance on downstream tasks, we conducted the following experiment on all models and benchmarks reported in Section~\ref{sec:instruct_results}:
\begin{enumerate}
    \item Identify problems from the test set that have at least one 13-gram match in the training dataset.
    \item Evaluate the model on a decontaminated benchmark version obtained by removing problems that were identified.
\end{enumerate}

\begin{table}[htb]
\caption{Benchmark test set sizes (number of examples) for the original benchmarks versus the final decontaminated versions. The 'Decontaminated' column shows the reduced set size after removing all examples with detected 13-gram training data overlap.}
\centering
\begin{tabular}{lcc}
\toprule
\textbf{Dataset} & \textbf{Original} & \textbf{Decontaminated} \\
\midrule
MBPP        & 500  & 331 \\
IFEval      & 541  & 429 \\
GSM8K       & 1319 & 1235 \\
MBPP+       & 378  & 339 \\
\bottomrule
\end{tabular}

\label{tab:decontam-sizes}
\end{table}


As we can see from Table~\ref{tab:decontam-results-validation}, the performance of the evaluated models is consistent between the original and decontaminated versions, aside from some upward bias in the decontaminated versions of MBPP+ and IFEval. Crucially, the result for the model trained on \MixtureVitae{} was not affected by the strict decontamination procedure applied to the benchmarks. This rules out the possibility that the strong performance of \MixtureVitae{} on math and coding is due to  benchmark leakage.

\subsection{Decontamination Case Study}
To further alleviate concerns about contamination issues, we performed an experiment where we trained a 1.7B model on a version of \MixtureVitae{} that excludes  three dataset shards that contribute 27\% of the total contaminated docs--- which in particular showed MMLU contamination rates that are high relative to the rest of the dataset, as shown in Table~\ref{tab:contam-breakdown-mmlu}. The shards we removed were \textbf{Misc-Instruct}, \textbf{DART-Math} and \textbf{Nemotron Science \& Math}. The results are shown in Figure~\ref{fig:ablate_app} and demonstrate that removing these shards had no effect on MMLU performance.

\begin{table}[htb]
\caption{Contamination sources for the \textbf{MMLU} benchmark in \MixtureVitae{}, sorted by the number of contaminated documents, high to low. }

\centering
\begin{tabular}{lr}
\toprule
\textbf{Dataset Shard} & \textbf{Contaminated Docs} \\
\midrule
Misc-Instruct & \num{14649} \\
DART-Math~\citep{tong2024dartmath} & \num{11102} \\
Nemotron Science \& Math~\citep{bercovich2025llamanemotronefficientreasoningmodels} & \num{4793} \\
MAGACorpus~\citep{hao2025reformulationpretrainingdataaugmentation} & \num{241} \\
(All Remaining) & \num{3137} \\
\bottomrule
\end{tabular}
\label{tab:contam-breakdown-mmlu}
\end{table}

\begin{figure}[htb]
    \centering
    \includegraphics[width=0.5\linewidth]{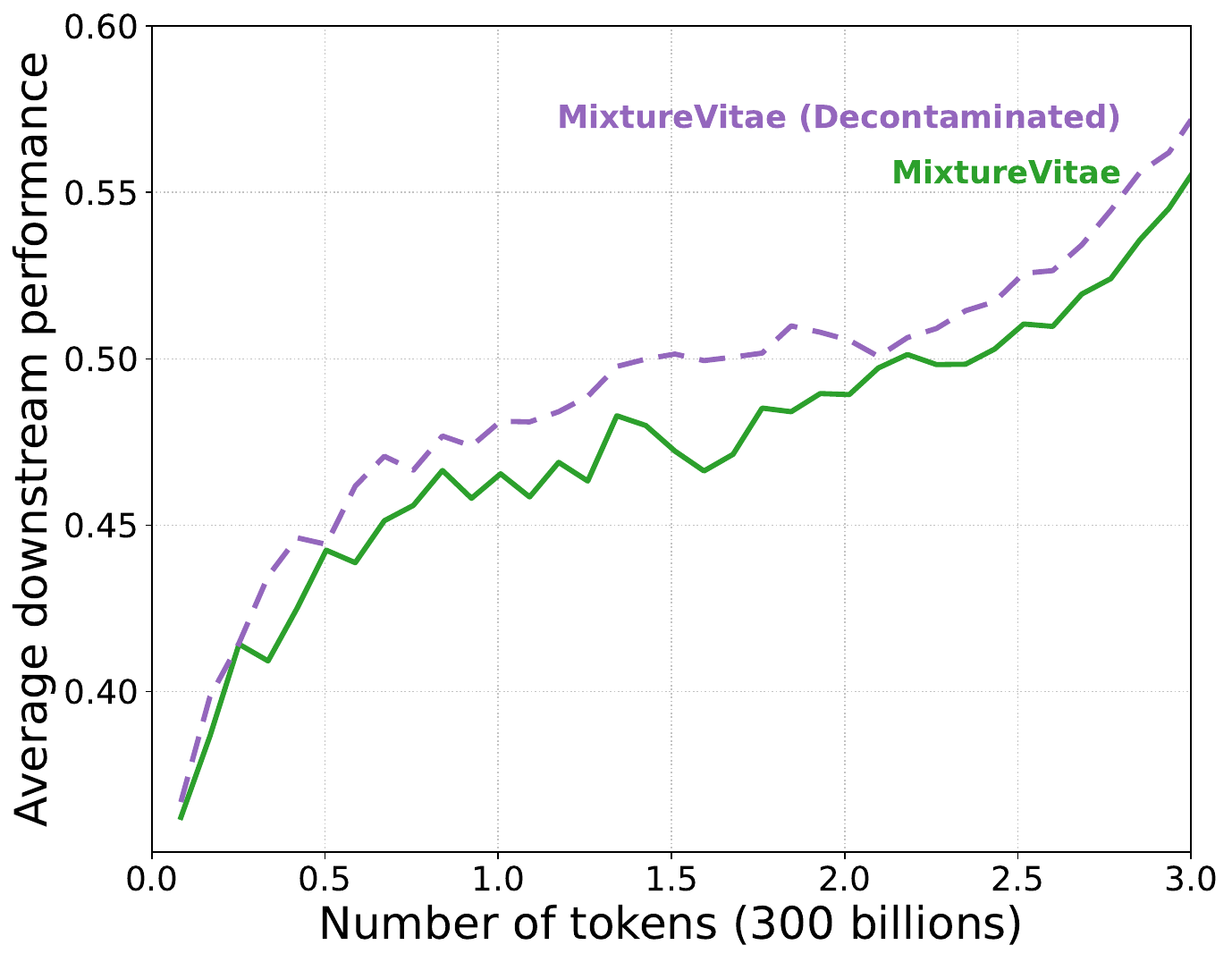}
    \caption{\textbf{1.7B model performance on a fully decontaminated dataset}. The model trained on the fully decontaminated \MixtureVitae{} corpus (purple, dashed) performs slightly better than the model trained on the full \MixtureVitae{} dataset (green, solid), further indicating that benchmark gains are not driven by contaminated examples.
 }
    \label{fig:full_decontam}
\end{figure}

\textbf{D.5 Full Decontamination Experiment} \\
We also performed a stronger decontamination experiment in which we removed \emph{every} document in \MixtureVitae{} that was flagged as contaminated by our 13-gram procedure (Appendix~\ref{app:contam_report}) and retrained a 1.7B model for 300B tokens under the \texttt{open-sci-ref} setup. The results---shown in Figure~\ref{fig:full_decontam}---indicate that the fully decontaminated variant performs slightly \emph{better} than the original \MixtureVitae{} model, further addressing concerns that our benchmark results might be inflated by data leakage.

\subsection{Discussion on Decontamination Methodology}
\label{app:decontam_discussion}

Our decontamination pipeline employs the standard 13-gram exact-match procedure~\citep{abdin2024phi4technicalreport} to ensure high precision, scalability and comparability with prior baselines. While we acknowledge that exact matching overlooks paraphrased content, we avoided approximate methods (e.g., LSH, embedding-based) due to their tendency to produce false positives on common factual or algorithmic templates \citep{lee2022deduplicating}. As noted in \citet{penedo2024}, aggressive removal of semantically similar content risks distorting the training distribution by discarding high-value instructional data.
\section{Synthetic Math Data Generation}
\label{app:mwp}
The synthetic math dataset was programmatically generated to produce a diverse range of mathematical problems and their solutions. The generation process covers a wide array of mathematical domains, including fundamental arithmetic operations, multi-term fractional expressions, and the step-by-step solution of algebraic linear equations. A key component of the dataset consists of word problems, where numerical challenges are embedded in narrative scenarios.

A significant feature of this generation pipeline is the creation of detailed, step-by-step solutions formatted as a chain-of-thought. For many problem categories, the scripts produce a human-readable explanation of the entire solution process. This is achieved by using a variety of randomized natural language templates to describe each logical step, such as carrying a digit in addition or isolating a variable in an equation.

Following the initial generation, a final post-processing step is applied to format the dataset for model training. This stage programmatically identifies data entries containing human-like, explanatory text by searching for common instructional words. For these selected entries, a descriptive header (e.g., "Here are examples of addition, division exercises") is dynamically generated. The content and phrasing of this header are randomized and based on the mathematical operations present within the text, adding significant linguistic diversity.


For example, in the generated math problem below, a model may be able to generalize to new numbers, but if the problem were to add three students instead of two, the model may not be robust enough to generalize. We leave this analysis for future work.

\begin{verbatim}

The age difference between Sarah and Asaf's age is half the 
total number of pencils Sarah has. The sum of their ages is 
132, and Sarah is 27 years old. If Asaf has 60 more pencils than 
Sarah, calculate the total number of pencils they have together.
Solution: If the sum of their ages is 132, and Sarah is 27 years 
old, Asaf is 105 years old.
The age difference between Sarah 
and Asaf's age is 105-27 = 78.
Since the age difference between Sarah and Asaf's age is half 
the total number of pencils Sarah has, Sarah has 2*78 = 156 pencils.
If Asaf has 60 more pencils than Sarah, Asaf has 156+60= 216 pencils.
Together, they have 156 + 216 = 372 pencils.
\end{verbatim}

\section{Our Position on Using Governmental and Other Works Under Fair Use and Related Ethical and Legal Basis}
\label{app:govt_works}

To contextualize our licensing tiers and clarify the rationale behind including certain higher-risk but legally supportable sources, we outline here the ethical and legal considerations underlying \MixtureVitae{}’s construction. Our goal is not to offer legal advice or definitive interpretations of copyright law, but rather to articulate the principles—fair use, permissive upstream licensing, government-works doctrine, and the EU text-and-data-mining (TDM)~\citep{eu_directive_tdm_2019,10.1093/grurint/ikac054} exception—that inform our “permissive-first, risk-mitigated” design philosophy. We provide this discussion so downstream users can understand how specific dataset subsets were evaluated and what residual risks remain despite our filtering and provenance-tracking efforts.

\subsection{Fair Use of Government Works}
In order to increase the diversity of our dataset, we included $\approx$11B tokens of governmental website data from US federal, US non‑federal, and non‑US government sources. While works created by the US federal government are generally not copyrightable, other governmental website content may neither be expressly in the public domain nor explicitly licensed. For those sources, we rely on fair use principles (United States Congress, 1976~\citep{uscopyright} ; Lemley \& Casey, 2017~\citep{lemley2017fair}) and the EU text and data mining exceptions (The European Parliament and the Council of the European Union, 2019; Margoni, 2019), which together mitigate the risk associated with using this subset.

Our ethical and legal reasoning for using this government web content---sourced from Common Crawl–related datasets~\citep{commoncrawl} that respect \textit{robots.txt} opt‑out---is as follows:

\begin{itemize}
\item \textbf{Public Purpose Alignment}: The content created by governments is normally meant to be shared with the public, and by using the data for training we are assisting this purpose.

\item \textbf{Purpose of Use}: From a legal perspective, the government works are being redistributed as part of an open source, no-fee dataset, used to create models are less likely to violate copyright. This purpose is clearly not to compete with the government’s own usage.

\item \textbf{Effect on Potential Market}: Our use of government website content is unlikely to affect any potential market for that content, as governments typically do not exploit these materials commercially in ways that would compete with our dataset or downstream models. This factor favors a finding of fair use.

\item \textbf{Nature of the Content}: The nature of the content is mostly public announcements, content of public interest, governmental functions or the like. Again, we believe there is strong public policy interest for fair use of this type of information.

\item \textbf{Amount Used}: While we use all or almost all of the content of the government website, the amount of usage is not determinative of fair-use or not fair-use.

\item \textbf{Federal vs. Non-Federal Works}: Lastly, US works created by the federal governments are generally not copyrightable. However, we recognize that this is not the case for other foreign governmental works, or non-federal works.
\end{itemize}

For these reasons, we believe using government website data presents relatively lower copyright risk. To further minimize risk---for example, the potential inclusion of third‑party copyrighted works embedded in government web pages---we apply keyword filters such as “All Rights Reserved” and “Copyright ©” to exclude pages that contain such terms.

Recent court cases, as of the writing of this paper, include:
\begin{itemize}
    \item \textbf{Bartz v. Anthropic PBC}: district court ruling that use of purchased copies of books for AI training is fair use.
    \item \textbf{Kadrey v. Meta Platforms, Inc.,}: district court ruling that training on authors’ books was transformative fair use.
\end{itemize}
 
These developments lend some support to the argument that AI training on web-text data—including our relatively small, public-facing government subset—can fall within fair use, though the case law is still evolving.

\subsection{Other Tier‑2 Data With Opaque or Mixed Provenance}
Similarly, our dataset includes data whose provenance is not entirely transparent even though the license on the upstream dataset appears permissive, such as \texttt{The Stack V1} and other Tier‑2 sources identified in Appendix~\ref{app:synth_prov}. In the case of \texttt{The Stack V1}, a line‑by‑line audit to remove copyrighted content has not been performed, and therefore some risk remains in its usage. Nonetheless, we rely on fair use to justify this usage because the data are used to train models, rather than to provide a substitutive or competing software product. For a more detailed discussion of \texttt{The Stack V1}, see Section~\ref{app:stack}.

For other Tier‑2 data, some upstream generator models impose conditions on downstream use—such as the Llama license, which requires model users to adhere to certain limitations. We do not believe we are bound by terms that were not contractually passed through to us by our direct licensor, although this issue is subject to debate. We therefore classify this small portion ($\approx$4\%) of the dataset as \textbf{Tier 2(b)}.

There are additional Tier‑2 data where the provenance is partially opaque. For example, a small portion of our P3 dataset, when converted into a few‑shot format, may pose higher risk than other Tier‑1 data. While the ultimate source datasets that constitute P3 are well‑known academic benchmarks, some of those component datasets do not provide explicit licenses. Nonetheless, we consider the resulting few‑shot datasets to be highly transformative and unlikely to compete with the underlying works: they are mixed and reformatted multiple times for the specific purpose of training classification and few‑shot models, rather than, for example, serving as standalone product reviews. We classify these higher‑risk works as \textbf{Tier 2(b)} and include them in our dataset with that caveat.

\subsection{Reliance on EU Text and Data Mining Exceptions}
We also rely, to some extent, on the EU text and data mining exception~\citep{eu_directive_tdm_2019} for our inclusion of web‑crawled data. This regime is complementary to US fair‑use doctrine, and we mention it here for completeness. In particular, we depend on Common Crawl’s practice of respecting \textit{robots.txt} at the time of crawling. We do not believe retroactive recrawling is legally necessary to determine whether a work was subsequently opted out, but we nonetheless commend efforts towards doing so, such as Apertus~\citep{apertus2025apertusdemocratizingopencompliant}.

\subsection{Residual Copyright and Trademark Risks}
The copyright risks in machine learning are complex. For example, copyrighted materials may appear as limited fair‑use quotations in Wikipedia articles~\footnote{\url{https://en.wikipedia.org/wiki/Wikipedia:Quotations##Copyrighted_material_and_fair_use}}. A model trained on such materials in the aggregate could, in principle, generate more substantial and potentially infringing text than the short quotations present in the dataset. Future work should address this risk, including (i) copyright evaluation audits of datasets, and (ii) model‑level mitigations that encourage limited direct quotation and discourage reproduction of substantial protected passages.

As with other large, permissively licensed datasets, additional legal risks remain, including trademark risks. For instance, while training on a Wikipedia article about “Spiderman” may be relatively low risk (given its CC‑BY‑SA license and the educational, summarizing nature of the article), a model that subsequently generates new stories featuring the character name “Spiderman”—even if the plots themselves are not derived from existing human‑created stories—may still implicate trademark rights. Addressing those issues thoroughly is beyond the scope of this work and is left for future research.

We do not and cannot guarantee that, even with rigorous provenance tracking and standard filtering, the dataset is free of legal risk. Nothing in this section constitutes legal advice. We recommend that anyone who uses our datasets consult their own legal counsel in their jurisdiction before deploying models trained on this data in commercial settings.

\section{Provenance and Rationale for The Stack v1 (OpenRAIL-M and terms of use)}
\label{app:stack}

Our inclusion of 53.2B tokens sourced from \textbf{The Stack v1} \citep{kocetkov2023the}, which we categorize by its governing dataset card terms of use and which subsequent model uses the \textbf{OpenRAIL-M} license, warrants this specific note on provenance. The data was included based on the following rationale:

\begin{itemize}
    \item \textbf{Source and Filtering Methodology:} The dataset originates from a large-scale scrape of GitHub. The BigCode project curated this data by applying a filter to include only those repositories that contained a clear permissive license file (e.g., MIT, Apache 2.0, BSD) at the root level.

    \item \textbf{Acknowledged Heuristic:} This repository-level filtering is a \textit{heuristic} and not a file-level guarantee. As acknowledged by the dataset's creators, this process cannot perfectly resolve complex cases of multi-licensing within a single repository, such as the inclusion of non-permissively licensed vendor libraries or mixed-license assets alongside permissively-licensed code.

    \item \textbf{Inclusion Justification:} Despite this caveat, The Stack v1 represents the largest-available public corpus curated with the \textit{explicit goal} of permissive filtering. Excluding it would make training a high-performance, open, and risk-mitigated code model nearly impossible. Its ``best-effort'' permissive curation philosophy directly aligns with our dataset's core principle of risk-mitigation.
\end{itemize}

Thus we include it in our dataset with the classification of Tier-2, as defined in Section~\ref{sec:license_tiers}.

\section{Data Filtering Reasoning and Protocol}
To promote transparency, we describe our protocol for defining and checking the lists and content of the pseudo-crawled portion of \MixtureVitae{}.

\subsection{Governmental and NGO Domain Patterns}
\label{app:gov_domains}
The following list of URL patterns was used to filter for governmental, non-governmental, and international organization websites from the web datasets. We gathered the list by examining public records, Wikipedia lists, and the like. The list is not as simple as \textbf{gov.} because international governments use different TLDs. Moreover, some spam websites masquerades as \textbf{{.gov}} websites. 
Two of the authors examined each domain either online or through the \textit{Internet Archives' Way Back Machine} to confirm they belonged to a government website. After performing a pseudo-crawl on FineFineWeb, Nemotron-CC and MagaCorpus, the authors manually audited the data for quality, and filtered out spam websites with similar website names, which were added to blocklists. 

The\textbf{ .gov, .gov/, and .mil/ }websites are US Federal governmental works. To the extent we could, we filtered any sites that had keywords indicating reservations of rights. We believe this lowers the risk of inadvertent third party copyrighted works appearing on US Federal works, and is in the spirit of the EU text data mining opt-out conventions. We also note that the ultimate source of these websites is from Common Crawl which already also respects the \textit{robots.txt} opt-out.
{\footnotesize
\begin{itemize}
    \item  \texttt{gov} (as a suffix)
    \item \texttt{gov/}
    \item \texttt{mil/}
\end{itemize}
}

All other websites in this category are specifically international governments or NGOs.

{\footnotesize
    \texttt{vlada.mk},
    \texttt{vlada.cz},
    \texttt{kormany.hu},
    \texttt{regeringen.*},
    \texttt{rijksoverheid.nl},
    \texttt{government.nl},
    \texttt{bund.de},
    \texttt{bundesregierung.de},
    \texttt{government.ru},
    \texttt{gc.ca},
    \texttt{admin.ch},
    \texttt{www.gob.cl/},
    \texttt{www.gob.ec/},
    \texttt{guatemala.gob.gt/},
    \texttt{presidencia.gob.hn/},
    \texttt{www.gob.mx/},
    \texttt{presidencia.gob.pa/},
    \texttt{www.gob.pe/},
    \texttt{gob.es/},
    \texttt{argentina.gob.ar/},
    \texttt{tanzania.go.tz/},
    \texttt{indonesia.go.id/},
    \texttt{go.kr/},
    \texttt{go.jp/},
    \texttt{thailand.go.th/},
    \texttt{europa.eu/},
    \texttt{un/},
    \texttt{int/},
    \texttt{govt.},
    \texttt{www.gub.uy},
    \texttt{gov.},
    \texttt{gouv.}
}

\subsection{Curated Permissive Domain List}
\label{app:permissive_domains}
The following list of approximately 50 domains was curated based on their known public domain or CC-BY-SA* license status or a permissive status. The websites were chosen for their diversity of content.  Two of the authors---one of which has a legal background---examined the websites' terms of use, or relevant sections online or on the Way Back Machine to confirm licensing and permission status.  After performing a pseudo-crawl on FineFineWeb, Nemotron-CC and MagaCorpus, the authors manually reviewed the data for quality, and filtered out spam websites with similar website names as the below. These spam sites  were added to blocklists. 
{\footnotesize

     \texttt{free.law}, \texttt{europeana.eu},
    \texttt{publicdomainreview.org},
    \texttt{wisdomcommons.org},
    \texttt{intratext.com},
    \texttt{mediawiki.org},
    \texttt{wikimedia.org},
    \texttt{wikidata.org},
    \texttt{wikipedia.org},
    \texttt{wikisource.org},
    \texttt{wikifunctions.org},
    \texttt{wikiquote.org},
    \texttt{wikinews.org},
    \texttt{wikivoyage.org},
    \texttt{wiktionary.org},
    \texttt{wikibooks.org},
    ,\texttt{courtlistener.com/\footnote{For \texttt{courtlistener.com}, the terms of use says it is CC-BY-ND, but the underlying court cases are public domain, and the content from this website is merely 176KB and is de minimus.}},
    \texttt{case.law},
    \texttt{pressbooks.oer.hawaii.edu},
    \texttt{huggingface.co/docs},
    \texttt{opencourselibrary.org},
    \texttt{medbiq.org},
    \texttt{doabooks.org},
    \texttt{bccampus.ca},
    \texttt{open.umn.edu/opentextbooks},
    \texttt{www.gutenberg.org},
    \texttt{mozilla.org},
    \texttt{www.eclipse.org},
    \texttt{apache.org},
    \texttt{python.org},
    \texttt{pytorch.org},
    \texttt{numpy.org},
    \texttt{scipy.org},
    \texttt{opencv.org},
    \texttt{scikit-learn.org},
    \texttt{pydata.org},
    \texttt{matplotlib.org},
    \texttt{palletsprojects.com},
    \texttt{sqlalchemy.org},
    \texttt{pypi.org},
    \texttt{sympy.org},
    \texttt{nltk.org},
    \texttt{scrapy.org},
    \texttt{owasp.org},
    \texttt{creativecommons.org},
    \texttt{wikia.com},
    \texttt{foodista.com},
    \texttt{fandom.com},
    \texttt{attack.mitre.org}

}

The vast majority of these sites are CC-BY licensed. However, there are some that have other open licenses as shown in Table~\ref{tab:licenses}.

\begin{table}[ht]
\caption{Software Licenses and Associated Websites}

\centering
\begin{tabular}{|l|l|}
\hline
\textbf{License} & \textbf{Websites} \\
\hline
BSD 3-Clause & \begin{tabular}[t]{@{}l@{}}
scipy.org, sympy.org, matplotlib.org, scrapy.org, \\
scikit-learn.org, pydata.org, pytorch.org, \\
palletsprojects.com
\end{tabular} \\
\hline
Mozilla Public License & mozilla.org \\
\hline
Python Software Foundation License 2.0 & python.org \\
\hline
Apache 2.0 & apache.org, nltk.org, opencv.org \\
\hline
MIT License & sqlalchemy.org \\
\hline
Eclipse Public License & www.eclipse.org \\
\hline
MedBiquitous Standards Public License & medbiq.org \\
\hline
\end{tabular}
\label{tab:licenses}
\end{table}

\section{Synthetic Data Source Provenance}
\label{app:synth_prov}

To ensure full transparency regarding the "permissive-first" nature of \MixtureVitae{}, we provide a detailed provenance audit of our synthetic data components in Table~\ref{tab:provenance}, including classification to tiers as defined in Section~\ref{sec:license_tiers}.

\begin{table*}[htbp]
\vspace{-1.3cm}
\caption{Detailed provenance of synthetic data sources in \MixtureVitae{}.}

\centering
\scriptsize
\setlength{\tabcolsep}{3pt}      
\setlength{\extrarowheight}{0pt}
\renewcommand{\arraystretch}{0.96} 
\vspace{-0.2cm}

\begin{tabularx}{\textwidth}{
  >{\raggedright\arraybackslash}p{2.6cm}   
  >{\centering\arraybackslash}p{1.3cm}     
  >{\centering\arraybackslash}p{1.0cm}     
  >{\raggedright\arraybackslash}X          
  >{\centering\arraybackslash}p{0.9cm}     
  >{\raggedright\arraybackslash}X          
}
\textbf{Dataset Name} & \textbf{\shortstack[r]{Dataset \\ License}} & \textbf{Model} & \textbf{\shortstack[r]{Seed Data \\ Provenance}} & \textbf{\shortstack[c]{Token\\Count(B)}} & \textbf{Notes} \\
\rowcolor{green!7} \multicolumn{6}{l}{\textbf{Tier 1: Fully Permissive ($\approx$ 161B Tokens)}} \\
\href{https://huggingface.co/datasets/glaiveai/reasoning-v1-20m}{GlaiveAI Reasoning} & Apache 2.0 & Permissive & N/A & 38.366 & Fully synthetic \\
\href{http://nvidia/Llama-Nemotron-Post-Training-Dataset}{Nemotron (Science \& Math)} & CC-BY-4.0 & Permissive & Permissive (StackOverflow, WildChat) & 22.310 & Science \& Math subset of Llama-Nemotron-Post-Training-Dataset \\
\href{https://huggingface.co/datasets/inclusionAI/Ling-Coder-SyntheticQA}{Ling-Coder/SyntheticQA} & Apache 2.0 & Permissive & N/A & 19.852 &  \\
\href{https://huggingface.co/datasets/open-thoughts/OpenThoughts3-1.2M}{Open Thoughts} & Apache 2.0 & Permissive & Permissive (OpenMath-2-Math, CodeGolf, OpenCode, etc) & 18.786 & Excludes Organic Chemistry subset \\
EuroPat & Public Domain & Permissive & Permissive & 11.586 & Synthetic image captions created from patents \\
\href{https://huggingface.co/datasets/Muennighoff/P3}{P3 (Permissive Subset)} & Apache 2.0 & N/A &
Permissive (ARC, PIQA, BoolQ, etc) &   10.130  & \\
\href{https://huggingface.co/datasets/nvidia/Nemotron-CC-v2}{Nemotron-CC} & Common Crawl ToS & Permissive & Permissive (Common Crawl) & 6.230 & Using a Permissive-only subset \\
YouTube & CC-BY-4.0 & Permissive & Permissive (VALID, CommonCorpus) & 7.386 & Derived from CC-BY licensed YouTube content \\
\href{https://huggingface.co/datasets/nvidia/Nemotron-PrismMath}{Prism-Math} & CC-BY-4.0 & Permissive & Permissive (NuminaMath-1.5) & 5.682 &  \\
\href{https://huggingface.co/datasets/deepmind/math_dataset}{DeepMind Math} & Apache 2.0 & N/A & Permissive (Procedurally Generated) & 4.232 &  \\
\href{https://huggingface.co/datasets/nvidia/OpenMathInstruct-1}{Misc. Instruct. / NVidia OpenMathInstruct-1} & NVIDIA license & Permissive & Permissive (GSM8K, MATH) & 2.440 &  \\
 \href{https://huggingface.co/datasets/HuggingFaceM4/WebSight}{Websights} & CC-BY-4.0 & Permissive & N/A & 1.018 & Fully synthetic \\
\href{https://huggingface.co/datasets/oumi-ai/MetaMathQA-R1}{Misc Instruct. / MetaMathQA-R1}
(responses) & MIT & Permissive & Permissive (GSM8K, MATH) & 0.672 &  \\
Math Word Problems & Apache 2.0 & N/A & Permissive & 0.456 & Procedurally Generated \\
\href{https://huggingface.co/datasets/inclusionAI/Ling-Coder-DPO}{Ling-Coder/DPO} & Apache 2.0 & Permissive & Unknown (Common-Crawl) & 0.398 &  \\
\href{https://huggingface.co/datasets/open-r1/OpenR1-Math-220k}{Misc. Instruct. / \newline  OpenR1-Math-220k} & Apache 2.0 & Permissive & Permissive (NuminaMath-1.5) & 0.320 &  \\
\href{https://huggingface.co/datasets/nvidia/sft_datablend_v1}{Misc. Instruct. / \newline  NVIDIA SFT Datablend} & CC-BY-4.0 & Permissive & Permissive (MNLI, COPA, PIQA, etc) & 0.286 &  \\
\href{https://huggingface.co/datasets/open-r1/OpenThoughts-114k-Code_decontaminated}{Misc. Instruct. / \newline  OpenThoughts-114k-Code (decontaminated)} & Apache 2.0 & Permissive & Permissive (TACO, Apps , CodeContests, etc) & 0.150 &  \\
\href{https://huggingface.co/datasets/VishaalY/synthetic-code-generations}{Misc. Instruct. / \newline Synthetic Code Generations} & Apache 2.0 & Permissive & N/A & 0.104 & Fully synthetic \\
\href{https://huggingface.co/datasets/PrimeIntellect/stackexchange-question-answering}{Misc. Instruct. / \newline  PrimeIntellect StackExchange QnA} & Apache 2.0 & N/A & N/A & 0.076 &  \\
\href{https://huggingface.co/datasets/PrimeIntellect/real-world-swe-problems}{Misc. Instruct. / \newline  PrimeIntellect Real World SWE Problems} & Apache 2.0 & Permissive & Permissive (CommitPack) & 0.006 &  \\
\href{https://huggingface.co/datasets/PrimeIntellect/synthetic-code-understanding}{Misc. Instruct. / \newline  PrimeIntellect Synthetic Code Understanding} & Apache 2.0 & Permissive & N/A & 0.004 & Fully synthetic \\
\href{https://huggingface.co/datasets/openai/gsm8k}{Misc. Instruct. / \newline  GSM8K (train)} & MIT & Permissive & N/A & 0.004 & Fully synthetic \\
\rowcolor{yellow!10} \multicolumn{6}{l}{\textbf{Tier 2(a): Permissive with Upstream Opacity ($\approx$ 35B Tokens)}} \\
\href{https://huggingface.co/datasets/nvidia/OpenScience}{OS-Q2 (OpenScience)} & CC-BY-4.0 & Permissive & N/A & 17.366 & Fully synthetic \\
\href{https://huggingface.co/datasets/inclusionAI/Ring-lite-sft-data}{Ring-lite SFT Data} & Apache 2.0 & Permissive & Permissive (CodeContest, APPS, TACO, etc) & 14.968 &  \\
\href{https://huggingface.co/datasets/hkust-nlp/CodeIO-PyEdu-Reasoning}{PyEdu Reasoning} & Stack V1, ODC-BY & Permissive & Permissive (The Stack V1) & 3.138 &  \\
\href{https://huggingface.co/datasets/Magpie-Align/Magpie-Phi3-Pro-1M-v0.1}{Misc. Instruct. / \newline  Magpie-Phi3-Pro-1M-v0.1} & N/A & Permissive & N/A & 0.386 & Fully synthetic \\
\href{https://huggingface.co/datasets/hltcoe/megawika}{MegaWika} & CC-By-SA/4.0 & Permissive & Permissive (Wikipedia) & 0.356 &  \\
\href{https://huggingface.co/datasets/Magpie-Align/Magpie-Qwen2.5-Coder-Pro-300K-v0.1}{Misc. Instruct. / \newline  Magpie-Qwen2.5-Coder-Pro-300K-v0.1} & N/A & Permissive & N/A & 0.120 & Fully synthetic \\
\href{https://huggingface.co/datasets/NovaSky-AI/Sky-T1_data_17k}{Misc. Instruct. / \newline  NovaSky-AI Sky-T1} & Apache 2.0 & Permissive & Permissive (AIME, MATH, etc) & 0.080 &  \\
\href{https://huggingface.co/datasets/bigcode/self-oss-instruct-sc2-exec-filter-50k}{Misc. Instruct. / \newline  BigCode Self-OSS-Instruct} & Stack v1 & Permissive & Permissive & 0.018 &  \\
\href{https://huggingface.co/datasets/openbmb/UltraFeedback}{Misc. Instruct. / \newline  UltraFeedback} & MIT & Permissive & Permissive (UltraChat, 
TruthfulQA, etc) & 0.014 &  \\
\href{https://huggingface.co/datasets/nguyenkhanh87/CaseHOLD_Phi4_Reasoning}{Misc. Instruct. / \newline  CaseHOLD (Phi4 Reasoning Traces)} & N/A & Permissive & Permissive (CaseHOLD) & 0.002 & Case law is public domain  \\
\rowcolor{orange!10} \multicolumn{6}{l}{\textbf{Tier 2(b): Restricted, Mixed or Opaque Provenance ($\approx$ 17B Tokens)}} \\
\href{https://huggingface.co/datasets/inclusionAI/Ling-Coder-SFT}{Ling-Coder/SFT} & Apache 2.0 & Permissive & Partially Unknown (Github, CommonCrawl, The Stack, etc) & 7.668 & Unknown provenance of CommonCrawl subset \\
P3 (Commercial Subset) & Apache 2.0  & N/A &
Ambiguous-license (Amazon, Rotten Tomatoes, IMDb, etc) & 5.730 &
User-generated content on commercial platforms with no verifiable licenses \\
\href{https://huggingface.co/datasets/nvidia/OpenMathInstruct-2}{Misc. Instruct. / \newline  OpenMathInstruct-2} & CC-By-4.0 & Restricted & Permissive (GSM8K and MATH) & 3.202 & Llama-generated responses \\
\href{https://huggingface.co/datasets/CharlieDreemur/OpenManus-RL}{Misc. Instruct. / \newline  OpenManus-RL} & Apache 2.0 & Mixed & Permissive (AgentTraj-L, Agent-FLAN, etc) & 0.024 & GPT-4 used for generating traces in the AgentTraj-L subset \\
\rowcolor{yellow!10} \multicolumn{6}{l}
{\textbf{Tier 3: Civic / Governmental Works ($\approx$ 15B Tokens)}} \\
\href{https://huggingface.co/datasets/nvidia/Nemotron-CC-v2}{Nemotron-CC (Gov. Portion)} & Common Crawl ToS & Permissive & Permissive (Common Crawl) & 8.531 & Using a Permissive-only subset \\
\href{https://huggingface.co/datasets/ByteDance-Seed/mga-fineweb-edu}{MagaCorpus} & ODC-By & Permissive & Permissive (FinewebEdu-dedup) & 6.596 & Using only permissive subsets \\
\end{tabularx}
\label{tab:provenance}
\end{table*}

\begin{figure}[htbp]
    \centering
        \includegraphics[width=0.48\linewidth]{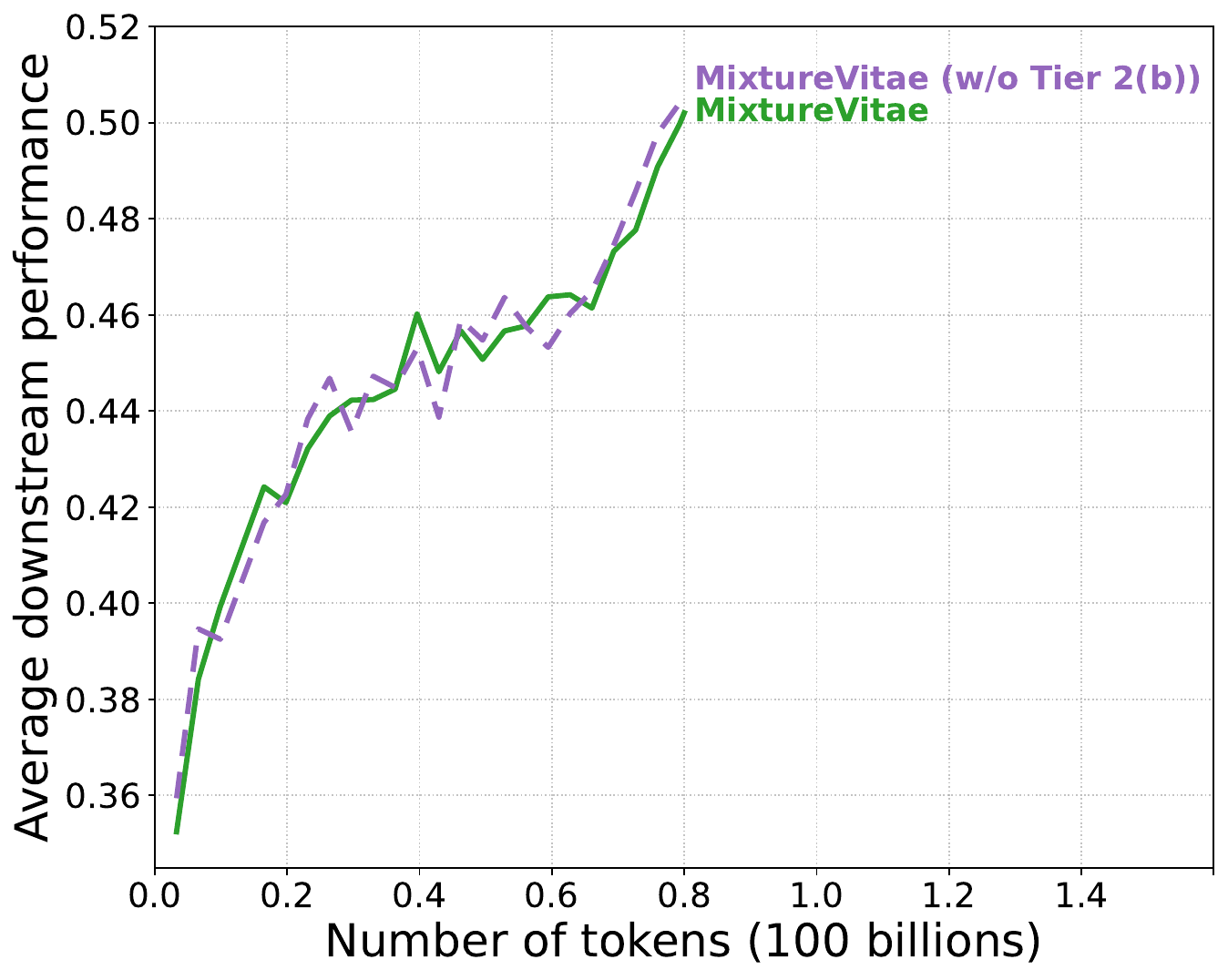}
        \caption{\textbf{Ablation of Tier 2(b) components.} We compare the training trajectory of the full \MixtureVitae{} dataset (green) against a version excluding Tier 2(b) (purple dashed). Tier 2(b) consists of synthetic data derived from non-permissive generators (e.g., Llama-3) or seeds with opaque provenance. The nearly identical performance curves demonstrate that users requiring strict permissive compliance can exclude these components with negligible impact on downstream model quality.}
        \label{fig:ablate_tier2b}
\end{figure}

To validate the robustness of our permissive-first strategy, we further analyze the contribution of synthetic components categorized as \textbf{Tier 2(b)}. This small part of \MixtureVitae{} is comprised of subsets which are permissively licensed (e.g., Apache 2.0) but are derived from generator models with restrictive community licenses (such as Llama-3) or seed data with partially opaque origins. As illustrated in Figure~\ref{fig:ablate_tier2b}, removing these Tier 2(b) components yields a training trajectory indistinguishable from the full \MixtureVitae{} baseline. This result confirms that our model's strong performance is driven by its core, fully verifiable permissive sources, ensuring that users with strict compliance requirements can safely exclude Tier 2(b) data without compromising downstream quality.

See Section~\ref{app:govt_works} for a further discussion on our justification for including Tier 2 and in particular Tier 2(b) data.

\section{Scaling Outlook and Future Directions}
\label{app:scaling_outlook}

While \MixtureVitae{} currently comprises \DatasetSize{} billion tokens---a scale smaller than frontier runs which often exceed 10 trillion tokens---our primary objective in this work was to establish a proof-of-concept for data efficiency and strong downstream performance within a strict permissive-first, risk-mitigated licensing framework. We identify several concrete avenues to scale this approach to the multi-trillion token regime required for larger foundation models:

\paragraph{Subset Upsampling.} 
Standard industry recipes for large-scale training often heavily upsample high-quality data. For instance, Llama~3~\citep{Meta2024_Llama3} employs upsampling factors of 4--10$\times$ for its highest-quality subsets to reach its training budget. In contrast, the current iteration of \MixtureVitae{} does not assign aggressive upsampling factors to individual shards. Applying standard upsampling techniques to our highest-value subsets (such as curated reasoning) would immediately scale their contribution to the total token count.

\paragraph{Multilingual Expansion.} 
The current release of \MixtureVitae{} is primarily English-centric. Expanding the sourcing strategy to include multilingual data represents an order-of-magnitude opportunity for scaling. This can be achieved through two primary methods: (1) identifying and allowing international permissively licensed sources, and (2) using machine translation to expand the existing data in \MixtureVitae{}.

\paragraph{Synthetic Expansion.} 
Our Math and Reasoning synthetic subsets are generated procedurally or via LLMs. This generation process is horizontally scalable. By increasing the compute budget for generation, these high-density subsets can be expanded significantly without incurring the legal risks associated with scraping organic web data.

\paragraph{Web Data Rephrasing.} 
Recent work has demonstrated the utility of rephrasing web data to improve quality and standardize style \citep{maini2024rephrasing}. Applying a similar rephrasing pipeline on top of the \MixtureVitae{} web data processing pipeline can further expand the corpus volume while maintaining the strict safety and licensing standards defined in our framework.

\section{Author contributions}
\label{appendix:author_contributions}

\begin{itemize}

\item \textbf{Huu Nguyen}: Overall lead, data pipeline engineering, data curation, dataset composition, data review, decontamination analysis, policy design, paper co-writing  

\item \textbf{Victor May}: Paper writing lead, project coordination, policy design, decontamination protocol design and analysis

\item \textbf{Harsh Raj}: Led model training, evaluations and ablation experiments, decontamination analysis, paper co-writing

\item \textbf{Marianna Nezhurina}: Model training, established major parts of the dataset tokenization and training infrastructure (Megatron-LM container based workflow), conducted scaling tests for distributed training, wrote routines for evaluation based on \textbf{lm-eval-harness}, conducted  model training, performed evaluations, ablations and paper co-writing  

\item \textbf{Yishan Wang}: Paper co-writing

\item \textbf{Yanqi Luo} Paper co-writing

\item \textbf{Minh Chien Vu} Dataset curation, multimodal text generation

\item \textbf{Taishi Nakamura} Implemented checkpoint conversion routines from Megatron to HuggingFace format, paper co-writing

\item \textbf{Ken Tsui}: Data curation, filtering and classifier training

\item \textbf{Van Khue Nguyen} Dataset curation, multimodal text generation

\item \textbf{David Salinas}:  Performing evaluations and visualizations, paper co-writing

\item \textbf{Aleksandra Krasnodębska}:  Red teaming evaluations, paper co-writing

\item \textbf{Christoph Schumann}: Synthetic math word problems dataset co-design, advising

\item \textbf{Mats Leon Richter}: Dataset curation and processing, advising

\item \textbf{Xuan-Son (Sonny) Vu}: advising and paper co-writing

\item \textbf{Jenia Jitsev}: Supervision, compute resource acquisition. Reference baseline experiments (open-sci-ref) design and model training. Organized dataset acquisition and transfer across the supercomputers. Co-established environments for experiments across various supercomputers. Evaluations co-design, results analysis. Paper co-writing.

\end{itemize}

\begin{table}[htbp]
\centering
\caption{Author contributions to this work. Large squares (\major) indicate a major contribution and plus symbols (\minor) indicate a supporting contribution.}
\label{tab:contributions}
\resizebox{.99\linewidth}{!}{
\begin{NiceTabular}{l*{9}{c}}[
    hvlines, 
    rules/color={gray!50}, 
    cell-space-limits=4pt  
]
& \shortstack{Dataset Curation \&\\ Processing}
& \shortstack{Policy}
& Decontamination
& \shortstack{Model \\ Training}
& \shortstack{Evaluation \& \\ Red Teaming}
& Writing
& Visualization
& \shortstack{Leadership \&\\ Coordination}
& \shortstack{Advising}
\\

Huu Nguyen              & \major & \major & \major &        &        &      \minor   &       & \major &        \\
Victor May              &        & \minor & \major &        &        & \major & \major & \major &        \\
Harsh Raj               &        &        &  \minor & \major & \major &  \minor      & \minor &        &        \\
Marianna Nezhurina      &        &        & \minor       & \major & \major &  \minor      &  \minor      &        &        \\
Yishan Wang             &        &        &        &        &        & \major &        &        &        \\
Yanqi Luo               &        &        &        &        &        &   \minor     & \major &        &        \\
Minh Chien Vu           & \minor &        &        &        &        &        &        &        &        \\
Taishi Nakamura         &        &        &        & \minor &        &   \minor     &        &        &        \\
Ken Tsui                & \minor &        &        &        &        &        &        &        &        \\
Van Khue Nguyen         & \minor &        &        &        &        &        &        &        &        \\
David Salinas           &        &        &        &        & \minor &   \major     &    \minor    &        &        \\
Aleksandra Krasnodębska &        &        &        &        & \minor &        &        &        &        \\
Christoph Schuhmann     & \minor &        &        &        &        &        &        &        &   \minor     \\
Mats Leon Richter       &   \minor      &        &        &        &        & &        &        &   \minor     \\
Xuan-Son (Sonny) Vu     &        &        &        &        &        &  \minor      &        &        & \minor \\
Jenia Jitsev            & \minor &        &        & \major & \major & \minor &  \minor      & \major & \major \\
\CodeAfter
    \tikz \draw (1-|1) rectangle (last-|last) ;
\end{NiceTabular}}
\end{table}
\end{document}